\documentclass[10pt]{article}

\usepackage[preprint]{tmlr}

\usepackage{microtype}
\usepackage{amsmath,amssymb,amsthm,mathtools,bm}
\usepackage{graphicx}
\usepackage{booktabs}
\usepackage{multirow}
\usepackage{array}
\usepackage{placeins}
\usepackage{tikz}
\usetikzlibrary{positioning,arrows.meta}
\usepackage{hyperref}
\usepackage{url}

\allowdisplaybreaks
\newtheorem{theorem}{Theorem}[section]
\newtheorem{lemma}[theorem]{Lemma}
\newtheorem{corollary}[theorem]{Corollary}
\newtheorem{proposition}[theorem]{Proposition}
\newtheorem{assumption}[theorem]{Assumption}
\theoremstyle{remark}

\newcommand{\R}{\mathbb{R}}
\newcommand{\E}{\mathbb{E}}
\newcommand{\PP}{\mathbb{P}}
\newcommand{\N}{\mathcal{N}}
\newcommand{\diag}{\operatorname{diag}}
\newcommand{\tr}{\operatorname{tr}}

\newcommand{\JMOGW}{J_{\mathrm{MOG}}^{W_2}}
\newcommand{\JMOGM}{J_{\mathrm{MOG}}^{\mathrm{MSE}}}

\title{Mixture-of-Gaussians-Guided Schedule Design for Brownian Bridge Diffusion Models}

\author{\name Ron Levi \email ron.levi@campus.technion.ac.il \\
  \addr Department of Computer Science\\
  Technion, Israel
  \AND
  \name Michael Elad \email elad@cs.technion.ac.il \\
  \addr Department of Computer Science\\
  Technion, Israel
}

\begin{document}

\maketitle

\begin{abstract}
Brownian Bridge Diffusion Models (BBDM) offer an appealing framework for image restoration and inverse problems by constructing a stochastic bridge from the clean signal directly to the degraded observation, rather than to pure noise. Despite their promise, the choice of bridge schedule is typically inherited from heuristics, and a principled analytical framework for schedule design has been lacking. In this work, we develop such a framework by offering a novel analysis of BBDM reverse dynamics under a Mixture-of-Gaussians (MoG) prior. This setting yields a closed-form ideal posterior and a corresponding MMSE denoiser, while the BBDM-induced reconstruction law is captured analytically through a tractable surrogate. Building on these expressions, we formulate two complementary schedule-design objectives: a Wasserstein criterion targeting perceptual quality and an MSE criterion targeting reconstruction fidelity. Our work exposes an inherent tradeoff between the two and proves the existence of universal schedules for both that are independent of the degradation and prior. Extensive experiments on controlled MoG settings confirm full alignment between theory and practice, and experiments on the FFHQ dataset across inpainting, deblurring, and super-resolution tasks validate the practical value of our schedule-design criteria.
\end{abstract}

\section{Introduction}

Diffusion models can be used beyond unconditional generation of images, handling image-to-image translation and solving inverse problems.\footnote{The authors used LLMs for assistance with
LaTeX formatting, manuscript presentation and limited code generation and
debugging. All scientific ideas,
claims, derivations, results, code, and final content were reviewed and
approved by the authors.}
In the standard DDPM/DDIM formulations~\citep{ho2020ddpm,song2021ddim}, the forward process gradually corrupts a clean signal towards a terminal state that is close to a Gaussian noise, and the reverse process starts from such a noise and iteratively carves out visual content via a series of denoising steps. When applied to linear inverse problems of the form $y=Hx+n$, where $H$ is a known degradation operator and $n$ is measurement noise, conditional diffusion variants can incorporate this observation by providing $y$ to the denoiser either explicitly or implicitly \citep{saharia2021sr3,saharia2022palette, lugmayr2022repaint,kawar2022ddrm,wang2023ddnm,chung2022dps,song2023pseudoinverse}.

Brownian Bridge Diffusion Models (BBDM) offer a different conditional construction \citep{li2023bbdm}. Instead of diffusing a clean signal towards pure noise, BBDM constructs a stochastic bridge from the clean signal to the conditioning observation $y$. Consequently, the reverse chain starts from the observation itself. For inverse problems, this is an appealing viewpoint: the measurement is not only side information supplied to a network or a guidance term added during inference, but the endpoint of the diffusion path.

A natural question for those two mechanisms is how their inference behavior depends on their schedules. In a conditional DDIM model, the schedule controls the usual data-to-noise path. In BBDM, the bridge schedule controls the path between the clean signal and the observation. These choices directly affect the reconstruction process, yet they are often inherited from heuristics. Our goal is to develop a method-agnostic analytical framework that describes the induced reconstruction laws and uses them for principled schedule design. This paper focuses on BBDM, yet the same framework is also applicable to a conditional DDIM sampler, allowing a direct comparison between the two approaches.

A useful route towards developing such a framework is to start from a tractable model for the clean-signal distribution. Under a single Gaussian prior, exact analysis becomes possible and can be used to derive closed-form reconstruction laws and schedule-design criteria. Two recent papers by \citet{benita2025spectral,benita2026posterior} follow this route. The first analyzes standard DDPM/DDIM prior sampling under a Gaussian signal model and uses the resulting transfer-function view for schedule design. The second studies training-free posterior sampling methods for inverse problems, including DPS and PiGDM-style samplers~\citep{chung2022dps,song2023pseudoinverse}, again under a Gaussian prior and in a setting where the degradation operator and the prior covariance are jointly diagonalizable, sharing a Fourier basis. These works demonstrate the value of exact analysis, but their single-Gaussian and shared-basis assumptions restrict the class of signal models and inverse problems that can be analyzed directly.

In this work we extend the above-described analytical program in two ways: Moving from a single Gaussian prior to a Mixture-of-Gaussians (MoG) one, and enabling an analysis of any linear degradation. The MoG model is richer while still permitting exact posterior formulas. We first derive the exact posterior and the corresponding MMSE denoiser, which gives a complete analytical description at the one-step level. The difficulty appears when this exact MMSE denoiser is inserted into the BBDM reverse sampler: the posterior component probabilities depend on the current latent state. We call the resulting nonlinear sampler the oracle MoG BBDM chain. Although its denoiser is exact, this chain no longer has a fixed affine form that can be unrolled into a simple end-to-end reconstruction law.

To recover tractability, we introduce an approximation that leans on a selected-label surrogate. We first draw a discrete label according to the component probabilities implied by the measurement alone, and then keep this label fixed throughout the reverse chain. 
Once the label is fixed, the reverse dynamics become an affine Gaussian recursion yielding an explicit reconstruction law. 
We show that this law can be analyzed in a component-wise basis, applicable for any prior and degradation operator. 
 
We further show that this surrogate is mean-exact relative to the corresponding target posterior law \(p(x_0\mid y)\), and its only discrepancy is a deficit in posterior covariance.

The above-described derivations enable two complementary schedule-design objectives: a distributional objective, based on the Wasserstein (W2)  distance between the surrogate and the target posterior laws, and an image-to-image objective based on Mean-Squared-Error (MSE) between reconstructions using the two approaches. Whereas the first targets perceptual quality, the second values distortion performance. 

By restricting the BBDM bridge's schedule to a bounded four-parameter family, we prove that the MSE objective admits a universal MSE-oriented schedule, independently of the degradation operator, the measurement-noise level, or the prior. Furthermore, we expose an inherent tradeoff between the MSE and W2 losses: Choosing the schedule that increases MSE leads to a decreased W2 and vice-versa. 

We conclude this work with a broad experimental study that validates the proposed analysis and its practical value. In controlled MoG settings, where the exact posterior is available, we show that the selected-label approximation closely matches oracle BBDM behavior. We then show  
that the frozen surrogate captures scheduling trends in trained BBDM models. Finally, we demonstrate on FFHQ that the resulting MSE- and W2 -oriented schedules produce the expected distortion--perception tradeoff across various restoration tasks.

In summary, our contributions are the following: (i) We derive the exact posterior and MMSE denoiser for BBDM under a Mixture-of-Gaussians prior \footnote{An analogous conditional-DDIM formulation is provided in the Appendix~\ref{app:conditional_ddim_mog_derivation}.}. (ii) Identifying that the MoG reverse process in BBDM loses a global affine structure, we introduce an effective approximation of this computational chain, getting a closed-form expression for the final reconstruction and its dependency on the chosen schedule. 
(iii) We present two complementary design criteria for the BBDM's schedule, trading perceptual quality versus reconstruction distortion. These enable a reliable and direct optimization of the scheduling parameters of BBDM.
 
(iv) We suggest universal problem-agnostic schedules for the two losses, and validate these in various trained BBDM models on the FFHQ dataset.

Sections~\ref{sec:preliminaries}--\ref{sec:comparison_v6} develop the theory, Section~\ref{sec:related_work_main_v76} discusses related work, Section~\ref{sec:experiments_v7} presents the experiments, and the appendices contain the full derivations and proofs.

\section{Background and Problem Setup}
\label{sec:preliminaries}

This section fixes the notation and recalls the ingredients of BBDM used in the analysis. We first describe the bridge construction, then the reverse sampling rule, and finally the inverse-problem handled in the present work.

\subsection{BBDM as a stochastic bridge}

Following \citet{li2023bbdm}, we index the subsequence of steps used at inference time by $s\in\{0,\dots,S\}$. The Brownian-bridge forward marginal is
\begin{equation}
q_{BB}(x_s\mid x_0,y)
=
\N\!\bigl(x_s;(1-m_s)x_0+m_sy,\delta_s I\bigr),
\label{eq:bbdm_forward_marginal_bg_v20}
\end{equation}
where $m_0\cong0$, $m_S\cong1$, and $\delta_0=\delta_S\cong0$. Equation \eqref{eq:bbdm_forward_marginal_bg_v20} shows that the bridge starts at $x_0$ and ends at $y$: as $s$ increases, the mean moves from the clean signal towards the corresponding measurement, while $\delta_s$ controls the randomness around that interpolating path. In our analysis we use the following standard BBDM derived quantities
\begin{gather}
\delta_{s\mid s-1} = \delta_s - \delta_{s-1}\frac{(1-m_s)^2}{(1-m_{s-1})^2},
\label{eq:bbdm_default_notation_2_v20}
\\[2pt]
\tilde{\delta}_s = \frac{{\delta}_{s\mid s-1}\cdot {\delta}_{s-1}}{{\delta}_s}, \qquad \sigma_s = \sqrt{\tilde{\delta}_s}.
\label{eq:bbdm_default_notation_1_v20}
\end{gather}
In the original BBDM formulation, a common reference choice is
\[
\delta_s^{\mathrm{ref}}=2(m_s-m_s^2),
\]
motivated by a variance-preservation argument that unrealistically assumes the bridge endpoints are independent \citep{li2023bbdm}. Because clean signals and target measurements are inherently dependent in inverse problems, we keep $\delta_s$ general to explicitly optimize its shape rather than fixing it in advance.

Throughout this paper we analyze the implementation setting, also used by the released BBDM code, in which the denoiser at step $s$ takes both $x_s$ and $y$ as inputs.

\subsection{Reverse sampling rule}
Once a denoiser supplies an estimate $\hat x_0(x_s,y)$, the BBDM reverse update along this sampling subsequence can be written as
\begin{equation}
x_{s-1}=a_s\hat x_0+b_sy+c_sx_s+\sigma_sz_s,
\qquad
z_s\sim\N(0,I),
\label{eq:bbdm_sampling_form_bg_v20}
\end{equation}
with
\begin{equation}
a_s=(1-m_{s-1})-(1-m_s)\sqrt{\frac{\delta_{s-1}-\sigma_s^2}{\delta_s}},
\label{eq:bbdm_as_bg_v20}
\end{equation}
\begin{equation}
b_s=m_{s-1}-m_s\sqrt{\frac{\delta_{s-1}-\sigma_s^2}{\delta_s}},
\qquad
c_s=\sqrt{\frac{\delta_{s-1}-\sigma_s^2}{\delta_s}}.
\label{eq:bbdm_bc_bg_v20}
\end{equation}

At the first reverse step \(s=S\) the coefficient formulas above are interpreted in the limiting endpoint sense:
\[
x_{S-1}=(1-m_{S-1})\hat x_0+m_{S-1}y+\sqrt{\delta_{S-1}}z_S .
\]

\subsection{Linear measurement model and MOG prior}

The present paper studies BBDM in a linear inverse-problem setting. This is an additional modeling layer on top of the original bridge construction. We assume that the clean signal $x_0\in\R^d$ is observed through the linear model
\begin{equation}
y \mid x_0 \sim \N(Hx_0,\sigma_y^2 I),
\label{eq:measurement_model_main_v20}
\end{equation}
where $H\in\R^{d\times d}$ is a known degradation operator and $\sigma_y^2>0$ is the measurement-noise variance. We further model the data distribution of the clean signal $x_0$ by a Mixture-of-Gaussians,
\begin{equation}
\begin{aligned}
p(x_0)
&=
\sum_{r=1}^{R}\pi_r\,\N(x_0;\mu_r,\Sigma_r),
\\
\Sigma_r&\succ 0,
\qquad r=1,\dots,R,
\\
\pi_r&>0,
\qquad
\sum_{r=1}^{R}\pi_r=1.
\end{aligned}
\label{eq:mog_prior_mainpaper_v20}
\end{equation}
To separate the continuous and discrete uncertainties, we introduce a latent component variable
\begin{equation}
\PP(C=r)=\pi_r,
\qquad
x_0\mid (C=r)\sim \N(\mu_r,\Sigma_r).
\label{eq:mog_latent_C_mainpaper_v20}
\end{equation}
Conditioned on $C=r$, the model remains single Gaussian.

\section{Bayes-Optimal Denoiser under MoG Prior}
\label{sec:mog_posterior}

This section derives the posterior distribution $p(x_0\mid x_s,y)$ under the MoG prior. This posterior is the basic quantity required for the rest of the paper, because BBDM is trained to predict the bridge residual $x_s-x_0$. Under squared loss, the Bayes-optimal residual predictor is
\[
\hat r_s^{\mathrm{MMSE}}(x_s,y)
=
\E[x_s-x_0\mid x_s,y]
=
x_s-\E[x_0\mid x_s,y],
\]
and therefore the induced Bayes-optimal estimate of the clean signal is
\begin{equation}
\hat x_{0,s}^{\mathrm{MMSE}}(x_s,y)
\triangleq
\E[x_0\mid x_s,y].
\label{eq:mog_mmse_def_mainpaper}
\end{equation}

To compute $p(x_0\mid x_s,y)$, we decompose it over the latent component label:
\begin{eqnarray}
p(x_0\mid x_s,y)
=
\sum_{r=1}^{R}p(x_0\mid x_s,y,C=r)\,\PP(C=r\mid x_s,y).
\label{eq:mog_posterior_total_probability_mainpaper}
\end{eqnarray}
Equation~\eqref{eq:mog_posterior_total_probability_mainpaper} splits the derivation into two parts. We first derive the component-conditioned posterior $p(x_0\mid x_s,y,C=r)$, namely the posterior obtained when the active mixture component is fixed. We then derive the posterior probability $\PP(C=r\mid x_s,y)$ of that component and combine the two ingredients. Detailed derivations are collected in Appendix~\ref{app:mog_exact_posterior_derivation}.

\subsection{Component-conditioned posterior}

We begin with the posterior obtained when the active component is fixed. Once $C=r$ and the measurement $y$ are known, the posterior $p(x_0\mid y,C=r)$ is Gaussian. When additionally observing the bridge state $x_s$, the posterior $p(x_0\mid x_s,y,C=r)$ remains Gaussian as well.

\begin{lemma}
\label{lem:component_conditioned_posterior}
Fix a component index $r$. Conditioned on $C=r$, the posterior after observing the measurement is
\[
p(x_0\mid y,C=r)=\N\!\bigl(x_0;\mu_{r\mid y},\Sigma_{r\mid y}\bigr),
\]
with
\begin{equation}
\begin{aligned}
\Sigma_{r\mid y}^{-1}
&=
\Sigma_r^{-1}+\frac{1}{\sigma_y^2}H^\top H, \quad
\mu_{r\mid y}
=
\Sigma_{r\mid y}
\left(
\Sigma_r^{-1}\mu_r+\frac{1}{\sigma_y^2}H^\top y
\right).
\end{aligned}
\label{eq:mog_given_y_component}
\end{equation}
Conditioned on the same component, for an interior bridge step $1\le s\le S-1$, the posterior after observing both the measurement and the bridge state is
\[
p(x_0\mid x_s,y,C=r)=\N\!\bigl(x_0;\mu_{r\mid s}(x_s,y),\Sigma_{r\mid s}\bigr),
\]
where
\begin{equation}
\Sigma_{r\mid s}
=
\left(
\Sigma_r^{-1}+\frac{1}{\sigma_y^2}H^\top H+\frac{(1-m_s)^2}{\delta_s}I
\right)^{-1},
\label{eq:mog_sigma_given_xs_y_mainpaper}
\end{equation}
\begin{equation}
\begin{aligned}
\mu_{r\mid s}(x_s,y)
&=
\Sigma_{r\mid s}
\Biggl(
\Sigma_r^{-1}\mu_r+\frac{1}{\sigma_y^2}H^\top y
+\frac{1-m_s}{\delta_s}(x_s-m_s y)
\Biggr).
\end{aligned}
\label{eq:mog_mu_given_xs_y_mainpaper}
\end{equation}
\end{lemma}

Lemma~\ref{lem:component_conditioned_posterior} shows that, if the active mixture component were known, then the relevant posterior distributions remain Gaussian, with corresponding posterior means that are affine in $(x_s,y)$.

\subsection{Posterior component probabilities and the exact posterior}

We now turn to the second ingredient in \eqref{eq:mog_posterior_total_probability_mainpaper}: the posterior probability of each component. We first compute the measurement-only probabilities $\PP(C=r\mid y)$, and then refine them after observing the bridge state $x_s$.

\begin{lemma}
\label{lem:posterior_component_probabilities}
The posterior probability of component $r$ after observing only $y$ is
\begin{equation}
\begin{aligned}
\gamma_{r\mid y}(y)
&\triangleq \PP(C=r\mid y)
=
\frac{
\pi_r\,\N\!\bigl(y;H\mu_r,H\Sigma_rH^\top+\sigma_y^2I\bigr)
}{
\sum_{j=1}^{R}\pi_j\,\N\!\bigl(y;H\mu_j,H\Sigma_jH^\top+\sigma_y^2I\bigr)
}.
\end{aligned}
\label{eq:mog_gamma_y_mainpaper}
\end{equation}
Moreover, conditioned on $(y,C=r)$,
\begin{equation}
\begin{aligned}
p(x_s\mid y,C=r)
&=
\N\!\bigl(
 x_s;\,(1-m_s)\mu_{r\mid y}+m_sy,
(1-m_s)^2\Sigma_{r\mid y}+\delta_sI\bigr).
\end{aligned}
\label{eq:mog_xs_given_y_mainpaper_v6}
\end{equation}
therefore, after observing both $x_s,y$, we get by Bayes' rule and the law of total probability,
\begin{equation}
\begin{aligned}
\gamma_{r\mid s}(x_s,y)
&\triangleq \PP(C=r\mid x_s,y)
=
\frac{\gamma_{r\mid y}(y)\,p(x_s\mid y,C=r)}{
\sum_{j=1}^{R}\gamma_{j\mid y}(y)\,p(x_s\mid y,C=j)}.
\end{aligned}
\label{eq:mog_gamma_xs_y_mainpaper}
\end{equation}
\end{lemma}

Combining \eqref{eq:mog_posterior_total_probability_mainpaper} with Lemmas~\ref{lem:component_conditioned_posterior} and \ref{lem:posterior_component_probabilities} yields the exact posterior
\begin{equation}
p(x_0\mid x_s,y)
=
\sum_{r=1}^{R}
\gamma_{r\mid s}(x_s,y)
\,\N\!\bigl(x_0;\mu_{r\mid s}(x_s,y),\Sigma_{r\mid s}\bigr),
\label{eq:mog_exact_posterior_mainpaper}
\end{equation}
Since BBDM uses the posterior mean as the Bayes-optimal estimate of $x_0$ under squared loss, we get,
\begin{equation}
\hat x_{0,s}^{\mathrm{MMSE}}(x_s,y)
=
\E[x_0\mid x_s,y]
=
\sum_{r=1}^{R}\gamma_{r\mid s}(x_s,y)\,\mu_{r\mid s}(x_s,y).
\label{eq:mog_exact_mmse_mainpaper}
\end{equation}

Equation~\eqref{eq:mog_exact_mmse_mainpaper} is the exact MoG analogue of the Gaussian oracle denoiser. Closed-form posterior inference is therefore not lost in the MoG case. What changes is the form of the posterior mean: it is now a weighted sum of affine estimators, with state-dependent component probabilities. This is exactly the mechanism that breaks the global affine reverse law in the next section.

\section{The Loss of Global Affinity}
\label{sec:mog_reverse_time}

The previous section provided the exact oracle denoiser. We now insert that denoiser into the BBDM sampler and inspect the resulting reverse update. We will see that in the MoG case, this recursion cannot be unrolled into a direct and simplified closed-form expression for the final reconstruction in terms of the measurement and the reverse Gaussian innovations. The fully expanded derivation is in Appendix~\ref{app:exact_reverse_expansion_v6}.

\begin{lemma}
\label{lem:oracle_reverse_update}
After substituting the exact MMSE denoiser \eqref{eq:mog_exact_mmse_mainpaper} into the BBDM reverse update \eqref{eq:bbdm_sampling_form_bg_v20} for \(1\le s\le S-1\), the oracle MoG BBDM chain becomes
\begin{equation}
\begin{aligned}
x_{s-1}
&= \left( c_s I + a_s\frac{1-m_s}{\delta_s} \sum_{r=1}^{R}\gamma_{r\mid s}(x_s,y)\,\Sigma_{r\mid s} \right) x_s
\\
&\quad+ \Biggl[ b_s I + a_s \sum_{r=1}^{R}\gamma_{r\mid s}(x_s,y)\,\Sigma_{r\mid s} \left( \frac{1}{\sigma_y^2}H^\top - \frac{m_s(1-m_s)}{\delta_s}I \right) \Biggr] y
\\
&\quad+ a_s\sum_{r=1}^{R}\gamma_{r\mid s}(x_s,y)\,\Sigma_{r\mid s}\Sigma_r^{-1}\mu_r + \sigma_sz_s.
\end{aligned}
\label{eq:oracle_reverse_expanded_main_v72}
\end{equation}
where $z_s\sim\N(0,I)$.
\end{lemma}

If $R=1$, then $\gamma_{1\mid s}(x_s,y)\equiv 1$, and all coefficients in \eqref{eq:oracle_reverse_expanded_main_v72} depend only on the schedule and the model parameters. In this case, the oracle MoG BBDM chain is globally affine. Otherwise, the coefficients multiplying $x_s$ and $y$ depend on the current bridge state through the posterior component probabilities $\gamma_{r\mid s}(x_s,y)$. Consequently, unlike the single-Gaussian case, consecutive reverse steps do not share fixed linear operators. This prevents us from recursively composing the equations and obtaining a direct closed-form expression for the final reconstruction $\hat x_0$ in terms of the measurement and the accumulated reverse Gaussian innovations.

\section{The Selected-Label Approximation}
\label{sec:mog_selection_basis}

Section~\ref{sec:mog_reverse_time} showed that the oracle MoG BBDM chain loses global affinity only because the posterior component probabilities $\gamma_{r\mid s}(x_s,y)$ vary with the current state. The selected-label surrogate freezes exactly this source of nonlinearity: after sampling a label from the measurement posterior, it keeps that label fixed throughout the reverse chain.

The approximation proceeds in two steps. First, we choose an auxiliary label $J\in\{1,\dots,R\}$ from the measurement posterior,
\begin{equation}
\PP(J=r\mid y)=\gamma_{r\mid y}(y).
\label{eq:selected_label_rule_main_v74}
\end{equation}
Second, once $J=r$ is selected, we keep that label fixed throughout the reverse chain and use the corresponding component-conditioned posterior mean $\mu_{r\mid s}(x_s,y)$ from Section~\ref{sec:mog_posterior} inside the BBDM update. Detailed derivations are collected in Appendix~\ref{app:selected_label_formulas_v6}.

At the terminal update \(s=S\), since \(x_S=y\), the selected-label
denoiser is interpreted as the measurement-only posterior mean
\(\hat x_0=\mu_{r|y}\). For \(1\le s\le S-1\), substituting
\(\hat x_0=\mu_{r|s}(x_s,y)\) into the BBDM reverse rule gives, conditioned on $J=r$,
\begin{equation}
\begin{aligned}
x_{s-1}
&=
\left(c_s I + a_s\frac{1-m_s}{\delta_s}\Sigma_{r\mid s}\right)x_s
+
\left(b_s I - a_s\frac{m_s(1-m_s)}{\delta_s}\Sigma_{r\mid s}\right)y
+
a_s\Sigma_{r\mid s}\Sigma_{r\mid y}^{-1}\mu_{r\mid y}
+\sigma_s z_s,
\qquad z_s\sim\N(0,I).
\end{aligned}
\label{eq:selected_label_reverse_update_main_v74}
\end{equation}
This is already a major simplification compared with the exact MoG reverse chain of Section~\ref{sec:mog_reverse_time}: once the label is fixed, the one-step update becomes affine in $x_s$ and $y$, with coefficients determined only by the selected component and the schedule.

The next question is whether these one-step updates can be analyzed in one convenient coordinate system. The key observation is that, for a fixed component $r$, all matrices $\Sigma_{r\mid s}$ are obtained from $\Sigma_{r\mid y}^{-1}$ by adding a scalar multiple of the identity and inverting. Therefore, all of them are diagonalized by the same basis.

\begin{lemma}
\label{lem:problem_adapted_basis}
For each fixed component $r$, there exists an orthogonal matrix $U_r$ and positive scalars $\lambda_{r,1},\dots,\lambda_{r,d}$ such that
\begin{equation}
\Sigma_{r\mid y}^{-1}
=
U_r\diag(\lambda_{r,1},\dots,\lambda_{r,d})U_r^\top.
\label{eq:problem_adapted_basis_mainpaper_v74}
\end{equation}
Moreover, for every interior step \(1\le s\le S-1\),
\begin{equation}
\Lambda_{r\mid s}
=
\diag\!\left(
\frac{1}{\lambda_{r,1}+\frac{(1-m_s)^2}{\delta_s}},\dots,
\frac{1}{\lambda_{r,d}+\frac{(1-m_s)^2}{\delta_s}}
\right)
\label{eq:Lambda_r_mainpaper_v74}
\end{equation}
 satisfies
\begin{equation}
\Sigma_{r\mid s}=U_r\Lambda_{r\mid s}U_r^\top.
\label{eq:sigma_rs_problem_basis_mainpaper_v74}
\end{equation}
\end{lemma}

Lemma~\ref{lem:problem_adapted_basis} is the structural replacement for the single global basis available in the one-Gaussian setting. It says that each selected component comes with its own basis $U_r$, and in that basis every posterior covariance along the reverse chain is diagonal. Unlike \citet{benita2025spectral,benita2026posterior}, this construction does not restrict the degradation operator and prior covariance to share a global basis.

We now project the reverse update \eqref{eq:selected_label_reverse_update_main_v74} onto this basis. Define
\begin{equation}
\begin{aligned}
x_s^{U_r}=U_r^\top x_s,
\quad y^{U_r}=U_r^\top y, \quad 
\mu_{r\mid y}^{U_r}=U_r^\top\mu_{r\mid y},
\quad z_s^{U_r}=U_r^\top z_s.
\end{aligned}
\label{eq:projected_quantities_mainpaper_v74}
\end{equation}
Because $U_r$ is orthogonal, $z_s^{U_r}\sim\N(0,I)$.

\begin{lemma}
\label{lem:fixed_label_reverse_recursion}
Conditioned on $J=r$, the projected reverse chain satisfies
\begin{equation}
x_{s-1}^{U_r}
=
G_r(s)x_s^{U_r}+N_r(s)y^{U_r}+M_r(s)\mu_{r\mid y}^{U_r}+\sigma_sz_s^{U_r},
\label{eq:reverse_basis_vector_mainpaper_v74}
\end{equation}
where
\begin{align}
G_r(s)=c_sI+\frac{a_s(1-m_s)}{\delta_s}\Lambda_{r\mid s}, \quad
N_r(s)=b_sI-\frac{a_sm_s(1-m_s)}{\delta_s}\Lambda_{r\mid s}, \quad
M_r(s)=a_s\Lambda_{r\mid s}\diag(\lambda_{r,1},\dots,\lambda_{r,d}).
\end{align}

The formulas above apply for \(1 \le s \le S-1\). For the terminal
reverse update \(s=S\), we use the endpoint convention
\[
\begin{aligned}
G_r(S) = 0, \quad N_r(S) = m_{S-1}I, \quad
M_r(S) = (1-m_{S-1})I, \quad \sigma_S^2 = \delta_{S-1}.
\end{aligned}
\]

For all steps, all three matrices are diagonal. Hence, conditioned on $J=r$, the reverse process decouples into $d$ scalar recursions in the basis $U_r$.
\end{lemma}

Lemma~\ref{lem:fixed_label_reverse_recursion} says that once the label is frozen and the process is written in the basis $U_r$, the reverse chain becomes a diagonal step-by-step update with Gaussian noise. This is exactly what makes it possible to unroll the reconstruction law explicitly.

\begin{corollary}
\label{cor:selected_label_reconstruction_law}
Starting from the BBDM initialization $x_S=y$, the fixed-label reverse chain can be unrolled as
\begin{equation}
\hat x_0^{U_r,(r)}
=
D_1^{(r)}y^{U_r}+D_2^{(r)}\mu_{r\mid y}^{U_r}
+
\sum_{i=1}^{S}\left(\prod_{j=1}^{i-1}G_r(j)\right)\sigma_i z_i^{U_r},
\label{eq:selected_unrolled_mainpaper_v74}
\end{equation}
where
\begin{equation}
\begin{aligned}
D_1^{(r)}
&=
\prod_{s'=1}^{S}G_r(s')
+
\sum_{i=1}^{S}\left(\prod_{j=1}^{i-1}G_r(j)\right)N_r(i), \quad
D_2^{(r)}
=
\sum_{i=1}^{S}\left(\prod_{j=1}^{i-1}G_r(j)\right)M_r(i).
\end{aligned}
\label{eq:D1D2_mainpaper_v74}
\end{equation}
Consequently, the selection approximation law conditioned on a specific gaussian is,
\begin{equation}
p_{\mathrm{Select}}(\hat x_0\mid y,J=r)
=
\N\!\bigl(\hat x_0;\mu_{\mathrm{BBDM}}^{(r)}(y),\Sigma_{\mathrm{BBDM}}^{(r)}\bigr),
\label{eq:selected_law_component_mainpaper_v74}
\end{equation}
with
\begin{equation}
\mu_{\mathrm{BBDM}}^{(r)}(y)
=
U_r\Bigl(D_1^{(r)}y^{U_r}+D_2^{(r)}\mu_{r\mid y}^{U_r}\Bigr),
\label{eq:selected_law_mean_mainpaper_v74}
\end{equation}
\begin{equation}
\Sigma_{\mathrm{BBDM}}^{(r)}
=
U_r\Biggl[
\sum_{i=1}^{S}
\left(\prod_{j=1}^{i-1}G_r(j)\right)\sigma_i^2I
\left(\prod_{j=1}^{i-1}G_r(j)\right)^\top
\Biggr]U_r^\top.
\label{eq:selected_law_cov_mainpaper_v74}
\end{equation}
After removing the conditioning on $J$,
\begin{equation}
p_{\mathrm{Select}}(\hat x_0\mid y)
=
\sum_{r=1}^{R}\gamma_{r\mid y}(y)
\,\N\!\bigl(\hat x_0;\mu_{\mathrm{BBDM}}^{(r)}(y),\Sigma_{\mathrm{BBDM}}^{(r)}\bigr).
\label{eq:selection_induced_mixture_mainpaper_v74}
\end{equation}
\end{corollary}

Corollary~\ref{cor:selected_label_reconstruction_law} summarizes the full payoff of the selected-label approximation. Each possible selected component produces one explicit Gaussian reconstruction law, analyzed in its own basis. The overall selected-label law is then obtained by mixing these Gaussian outputs with the measurement-posterior probabilities.

The unrolled law contains a schedule-dependent mean term through the matrices $D_1^{(r)}$ and $D_2^{(r)}$. However, the next result shows that these matrices simplify quite significantly. 

\begin{proposition}
\label{prop:selected_label_mean_exactness}
Conditioned on the selected label $J=r$, the selected-label BBDM output satisfies
\begin{equation}
\mu_{\mathrm{BBDM}}^{(r)}(y)
\triangleq 
\E[\hat x_0\mid y,J=r]=\mu_{r\mid y}.
\label{eq:mean_exactness_statement_main}
\end{equation}
Equivalently, in the notation of Corollary~\ref{cor:selected_label_reconstruction_law},
\begin{equation}
D_1^{(r)}=0,
\qquad
D_2^{(r)}=I.
\label{eq:D1D2_mean_exact_main}
\end{equation}
\end{proposition}

This proposition shows that the selected-label surrogate does not introduce any component-wise bias. The next proposition shows that the remaining discrepancy is a loss of covariance.

\begin{proposition}
\label{prop:selected_label_covariance_deficit}
In the component-wise basis $U_r$, write
\begin{equation}
U_r^\top\Sigma_{\mathrm{BBDM}}^{(r)}U_r
=
\diag\!\bigl(
\sigma_{\mathrm{BBDM},r,1}^2,\dots,
\sigma_{\mathrm{BBDM},r,d}^2
\bigr),
\label{eq:sigma_bbdm_basis_definition_main}
\end{equation}
where
\begin{equation}
\sigma_{\mathrm{BBDM},r,k}
=
\left(
\sum_{i=1}^{S}\sigma_i^2
\prod_{j=1}^{i-1}g_{r,k}(j)^2
\right)^{1/2},
\label{eq:bddm_sigma_main_v72}
\end{equation}
and $g_{r,k}(j)$ denotes the $k$-th diagonal entry of $G_r(j)$. Then
\begin{equation}
\Sigma_{r\mid y}-\Sigma_{\mathrm{BBDM}}^{(r)}\succeq 0.
\label{eq:cov_deficit_psd_main}
\end{equation}
Equivalently, in the component-wise basis,
\begin{equation}
0
\le
\sigma_{\mathrm{BBDM},r,k}^2
\le
\frac{1}{\lambda_{r,k}},
\qquad
k=1,\dots,d.
\label{eq:variance_underdispersion_main}
\end{equation}
\end{proposition}

Proposition~\ref{prop:selected_label_covariance_deficit} shows that the selected-label BBDM sampler is covariance-deficient. For any valid schedule, it may approach the component posterior covariance $\Sigma_{r\mid y}$ from below, but it cannot exceed it in any direction.

\section{Two Schedule Design Objectives}
\label{sec:comparison_v6}

The previous section produced an explicit surrogate reconstruction law. Schedule design now becomes a comparison problem: for a given schedule, how close is the selected-label reconstruction law $p_{\mathrm{Select}}(\hat x_0\mid y)$ to the target measurement posterior $p(x_0\mid y)$? We begin with the exact posterior after observing only the measurement, then compare it with the selected-label law for the same observation, and finally average the resulting comparison over the measurement distribution. This leads to two complementary objectives. The first is a matched-component Wasserstein upper-bound surrogate for comparing posterior laws. The second is based on expected mean-squared error under an explicit matched-label product coupling and is aligned with
low-distortion, pixel-wise reconstruction accuracy. The full derivations appear in Appendix~\ref{app:schedule_objectives_derivations}.

\begin{lemma}
\label{lem:true_measurement_posterior}
After observing only the measurement $y$,
\begin{equation}
\begin{aligned}
p(x_0\mid y)
&=
\sum_{r=1}^{R}\PP(C=r\mid y)\,p(x_0\mid y,C=r)
=
\sum_{r=1}^{R}
\gamma_{r\mid y}(y)
\,\N\!\bigl(x_0;\mu_{r\mid y},\Sigma_{r\mid y}\bigr).
\end{aligned}
\label{eq:true_posterior_y_mainpaper_v6}
\end{equation}
\end{lemma}

Lemma~\ref{lem:true_measurement_posterior} follows directly from the formulas of Section~\ref{sec:mog_posterior}. The important point for what follows is that the true measurement posterior and the selected-label law share the same component weights $\gamma_{r\mid y}(y)$. They differ only in the Gaussian component attached to each possible label.

\begin{lemma}
\label{lem:pointwise_wasserstein_upper_bound}
For fixed $y$,
\begin{align}
W_2^2\!\bigl(p(x_0\mid y),p_{\mathrm{Select}}(\hat x_0\mid y)\bigr) 
\le 
\sum_{r=1}^{R}\gamma_{r\mid y}(y)
\,W_2^2\!\Bigl(\N(\mu_{r\mid y},\Sigma_{r\mid y}),
\N\!\bigl(\mu_{\mathrm{BBDM}}^{(r)}(y),\Sigma_{\mathrm{BBDM}}^{(r)}\bigr)\Bigr).
\label{eq:pointwise_w2_bound_v72}
\end{align}
\end{lemma}

The bound in Lemma~\ref{lem:pointwise_wasserstein_upper_bound} follows from the definition of $W_2^2$ as an infimum over all couplings. Since the true posterior and the selected-label law have the same mixture weights $\gamma_{r\mid y}(y)$, we build one explicit coupling by using the same random component index in both mixtures and, conditioned on that index, coupling the corresponding Gaussian pair optimally. Evaluating its transport cost therefore gives an explicit upper bound, which reduces the comparison to a weighted sum of Gaussian costs.

\begin{lemma}
\label{lem:matched_gaussian_cost}
Using the standard closed form for the quadratic Wasserstein distance between
Gaussian measures~\citep{dowson1982frechet,peyre2019computational}, the
$r$-th Gaussian cost in Lemma~\ref{lem:pointwise_wasserstein_upper_bound}
is
\begin{align}
&W_2^2\!\Bigl(
\N(\mu_{r\mid y},\Sigma_{r\mid y}),
\N\!\bigl(\mu_{\mathrm{BBDM}}^{(r)}(y),\Sigma_{\mathrm{BBDM}}^{(r)}\bigr)
\Bigr)
\notag
=
\sum_{k=1}^{d}
\left(\sigma_{\mathrm{BBDM},r,k}-\frac{1}{\sqrt{\lambda_{r,k}}}\right)^2.
\label{eq:matched_gaussian_cost_main_v72}
\end{align}
\end{lemma}

Lemma~\ref{lem:matched_gaussian_cost} shows that, after mean exactness is taken into account, the matched-component comparison is purely a covariance comparison. Averaging this expression over the measurement distribution gives the Wasserstein upper-bound schedule objective.

\begin{corollary}
\label{cor:wasserstein_schedule_objective}
The matched-component Wasserstein upper-bound selected-label schedule objective is
\begin{equation}
\JMOGW
=
\sum_{r=1}^{R}\pi_r
\sum_{k=1}^{d}
\left(
\sigma_{\mathrm{BBDM},r,k}-\frac{1}{\sqrt{\lambda_{r,k}}}
\right)^2.
\label{eq:JMoG_mainpaper_v71}
\end{equation}
\end{corollary}

The Wasserstein objective compares the selected-label law with the true measurement posterior at the level of entire Gaussian components. The next objective will evaluate reconstruction similarity directly through expected mean-squared error, rather than compare full distributions.

\begin{corollary}
\label{cor:expected_mse_schedule_objective}
Under the natural product coupling between the matched Gaussian pairs, the expected-MSE objective is
\begin{equation}
\JMOGM
=
\sum_{r=1}^{R}\pi_r
\sum_{k=1}^{d}
\left(
\sigma_{\mathrm{BBDM},r,k}^2
+
\frac{1}{\lambda_{r,k}}
\right).
\label{eq:mog_expected_mse_full_v72}
\end{equation}

\end{corollary}

This is the matched-label product-coupling MSE. If the true posterior sample and the selected-label reconstruction are instead assigned independent mixture labels, an additional component-mean term appears. However, the extra term is independent of the schedule. Thus the schedule optimizer is unchanged, although the absolute MSE value is different.

Corollaries~\ref{cor:wasserstein_schedule_objective} and \ref{cor:expected_mse_schedule_objective} differ only in how they treat the remaining component covariance. The Wasserstein criterion tries to match the posterior standard deviations $1/\sqrt{\lambda_{r,k}}$, whereas the expected-MSE criterion penalizes the sampler variance itself. Thus the Wasserstein objective favors posterior-distribution matching, while the MSE objective favors concentration around the posterior mean. This observation is closely related to the perception--distortion tradeoff of
\citet{blau2018perception}, which formalizes the tension
between distortion-based fidelity and perceptual quality in image restoration.

\begin{figure}[!htbp]
    \centering
    \includegraphics[width=0.9\textwidth]
    {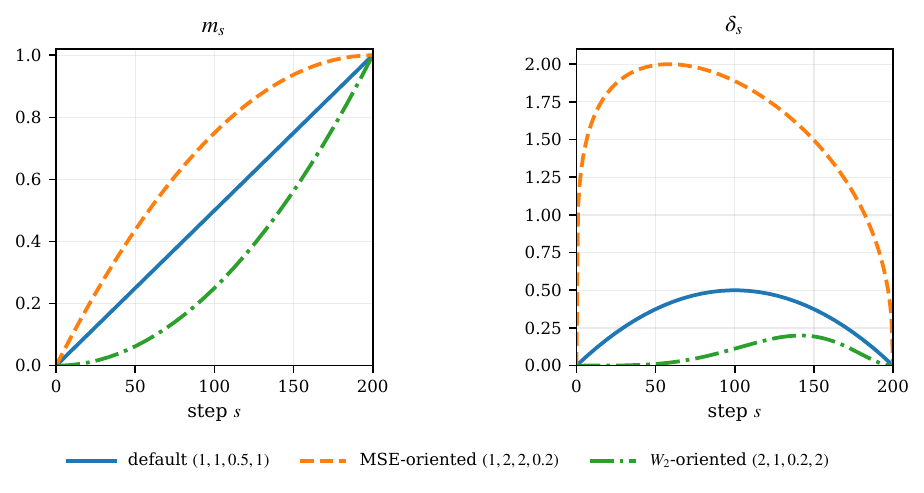}
    \caption{Schedule shapes for the three BBDM schedules used in the FFHQ
    experiments, shown for \(S=200\) sampling steps. Left: interpolation
    coefficient \(m_s\). Right: bridge variance coefficient \(\delta_s\).}
    \label{fig:bbdm_schedule_shapes_s200}
\end{figure}

\medskip
\noindent {\bf Bounded schedule family:} To design an optimal measurement-aware BBDM schedule for the Mixture-of-Gaussians setting, we seek parameters
\[
\Theta=\{(m_s,\delta_s)\}_{s=1}^{S}
\]
that minimize either $\JMOGW(\Theta)$ or $\JMOGM(\Theta)$. Rather than optimize over all discrete pairs directly, we use the bounded four-parameter family
\begin{equation}
\begin{aligned}
m_s &= 1-(1-\tau_s^{\alpha})^{\beta}, \quad
\delta_s = c\,[4m_s(1-m_s)]^{\gamma},
\quad \tau_s=\frac{s}{S},
\end{aligned}
\label{eq:bounded_schedule_family_mainpaper_v6}
\end{equation}
with
\begin{equation}
\begin{aligned}
\alpha\in[1,2],\quad
&\beta\in[1,2],\quad
c\in[0.2,2],\quad
\gamma\in[0.2,2].
\end{aligned}
\label{eq:bounded_schedule_box_mainpaper_v6}
\end{equation}
This keeps the search low-dimensional while enforcing the Brownian-bridge constraints by construction and avoiding degenerate schedules. Appendix~\ref{app:mog_free_schedule_heuristics} derives MoG-free edge rules for this family: the MSE variance bound is minimized at $(\alpha,\beta,c,\gamma)=(1,2,2,0.2)$, while the $W_2$ heuristic uses the opposite edge $(2,1,0.2,2)$ to increase sampler variance toward the posterior spread. These are the problem-agnostic schedules used when reliable high-dimensional MoG covariance estimates are unavailable. Figure~\ref{fig:bbdm_schedule_shapes_s200} visualizes the three BBDM
schedules considered in this work for \(S=200\) sampling steps.

Appendix~\ref{app:conditional_ddim_mog_derivation} repeats the same MoG and selected-label pipeline for a conditional DDIM sampler with an oracle posterior denoiser. The comparison isolates which parts of the analysis are specific to the Brownian bridge and which arise from the broader MoG posterior structure.

Finally, Theorem~\ref{thm:selected_chain_self_consistency} in
Appendix~\ref{app:onehot_proof_outline_v6} provides theoretical support for
the selected-label approximation itself. In a shared-covariance, linearly separated high-dimensional MoG regime, it proves that if the correct label is frozen, then the selected-label chain remains in states where the exact MoG responsibilities continue to favor that same label with overwhelming probability. At the one step level - the full MOG denoiser versus the selected label denoiser discrepancy is negligible.

\FloatBarrier

\section{Related work}
\label{sec:related_work_main_v76}

This paper is closest to three lines of work: bridge-based diffusion,
diffusion methods for inverse problems, and analytical studies of diffusion
inference and schedule design.

\paragraph{Bridge and stochastic-interpolation diffusion.}
BBDM models image-to-image translation through a stochastic bridge whose
endpoint is the target observation~\citep{li2023bbdm}. Closely related,
I$^2$SB learns diffusion bridges between degraded and clean image distributions
for restoration tasks~\citep{liu2023i2sb}. More broadly, Schr{\"o}dinger-bridge
and stochastic-interpolant formulations view diffusion models as stochastic
paths that transport samples between two endpoint distributions over a fixed
time interval~\citep{debortoli2021dsb,albergo2023stochastic}. These works mainly
develop bridge or interpolation constructions and the corresponding training
and sampling procedures. Our focus is not a new bridge
architecture, but an analysis of the reconstruction law induced by a BBDM
sampler and its schedule.

\paragraph{Diffusion models for inverse problems.}
A large body of work applies diffusion models to image restoration and inverse
problems. Conditional diffusion models such as SR3 and Palette train networks
conditioned on degraded observations~\citep{saharia2021sr3,saharia2022palette},
whereas zero-shot methods such as RePaint, DDRM, DDNM, DPS, and PiGDM use
pretrained diffusion models together with measurement constraints or posterior
guidance during sampling
~\citep{lugmayr2022repaint,kawar2022ddrm,wang2023ddnm,chung2022dps,song2023pseudoinverse}.
The main text focuses on the BBDM setting in which the measurement is built into the
bridge endpoint. However, Appendix~\ref{app:conditional_ddim_mog_derivation}
further shows that the same MoG and selected-label mechanism is not specific to
BBDM by deriving the analogous analysis for a conditional DDIM sampler on which many of those works are built.

\paragraph{Analytical studies of diffusion inference.}
Noise schedules and sampler parameterizations are known to strongly affect
diffusion behavior. Improved DDPMs and EDM-style analyses show that performance depends not only on the denoising network, but also on choices
such as the reverse variance, the noise or data parameterization, and the sampler dynamics~\citep{nichol2021improved,karras2022edm}.
Closer to our work, \citet{benita2025spectral,benita2026posterior} derive exact reconstruction laws and
schedule-design criteria under a single-Gaussian prior. We follow this analytical
program but replace the single-Gaussian model by a Mixture-of-Gaussians prior leading to our selected-label surrogate and component-wise basis analysis.

\section{Experiments}
\label{sec:experiments_v7}

The experiments follow a three-stage progression. First, in controlled MoG
models, where the posterior and oracle samplers are clearly available, we validate the
selected-label theory and compare the BBDM sampler with a conditional DDIM sampler. Second, on MNIST~\citep{lecun1998gradient}, where fitting a MoG is still
feasible, we test whether the frozen surrogate can faithfully capture scheduling parameter trends that align with BBDM with learned denoiser. We also show in this case that modeling the prior with more Gaussians directly improves the frozen model reconstruction capabilities. Finally, on FFHQ~\citep{karras2019stylegan}, where reliable high-dimensional MoG fitting is
impractical, we use the MoG-free MSE-oriented and $W_2$-oriented schedule heuristics of
Appendix~\ref{app:mog_free_schedule_heuristics}. Additional plots and
qualitative examples are found in  Appendix~\ref{app:extra_figures_v43}.

\subsection{Theory validation in the shared-covariance regime}

We begin with synthetic shared-covariance MoG data. The self-consistency result in Appendix~\ref{app:onehot_proof_outline_v6} predicts that, when the measurement separation grows with the dimension, the responsibilities should become nearly one-hot and the frozen selected label should remain stable along its own reverse chain. The synthetic study confirms empirically that as the ambient dimension increases, the realized separation statistic approaches its theoretical threshold, while the selected-label error, the denoiser discrepancy, and the final reconstruction discrepancy all decrease rapidly. Empirically, once the measurement margin is large enough, the exact MoG reverse process behaves almost as if a single component had been selected in advance. The full four-panel validation plot appears in Appendix~\ref{app:extra_figures_v43}.

\subsection{Blended tradeoff under oracle MoG and frozen reverse chains}

We next use the random means, shared-covariance toy model to visualize the tradeoff induced by blending the two schedule objectives of Section~\ref{sec:comparison_v6}. For different $\lambda$ values we optimize
\[
J_\lambda=(1-\lambda)\JMOGW+\lambda\JMOGM
\]
over the bounded schedule family and then evaluate the chosen schedule on two empirical axes over $n_y=100$ measurements: mean-squared error to the true signal and the sliced $W_2$ distance ~\citep{bonneel2015sliced} between reconstruction samples and posterior samples. The selected-label surrogate reconstructions are shown as dots and the oracle MoG BBDM reconstructions as crosses.
 For each measurement $y$, the frozen label is sampled from the measurement posterior and is kept fixed along the frozen chain.

\begin{figure}[htbp]
    \centering
    \includegraphics[width=0.498\columnwidth]{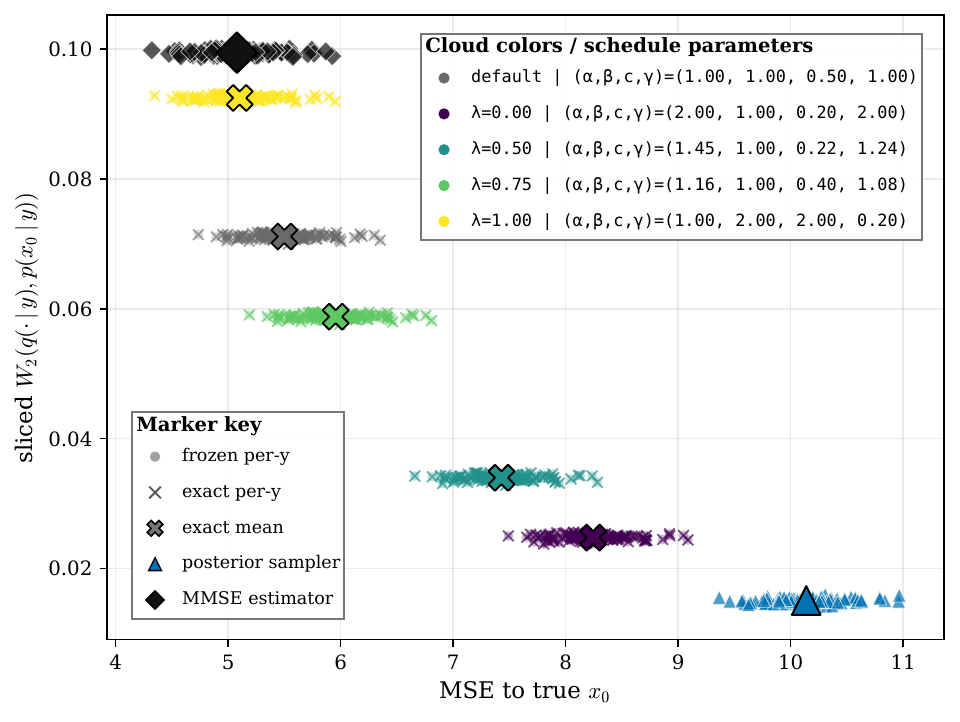}
    \caption{Shared-covariance toy experiment for schedules optimized with
    $J_\lambda$.
    The prior is a uniform-weight MoG with $R=32$ components in dimension
    $d=512$. The component means are sampled independently as
    $\mu_{r,k}\sim\mathrm{Unif}[-1,1]$, the inverse problem is denoising
    with $H=I$ and $\sigma_y=0.1$, and all components share the diagonal
    covariance
    $\Sigma=\operatorname{diag}(\operatorname{geomspace}(0.5,2,d))$.
    The BBDM chains use $S=20$ reverse steps. The plot evaluates the schedules
    optimized for $\lambda\in\{0,0.5,0.75,1\}$, using
    $256$ samples per measurement and $256$ random projections for the sliced
    $W_2$ estimate. The direct posterior sampler and posterior MMSE estimator are included as schedule-independent Bayesian references. The direct posterior sampler has nonzero empirical sliced-$W_2$ because it is compared to independent finite posterior reference samples.}
    \label{fig:toy_tradeoff_overlay_v88}
\end{figure}

\subsection{BBDM--DDIM comparison}
\label{sec:mog_denoise_w2_steps}

We next compare BBDM and conditional DDIM as a function of the number of
reverse steps \(S\). We use the same shared-covariance MoG setting as above,
with \(R=32\), \(d=512\), uniform mixture weights,
\(\mu_{r,k}\sim\mathrm{Unif}[-1,1]\), and
$\Sigma=\operatorname{diag}(\operatorname{geomspace}(0.5,2,d))$.
The inverse problem is denoising, \(H=I\), with \(\sigma_y=0.5\).For each \(S\in\{2,\ldots,1000\}\), we evaluate the two normalized
selected-label objectives
\begin{align*}
\frac{1}{d}J_{\mathrm{MOG}}^{W_2}
&=
\frac{1}{d}\sum_{k=1}^{d}
\left(
\sigma_{\mathrm{model},k}(S)
-
\frac{1}{\sqrt{\lambda_k}}
\right)^2,
\\
\frac{1}{d}J_{\mathrm{MOG}}^{\mathrm{MSE}}
&=
\frac{1}{d}\sum_{k=1}^{d}
\left(
\sigma_{\mathrm{model},k}^2(S)
+
\frac{1}{\lambda_k}
\right),
\qquad
\lambda_k=\Sigma_{kk}^{-1}+\sigma_y^{-2}.
\end{align*}
We compare the default conditional DDIM schedule, where \[
\beta_t
=
10^{-4}
+
\frac{t-1}{T-1}(2\cdot10^{-2}-10^{-4}),
\qquad
t=1,\ldots,T,
\]
and
\[
\bar\alpha_0=1,
\qquad
\bar\alpha_t
=
\prod_{u=1}^{t}(1-\beta_u).
\]

with three BBDM schedules:
the default bridge schedule \((1,1,0.5,1)\), the MSE-oriented schedule
\((1,2,2,0.2)\), and the \(W_2\)-oriented schedule \((2,1,0.2,2)\).

\begin{figure}[htbp]
    \centering
    \includegraphics[width=0.98\textwidth]
    {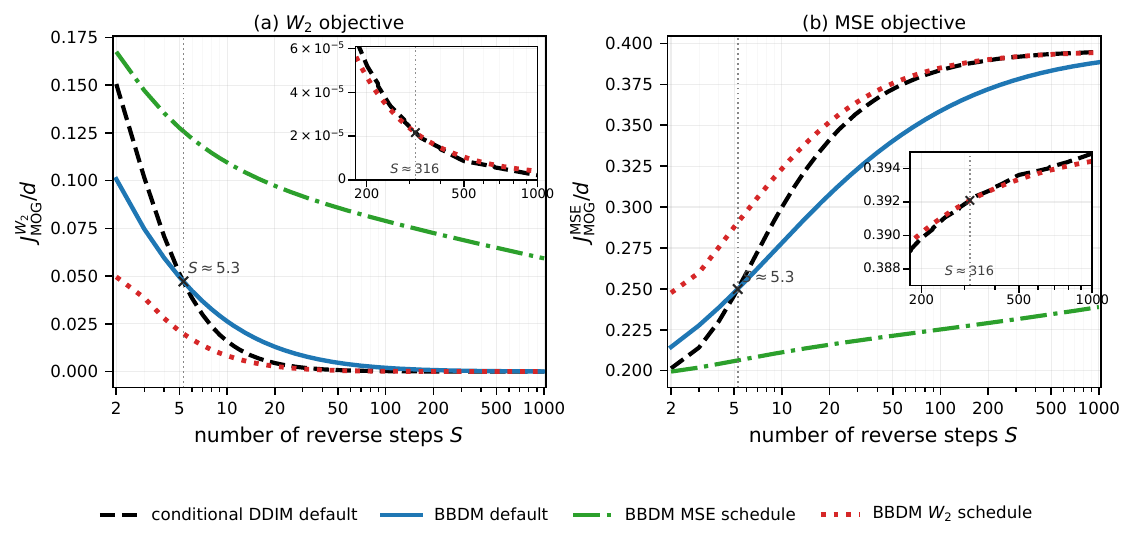}
    \caption{Step-wise analytic comparison between conditional DDIM and BBDM
    schedules in the shared-covariance MoG denoising setting with
    \(\sigma_y=0.5\). Panel (a) reports the normalized matched-component
    Wasserstein objective \(J_{\mathrm{MOG}}^{W_2}/d\), while panel (b) reports
    the normalized matched-label product-coupling MSE
    \(J_{\mathrm{MOG}}^{\mathrm{MSE}}/d\). The default BBDM and DDIM curves
    cross near \(S\simeq5.3\). The \(W_2\)-oriented BBDM schedule gives the
    lowest Wasserstein loss over most of the range, whereas the MSE-oriented
    schedule gives the lowest MSE throughout. }
    \label{fig:mog_denoise_w2_mse_steps_sigma05}
\end{figure}

Figure~\ref{fig:mog_denoise_w2_mse_steps_sigma05} shows the complementary
behavior of the two objectives. Under \(W_2\), the default BBDM schedule has a
small-\(S\) advantage over DDIM, while the \(W_2\)-oriented schedule gives the
lowest loss over most of the sampling range. Under MSE, the ranking reverses:
the MSE-oriented BBDM schedule gives the lowest loss throughout, whereas the
\(W_2\)-oriented schedule incurs a larger MSE because it preserves more
posterior variance. These results agree with the roles of the two objectives:
\(W_2\) favors posterior-spread matching, while MSE favors concentration around
the posterior mean.

\subsection{MNIST validation and transfer to trained BBDM models}

We use MNIST to test whether the selected-label surrogate can guide schedule
design for trained BBDM models. The models address Fourier low-pass
measurements of the form $y=Hx_0+n$, with retained frequency fraction $V$ and
$n\sim\N(0,\sigma_y^2I)$.

The detailed surrogate validation is reported in
Appendix~\ref{app:extra_figures_v43}.
Figure~\ref{fig:mnist_schedule_sweeps_app_v88} evaluates whether the surrogate
reproduces the schedule trends of a trained BBDM, while
Figure~\ref{fig:digit_strip_progression_app_v43} examines how the surrogate
depends on the richness of the fitted MoG prior. Together, these experiments
show that the surrogate provides a sufficiently informative model for guiding
the schedule search.

We then optimize the surrogate MSE objective for five degradation settings and
train corresponding BBDM models on the MNIST training set for 40 epochs using either the default or selected schedule.
The search consistently returns the MSE-oriented schedule $(\alpha,\beta,c,\gamma)\approx(1,2,2,0.2).$

Table~\ref{tab:mnist_default_vs_optimized_main_v75} reports paired
reconstruction metrics: PSNR and structural similarity index measure
(SSIM)~\citep{wang2004ssim}, calculated from piq library, together with an external pretrained MNIST classifier
negative log-likelihood (\texttt{super-j/vit-base-mnist}) that measures whether the reconstructed digit remains
semantically recognizable. We do not use Fr\'echet Inception Distance
(FID)~\citep{heusel2017fid} or Learned Perceptual Image Patch Similarity
(LPIPS)~\citep{zhang2018lpips} in this experiment: FID is primarily a
distribution-level natural-image metric, while LPIPS relies on deep perceptual
features that are not well matched to low-resolution grayscale digit images.
The optimized schedule improves PSNR and NLL in all five settings. SSIM
improves in four settings, and decreases only slightly on the baseline pair.

\begin{table}[htbp]
\caption{MNIST blur: default schedule versus the MSE-optimized schedule.}
\label{tab:mnist_default_vs_optimized_main_v75}

\begin{center}
\setlength{\tabcolsep}{3.2pt}
\begin{tabular*}{\linewidth}{@{\extracolsep{\fill}}cccccc@{}}
\toprule
$V$ & $\sigma_y$ & Sch. &
PSNR$\uparrow$ & SSIM$\uparrow$ & NLL$\downarrow$ \\
\midrule

\multirow{2}{*}{0.10}
& \multirow{2}{*}{0.10}
& def. & 25.360 & \textbf{0.9637} & 0.04250 \\
& & opt. & \textbf{26.335} & 0.9485 & \textbf{0.04128} \\
\midrule

\multirow{2}{*}{0.10}
& \multirow{2}{*}{0.20}
& def. & 23.233 & 0.9270 & 0.05477 \\
& & opt. & \textbf{24.558} & \textbf{0.9385} & \textbf{0.04840} \\
\midrule

\multirow{2}{*}{0.30}
& \multirow{2}{*}{0.10}
& def. & 30.565 & 0.9680 & 0.03150 \\
& & opt. & \textbf{30.818} & \textbf{0.9870} & \textbf{0.03140} \\
\midrule

\multirow{2}{*}{0.05}
& \multirow{2}{*}{0.10}
& def. & 22.928 & 0.9360 & 0.06640 \\
& & opt. & \textbf{24.238} & \textbf{0.9470} & \textbf{0.05940} \\
\midrule

\multirow{2}{*}{1.00}
& \multirow{2}{*}{0.50}
& def. & 23.736 & 0.9460 & 0.04300 \\
& & opt. & \textbf{25.568} & \textbf{0.9640} & \textbf{0.03440} \\

\bottomrule
\end{tabular*}
\end{center}
\end{table}

\FloatBarrier
\subsection{MoG-free heuristics for real-world datasets}
\label{sec:ffhq256_realworld_experiments}

We now evaluate the MoG-free schedule heuristics of
Appendix~\ref{app:mog_free_schedule_heuristics} on FFHQ images at resolution
$256\times256\times3$. Unless stated otherwise, all models are trained for
60 epochs on 54,000 training images, with 7,000 validation images and 7,000
test images, and evaluated with 200 sampling steps. We compare the default
BBDM schedule with the MSE-oriented schedule
$(\alpha,\beta,c,\gamma)=(1,2,2,0.2)$ and the $W_2$-oriented schedule
$(\alpha,\beta,c,\gamma)=(2,1,0.2,2.0)$, using FID, LPIPS, PSNR and SSIM. FID is computed using
\texttt{torch-fidelity} via the \texttt{fidelity} command, while
LPIPS is computed using the \texttt{lpips} package with the learned
AlexNet model.

For reference, we also train architecture-matched
conditional DDIM baseline. This baseline uses the same UNet architecture,
training data, optimizer, training budget, EMA, and evaluation protocol as the
BBDM models. The only algorithmic change is the diffusion path and sampler:
conditional DDIM uses the standard data-to-noise forward process, receives
$y$ only as denoiser conditioning, starts sampling from Gaussian noise, and uses
deterministic DDIM sampling.

We consider three families of linear inverse problems. The blur setting uses a
Fourier low-pass operator with retained frequency fraction $V$. The
super-resolution setting uses
\[
y = U_sD_sx_0+n,\qquad n\sim\N(0,\sigma_y^2I),
\]
where \(D_s\) downsamples by a factor \(s\) in each spatial dimension and
\(U_s\) upsamples the resulting low-resolution image back to the original grid
using nearest-neighbor interpolation. The distributed inpainting setting uses
\[
y=M_px_0+n,\qquad n\sim\N(0,\sigma_y^2I),
\]
where $M_p$ is a fixed binary mask that keeps a fraction $p$ of the RGB pixels.
Qualitative results are shown in Appendix~\ref{app:extra_figures_v43}.

\begin{table}[htbp]
\caption{FFHQ blur: BBDM schedules and conditional DDIM comparison.}
\label{tab:ffhq256_blur_main}

\begin{center}
\setlength{\tabcolsep}{2.7pt}
\begin{tabular*}{\linewidth}{@{\extracolsep{\fill}}ccclccc@{}}
\toprule
$V$ & $\sigma_y$ & Method &
FID$\downarrow$ & LPIPS$\downarrow$ &
PSNR$\uparrow$ & SSIM$\uparrow$ \\
\midrule

\multirow{4}{*}{0.10}
& \multirow{4}{*}{0.10}
& def. & 11.527 & 0.0709 & 30.656 & 0.866 \\
& & MSE & 21.807 & 0.1039 & \textbf{31.792} & \textbf{0.888} \\
& & $W_2$ & \textbf{5.007} & \textbf{0.0562} & 29.996 & 0.839 \\
& & cDDIM & 7.034 & 0.0603 & 29.891 & 0.843 \\
\midrule

\multirow{4}{*}{0.03}
& \multirow{4}{*}{0.10}
& def. & 15.942 & 0.1266 & 27.217 & 0.773 \\
& & MSE & 25.073 & 0.1731 & \textbf{28.547} & \textbf{0.806} \\
& & $W_2$ & \textbf{9.374} & 0.1138 & 27.324 & 0.756 \\
& & cDDIM & 12.400 & \textbf{0.1108} & 26.800 & 0.748 \\
\midrule

\multirow{4}{*}{0.30}
& \multirow{4}{*}{0.10}
& def. & 7.880 & 0.0396 & 33.372 & 0.914 \\
& & MSE & 13.166 & 0.0511 & \textbf{34.871} & \textbf{0.934} \\
& & $W_2$ & \textbf{2.967} & \textbf{0.0278} & 32.468 & 0.893 \\
& & cDDIM & 4.099 & 0.0295 & 32.731 & 0.901 \\
\midrule

\multirow{4}{*}{0.10}
& \multirow{4}{*}{0.20}
& def. & 17.294 & 0.0985 & 29.471 & 0.841 \\
& & MSE & 26.820 & 0.1222 & \textbf{30.815} & \textbf{0.868} \\
& & $W_2$ & \textbf{5.996} & \textbf{0.0754} & 28.958 & 0.811 \\
& & cDDIM & 8.006 & 0.0775 & 28.714 & 0.811 \\

\bottomrule
\end{tabular*}
\end{center}
\end{table}

\begin{table}[htbp]
\caption{FFHQ super-resolution: BBDM schedules and conditional DDIM comparison.}
\label{tab:ffhq256_sr_main}

\begin{center}
\setlength{\tabcolsep}{2.7pt}
\begin{tabular*}{\linewidth}{@{\extracolsep{\fill}}ccclccc@{}}
\toprule
Scale & $\sigma_y$ & Method &
FID$\downarrow$ & LPIPS$\downarrow$ &
PSNR$\uparrow$ & SSIM$\uparrow$ \\
\midrule

\multirow{4}{*}{$4\times$}
& \multirow{4}{*}{0.10}
& def. & 11.720 & 0.0851 & 28.888 & 0.830 \\
& & MSE & 23.997 & 0.1276 & \textbf{30.168} & \textbf{0.860} \\
& & $W_2$ & \textbf{5.066} & \textbf{0.0721} & 28.526 & 0.806 \\
& & cDDIM & 8.538 & 0.0766 & 28.325 & 0.806 \\
\midrule

\multirow{4}{*}{$8\times$}
& \multirow{4}{*}{0.10}
& def. & 19.950 & 0.1623 & 25.018 & 0.716 \\
& & MSE & 34.414 & 0.2049 & \textbf{26.060} & \textbf{0.754} \\
& & $W_2$ & \textbf{12.258} & 0.1533 & 24.837 & 0.696 \\
& & cDDIM & 18.165 & \textbf{0.1510} & 24.624 & 0.688 \\

\bottomrule
\end{tabular*}
\end{center}
\end{table}

\begin{table}[htbp]
\caption{FFHQ distributed inpainting: BBDM schedules and conditional DDIM comparison.}
\label{tab:ffhq256_distributed_inpainting_main}

\begin{center}
\setlength{\tabcolsep}{2.7pt}
\begin{tabular*}{\linewidth}{@{\extracolsep{\fill}}ccclccc@{}}
\toprule
$p$ & $\sigma_y$ & Method &
FID$\downarrow$ & LPIPS$\downarrow$ &
PSNR$\uparrow$ & SSIM$\uparrow$ \\
\midrule

\multirow{4}{*}{0.25}
& \multirow{4}{*}{0.10}
& def. & 14.289 & 0.0794 & 29.802 & 0.863 \\
& & MSE & 19.209 & 0.0855 & \textbf{31.311} & \textbf{0.889} \\
& & $W_2$ & \textbf{5.026} & \textbf{0.0670} & 27.890 & 0.808 \\
& & cDDIM & 9.591 & 0.0711 & 28.969 & 0.836 \\
\midrule

\multirow{4}{*}{0.125}
& \multirow{4}{*}{0.10}
& def. & 15.081 & \textbf{0.1061} & 27.245 & 0.807 \\
& & MSE & 22.808 & 0.1210 & \textbf{28.790} & \textbf{0.844} \\
& & $W_2$ & \textbf{11.099} & 0.1265 & 24.816 & 0.726 \\
& & cDDIM & 15.731 & 0.1089 & 26.592 & 0.777 \\

\bottomrule
\end{tabular*}
\end{center}
\end{table}

The results show that in most settings the conditional DDIM baseline improves substantially over the default
BBDM schedule in FID and LPIPS. However, the $W_2$-oriented BBDM schedule still
achieves the best FID in all FFHQ inverse problems, and usually gives the best
LPIPS as well. 

The distortion metrics follow a complementary pattern. The MSE-oriented BBDM
schedule gives the best PSNR and SSIM in every FFHQ setting, while both the
$W_2$-oriented BBDM schedule and conditional DDIM tend to sacrifice distortion
for more perceptual reconstructions. 

The additional sampling-step ablation in Appendix Table~\ref{tab:ffhq_sampling_steps_appendix}
shows that the same trend between the different BBDM schedules is mostly present also for other amounts of sampling steps.

\FloatBarrier
\section{Conclusion}

This work develops an analytical framework for schedule design in Brownian Bridge Diffusion Models. Under a Mixture-of-Gaussians prior, the selected-label surrogate reduces the effect of the bridge schedule to an explicit covariance-shaping problem. The resulting objectives clarify the distortion--perception tradeoff: MSE-oriented schedules favor concentration near the posterior mean, whereas Wasserstein-oriented schedules favor matching the posterior spread.

 A central direction for future work is to find a way to solve the derived MoG schedule objectives directly over the full valid schedule space for high dimensional datasets. This would replace the bounded parametric search and MoG-free directional heuristics used here with optimization procedures that more faithfully target the actual covariance criteria.

 \section*{Acknowledgements}

This research was partially supported by the Israel Science Foundation (ISF) under Grants 951/24 and 409/24, and by the Council for Higher Education--Planning and Budgeting Committee.

\bibliography{main_tmlr}
\bibliographystyle{tmlr}

\clearpage
\appendix
\section{Generic Gaussian conditioning identities}
\label{app:linear_gaussian_identities}

This appendix records a standard Gaussian conditioning identity that is used repeatedly throughout the paper.

\subsection*{Posterior under an affine Gaussian observation}

Let
\[
x \sim \mathcal N(\mu,\Sigma),
\]
and suppose that the observation $z \in \mathbb R^m$ is generated by
\begin{equation}
z = A x + b + \eta,
\qquad
\eta \sim \mathcal N(0,R),
\qquad
\eta \perp x,
\label{eq:app_linear_obs_model}
\end{equation}
where $A \in \mathbb R^{m\times d}$, $b \in \mathbb R^m$, and $R \succ 0$.

Then the posterior distribution of $x$ given $z$ is Gaussian:
\begin{equation}
x \mid z \sim \mathcal N(\mu_{x|z},\Sigma_{x|z}),
\label{eq:app_posterior_gaussian}
\end{equation}
with
\begin{equation}
\Sigma_{x|z}^{-1}
=
\Sigma^{-1} + A^\top R^{-1}A,
\label{eq:app_post_precision}
\end{equation}
\begin{equation}
\mu_{x|z}
=
\Sigma_{x|z}
\bigl(
\Sigma^{-1}\mu + A^\top R^{-1}(z-b)
\bigr).
\label{eq:app_post_mean_precision}
\end{equation}

\paragraph{Proof.}
By Bayes' rule,
\[
p(x \mid z) \propto p(z \mid x)\,p(x).
\]
Using \eqref{eq:app_linear_obs_model},
\[
p(z \mid x)
\propto
\exp\!\left(
-\frac12 (z-Ax-b)^\top R^{-1}(z-Ax-b)
\right),
\]
and since $x \sim \mathcal N(\mu,\Sigma)$,
\[
p(x)
\propto
\exp\!\left(
-\frac12 (x-\mu)^\top \Sigma^{-1}(x-\mu)
\right).
\]
Therefore,
\[
p(x \mid z)
\propto
\exp\!\left(
-\frac12 (x-\mu)^\top \Sigma^{-1}(x-\mu)
-\frac12 (z-Ax-b)^\top R^{-1}(z-Ax-b)
\right).
\]

Expanding only the terms that depend on $x$, and ignoring the factor of $\frac12$ we obtain
\begin{align*}
& (x-\mu)^\top \Sigma^{-1}(x-\mu)
+ (z-Ax-b)^\top R^{-1}(z-Ax-b) \\
&=
x^\top \Sigma^{-1}x - 2x^\top \Sigma^{-1}\mu
+ x^\top A^\top R^{-1}A x
- 2x^\top A^\top R^{-1}(z-b)
+ \text{const}.
\end{align*}
Grouping the quadratic and linear terms in $x$ gives
\[
x^\top\bigl(\Sigma^{-1}+A^\top R^{-1}A\bigr)x
-
2x^\top\bigl(\Sigma^{-1}\mu + A^\top R^{-1}(z-b)\bigr)
+ \text{const}.
\]
This is the canonical form of a Gaussian density in $x$. Hence the posterior
precision matrix is
\[
\Sigma_{x|z}^{-1}
=
\Sigma^{-1}+A^\top R^{-1}A,
\]
and the posterior mean is
\[
\mu_{x|z}
=
\Sigma_{x|z}
\bigl(\Sigma^{-1}\mu + A^\top R^{-1}(z-b)\bigr).
\]
This proves \eqref{eq:app_post_precision}--\eqref{eq:app_post_mean_precision}.

\clearpage
\section{Full derivation of the exact MoG posterior and the MMSE denoiser}
\label{app:mog_exact_posterior_derivation}

This appendix expands Section~\ref{sec:mog_posterior} step by step. The objective is to compute the posterior distribution
\[
p(x_0 \mid x_s,y)
\]
and its posterior mean, which is the MMSE-optimal estimator of $x_0$.

\subsection*{Model}

We assume that the clean signal $x_0 \in \mathbb R^d$ is drawn from the mixture
\begin{equation}
p(x_0)
=
\sum_{r=1}^{R} \pi_r\,\mathcal N(x_0;\mu_r,\Sigma_r),
\qquad
\pi_r > 0,
\qquad
\sum_{r=1}^{R} \pi_r = 1.
\label{eq:mog_prior}
\end{equation}

The measurement is generated through the linear model
\begin{equation}
y \mid x_0 \sim \mathcal N(Hx_0,\sigma_y^2 I),
\label{eq:mog_measurement}
\end{equation}
and the BBDM forward process at step $s$ is
\begin{equation}
x_s = (1-m_s)x_0 + m_s y + \sqrt{\delta_s}\,\epsilon,
\qquad
\epsilon \sim \mathcal N(0,I).
\label{eq:mog_forward}
\end{equation}

\subsection*{Roadmap of the derivation}

The MoG case contains two sources of uncertainty:

\begin{itemize}
    \item the \emph{continuous} uncertainty of $x_0$ within each Gaussian component.
    \item the \emph{discrete} uncertainty of which component generated the sample.
\end{itemize}

To separate them, we introduce a latent component variable
\[
C \in \{1,\dots,R\},
\]
and write the prior hierarchically as
\begin{equation}
\mathbb P(C=r)=\pi_r,
\qquad
x_0 \mid (C=r) \sim \mathcal N(\mu_r,\Sigma_r).
\label{eq:mog_latent_component}
\end{equation}

Our target is the posterior
\[
p(x_0 \mid x_s,y).
\]
Using the law of total probability over $C$, we decompose it as
\begin{equation}
p(x_0 \mid x_s,y)
=
\sum_{r=1}^{R}
p(x_0 \mid x_s,y,C=r)\,\mathbb P(C=r \mid x_s,y).
\label{eq:mog_posterior_decomposition}
\end{equation}

Thus, it is enough to compute two ingredients:

\begin{enumerate}
    \item the component-conditioned posterior
    \[
    p(x_0 \mid x_s,y,C=r),
    \]

    \item the posterior component probabilities
    \[
    \mathbb P(C=r \mid x_s,y),
    \]
  
\end{enumerate}

The key Gaussian conditioning identity used throughout is recorded in
Appendix~\ref{app:linear_gaussian_identities}.

We proceed in the following order:
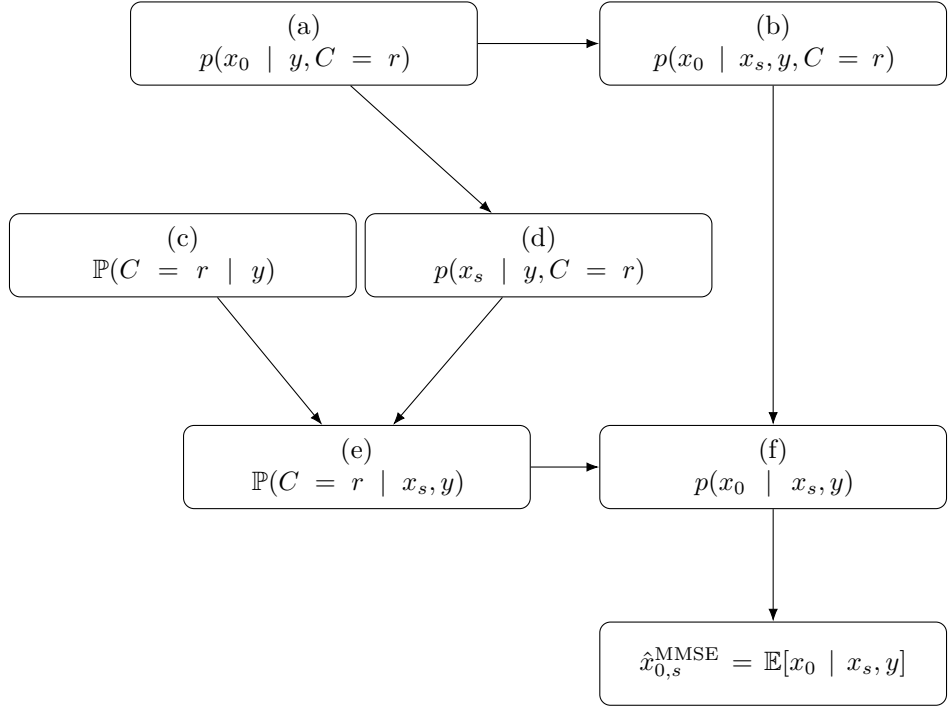
\begin{figure}[ht]
\centering
\begin{tikzpicture}[
  >=Latex,
  box/.style={
    draw,
    rounded corners,
    align=center,
    minimum height=11mm,
    text width=4.3cm,
    inner sep=4pt
  }
]

\node[box] (a) at (0,0) {(a)\\$p(x_0 \mid y,C=r)$};
\node[box] (b) at (6.2,0) {(b)\\$p(x_0 \mid x_s,y,C=r)$};

\node[box] (c) at (-1.6,-2.8) {(c)\\$\mathbb P(C=r \mid y)$};
\node[box] (d) at (3.1,-2.8) {(d)\\$p(x_s \mid y,C=r)$};

\node[box] (e) at (0.7,-5.6) {(e)\\$\mathbb P(C=r \mid x_s,y)$};
\node[box] (post) at (6.2,-5.6) {(f)\\$p(x_0 \mid x_s,y)$};

\node[box] (mmse) at (6.2,-8.2) {$\hat x_{0,s}^{\mathrm{MMSE}}=\mathbb E[x_0\mid x_s,y]$};

\draw[->] (a) -- (b);
\draw[->] (a) -- (d);
\draw[->] (c) -- (e);
\draw[->] (d) -- (e);
\draw[->] (b) -- (post);
\draw[->] (e) -- (post);
\draw[->] (post) -- (mmse);

\end{tikzpicture}
\caption{Dependency graph for the exact MoG posterior derivation. }
\end{figure}

\subsection*{(a) Component-conditioned posterior}

In this subsection, we derive
\[
p(x_0 \mid y,C=r).
\]

Fix a component index $r$. Conditioned on $C=r$, we have
\[
x_0 \mid (C=r) \sim \mathcal N(\mu_r,\Sigma_r),
\]
and the measurement model is
\[
y = Hx_0 + n,
\qquad
n \sim \mathcal N(0,\sigma_y^2 I).
\]

This is an affine Gaussian observation model of the form treated in
Appendix~\ref{app:linear_gaussian_identities}. Applying the generic posterior formula
with
\[
x=x_0,
\qquad
z=y,
\qquad
A=H,
\qquad
b=0,
\qquad
R=\sigma_y^2 I,
\]
we obtain
\[
x_0 \mid (y,C=r) \sim \mathcal N(\mu_{r|y},\Sigma_{r|y}),
\]
where
\begin{equation}
\Sigma_{r|y}^{-1}
=
\Sigma_r^{-1} + \frac{1}{\sigma_y^2} H^\top H,
\label{eq:mog_sigma_r_given_y}
\end{equation}
\begin{equation}
\mu_{r|y}
=
\Sigma_{r|y}
\left(
\Sigma_r^{-1}\mu_r + \frac{1}{\sigma_y^2} H^\top y
\right).
\label{eq:mog_mu_r_given_y}
\end{equation}

Thus, once both $y$ and the component index $r$ are fixed, the posterior over
$x_0$ is Gaussian.

\subsection*{(b) Component-conditioned posterior with $x_s$}

In this subsection, we derive
\[
p(x_0 \mid x_s,y,C=r),
\]
namely the posterior of $x_0$ after observing both $y$ and the bridge variable
$x_s$, while keeping the component index fixed.

From part (a), conditioned on $(y,C=r)$ we have
\[
x_0 \mid (y,C=r) \sim \mathcal N(\mu_{r|y},\Sigma_{r|y}).
\]
The bridge variable satisfies
\[
x_s = (1-m_s)x_0 + m_s y + \sqrt{\delta_s}\,\epsilon,
\qquad
\epsilon \sim \mathcal N(0,I).
\]
Hence, conditioned on $(y,C=r)$, $x_s$ is again an affine Gaussian observation of $x_0$.
Applying the same generic identity from
Appendix~\ref{app:linear_gaussian_identities} with
\[
x=x_0,
\qquad
z=x_s,
\qquad
A=(1-m_s)I,
\qquad
b=m_s y,
\qquad
R=\delta_s I,
\]
and prior $x_0 \mid (y,C=r)\sim\mathcal N(\mu_{r|y},\Sigma_{r|y})$, we obtain
\[
x_0 \mid (x_s,y,C=r)
\sim
\mathcal N\bigl(\mu_{r|s}(x_s,y),\Sigma_{r|s}\bigr),
\]
where
\begin{equation}
\Sigma_{r|s}
=
\left(
\Sigma_{r|y}^{-1}
+
\frac{(1-m_s)^2}{\delta_s} I
\right)^{-1},
\label{eq:mog_sigma_r_given_xs_y}
\end{equation}

\begin{equation}
\mu_{r|s}(x_s,y)
=
\Sigma_{r|s}
\left(
\Sigma_{r|y}^{-1}\mu_{r|y}
+
\frac{1-m_s}{\delta_s}(x_s-m_s y)
\right).
\label{eq:mog_mu_r_given_xs_y_compact}
\end{equation}

Using
\[
\Sigma_{r|y}^{-1}
=
\Sigma_r^{-1}+\frac{1}{\sigma_y^2}H^\top H,
\qquad
\Sigma_{r|y}^{-1}\mu_{r|y}
=
\Sigma_r^{-1}\mu_r+\frac{1}{\sigma_y^2}H^\top y,
\]
we equivalently obtain
\begin{equation}
\Sigma_{r|s}
=
\left(
\Sigma_r^{-1}
+
\frac{1}{\sigma_y^2}H^\top H
+
\frac{(1-m_s)^2}{\delta_s}I
\right)^{-1},
\label{eq:mog_sigma_r_given_xs_y_expanded}
\end{equation}

\begin{equation}
\mu_{r|s}(x_s,y)
=
\Sigma_{r|s}
\left(
\Sigma_r^{-1}\mu_r
+
\frac{1}{\sigma_y^2}H^\top y
+
\frac{1-m_s}{\delta_s}(x_s-m_s y)
\right).
\label{eq:mog_mu_r_given_xs_y}
\end{equation}

Thus, for every fixed component $r$, the posterior of $x_0$ given $(x_s,y,C=r)$
is Gaussian with explicitly known mean and covariance.

\subsection*{(c) Posterior component probabilities}

In this subsection, we derive
\[
\gamma_{r|y}(y)
\triangleq
\mathbb P(C=r \mid y),
\]

the posterior probability that component $r$ generated the sample after observing
the measurement $y$.

By Bayes' rule,
\begin{equation}
\mathbb P(C=r \mid y)
=
\frac{\mathbb P(C=r)\,p(y \mid C=r)}{p(y)}.
\label{eq:mog_bayes_cr_y}
\end{equation}
We already know that $\mathbb P(C=r)=\pi_r$, so it remains to compute $p(y \mid C=r)$.

Conditioned on $C=r$, we have
\[
x_0 \sim \mathcal N(\mu_r,\Sigma_r),
\qquad
y = Hx_0 + n,
\qquad
n \sim \mathcal N(0,\sigma_y^2 I).
\]
By an affine-Gaussian marginal identity, it follows that
\begin{equation}
p(y \mid C=r)
=
\mathcal N\bigl(y;\,H\mu_r,\;H\Sigma_r H^\top + \sigma_y^2 I\bigr).
\label{eq:mog_y_given_cr}
\end{equation}

Substituting \eqref{eq:mog_y_given_cr} into \eqref{eq:mog_bayes_cr_y}, we obtain
\begin{equation}
\gamma_{r|y}(y)
\triangleq
\mathbb P(C=r \mid y)
=
\frac{
\pi_r\,\mathcal N\!\bigl(y;\,H\mu_r,\;H\Sigma_r H^\top + \sigma_y^2 I\bigr)
}{
\sum_{j=1}^{R}
\pi_j\,\mathcal N\!\bigl(y;\,H\mu_j,\;H\Sigma_j H^\top + \sigma_y^2 I\bigr)
},
\qquad
\sum_{r=1}^{R} \gamma_{r|y}(y) = 1.
\label{eq:mog_gamma_given_y}
\end{equation}

\subsection*{(d) Marginal likelihood}

In this subsection, we derive
\[
p(x_s \mid y,C=r),
\]
the conditional marginal likelihood of the bridge variable $x_s$ after observing
$y$ and fixing the component index $r$.

From part (a), conditioned on $(y,C=r)$ we have
\[
x_0 \sim \mathcal N(\mu_{r|y},\Sigma_{r|y}),
\]
and the forward process is
\[
x_s = (1-m_s)x_0 + m_s y + \sqrt{\delta_s}\epsilon,
\qquad
\epsilon \sim \mathcal N(0,I).
\]
Again using an affine-Gaussian marginal identity, we obtain
\begin{equation}
p(x_s \mid y,C=r)
=
\mathcal N\bigl(
x_s;\,
(1-m_s)\mu_{r|y} + m_s y,\,
(1-m_s)^2 \Sigma_{r|y} + \delta_s I
\bigr).
\label{eq:mog_xs_given_y_cr}
\end{equation}
\subsection*{(e) Posterior component probabilities}

In this subsection, we derive
\[
\gamma_{r|s}(x_s,y)
\triangleq
\mathbb P(C=r \mid x_s,y),
\]
the posterior probability that component $r$ generated the sample after observing
both $y$ and $x_s$.

By Bayes' rule,
\begin{equation}
\mathbb P(C=r \mid x_s,y)
=
\frac{\mathbb P(C=r,x_s \mid y)}{p(x_s \mid y)}.
\label{eq:mog_bayes_cr_xs_y}
\end{equation}
Using the product rule on the numerator,
\[
\mathbb P(C=r,x_s \mid y)
=
\mathbb P(C=r \mid y)\,p(x_s \mid y,C=r).
\]
Therefore,
\begin{equation}
\mathbb P(C=r \mid x_s,y)
=
\frac{
\mathbb P(C=r \mid y)\,p(x_s \mid y,C=r)
}{
\sum_{j=1}^{R}
\mathbb P(C=j \mid y)\,p(x_s \mid y,C=j)
}.
\label{eq:mog_gamma_cr_xs_y_intermediate}
\end{equation}

Substituting the expressions from parts (c) and (d), we obtain
\begin{equation}
\gamma_{r|s}(x_s,y)
=
\frac{
\gamma_{r|y}(y)\,
\mathcal N\!\bigl(
x_s;\,
(1-m_s)\mu_{r|y}+m_s y,\,
(1-m_s)^2\Sigma_{r|y}+\delta_s I
\bigr)
}{
\sum_{j=1}^{R}
\gamma_{j|y}(y)\,
\mathcal N\!\bigl(
x_s;\,
(1-m_s)\mu_{j|y}+m_s y,\,
(1-m_s)^2\Sigma_{j|y}+\delta_s I
\bigr)
},
\qquad
\sum_{r=1}^{R}\gamma_{r|s}(x_s,y)=1.
\label{eq:mog_gamma_given_xs_y}
\end{equation}

Thus, the role of $x_s$ is to refine the measurement-only probabilities
$\gamma_{r|y}(y)$ into the posterior responsibilities $\gamma_{r|s}(x_s,y)$.

\subsection*{(f) Exact posterior and MMSE-optimal denoiser}

We now combine the component-conditioned posteriors from part (b)
with the posterior responsibilities from part (e) to obtain the full posterior
$p(x_0 \mid x_s,y)$ and its posterior mean. The formula for the full posterior will be:
\begin{equation}
p(x_0 \mid x_s,y)
=
\sum_{r=1}^{R}
\gamma_{r|s}(x_s,y)\,
\mathcal N\bigl(
x_0;\mu_{r|s}(x_s,y),\Sigma_{r|s}
\bigr).
\label{eq:mog_exact_posterior}
\end{equation}

Thus, the exact posterior remains a Mixture-of-Gaussians, with component means
$\mu_{r|s}(x_s,y)$, component covariances $\Sigma_{r|s}$, and weights
$\gamma_{r|s}(x_s,y)$.

Finally, since the Bayes-optimal denoiser under squared loss is the posterior mean,
we obtain
\begin{equation}
\hat x_{0,s}^{\mathrm{MMSE}}(x_s,y)
=
\mathbb E[x_0 \mid x_s,y]
=
\sum_{r=1}^{R}
\gamma_{r|s}(x_s,y)\,\mu_{r|s}(x_s,y).
\label{eq:mog_exact_mmse}
\end{equation}

\clearpage

\section{Full derivation of the oracle-induced reverse update}
\label{app:exact_reverse_expansion_v6}

This appendix expands the algebra behind Section~\ref{sec:mog_reverse_time}. The goal is to start from the exact posterior mean derived in Appendix~\ref{app:mog_exact_posterior_derivation}, substitute it into the BBDM sampler, and collect the resulting coefficients of $x_s$, $y$, and the constant term.

Using the notation of Section~\ref{sec:preliminaries}, the BBDM reverse update for \(1\le s\le S-1\) is
\begin{equation}
x_{s-1}
=
a_s \hat x_0 + b_s y + c_s x_s + \sigma_s z_s,
\qquad
z_s \sim \mathcal N(0,I),
\label{eq:mog_reverse_generic}
\end{equation}
where
\begin{equation}
a_s=(1-m_{s-1})-(1-m_s)\sqrt{\frac{\delta_{s-1}-\sigma_s^2}{\delta_s}},
\qquad
b_s=m_{s-1}-m_s\sqrt{\frac{\delta_{s-1}-\sigma_s^2}{\delta_s}},
\qquad
c_s=\sqrt{\frac{\delta_{s-1}-\sigma_s^2}{\delta_s}}.
\label{eq:mog_abc}
\end{equation}

From Appendix~\ref{app:mog_exact_posterior_derivation}, the exact MMSE denoiser is
\begin{equation}
\hat x_{0,s}^{\mathrm{MMSE}}(x_s,y)
=
\sum_{r=1}^{R} \gamma_{r|s}(x_s,y)\,\mu_{r|s}(x_s,y),
\label{eq:mog_mmse_repeat}
\end{equation}
with component-conditioned posterior mean
\begin{equation}
\mu_{r|s}(x_s,y)
=
\Sigma_{r|s}
\left(
\Sigma_r^{-1}\mu_r
+
\frac{1}{\sigma_y^2}H^\top y
+
\frac{1-m_s}{\delta_s}(x_s-m_s y)
\right).
\label{eq:mog_mu_r_given_xs_y_repeat}
\end{equation}

Substituting \eqref{eq:mog_mmse_repeat} into \eqref{eq:mog_reverse_generic} gives
\begin{equation}
x_{s-1}
=
a_s \sum_{r=1}^{R}\gamma_{r|s}(x_s,y)\,\mu_{r|s}(x_s,y)
+ b_s y + c_s x_s + \sigma_s z_s.
\label{eq:mog_reverse_after_mmse}
\end{equation}
Substituting \eqref{eq:mog_mu_r_given_xs_y_repeat} into \eqref{eq:mog_reverse_after_mmse} yields
\begin{align}
x_{s-1}
&=
a_s \sum_{r=1}^{R}\gamma_{r|s}(x_s,y)
\Sigma_{r|s}
\left(
\Sigma_r^{-1}\mu_r
+
\frac{1}{\sigma_y^2}H^\top y
+
\frac{1-m_s}{\delta_s}(x_s-m_s y)
\right)
+ b_s y + c_s x_s + \sigma_s z_s
\notag\\
&=
a_s \sum_{r=1}^{R}\gamma_{r|s}(x_s,y)\,\Sigma_{r|s}\Sigma_r^{-1}\mu_r
+
\frac{a_s}{\sigma_y^2}
\sum_{r=1}^{R}\gamma_{r|s}(x_s,y)\,\Sigma_{r|s}H^\top y
\notag\\
&\quad+
a_s\frac{1-m_s}{\delta_s}
\sum_{r=1}^{R}\gamma_{r|s}(x_s,y)\,\Sigma_{r|s}(x_s-m_s y)
+ b_s y + c_s x_s + \sigma_s z_s.
\label{eq:mog_reverse_expand_step1}
\end{align}
Next expand the term containing $(x_s-m_s y)$:
\begin{align}
x_{s-1}
&=
a_s \sum_{r=1}^{R}\gamma_{r|s}(x_s,y)\,\Sigma_{r|s}\Sigma_r^{-1}\mu_r
+
\frac{a_s}{\sigma_y^2}
\sum_{r=1}^{R}\gamma_{r|s}(x_s,y)\,\Sigma_{r|s}H^\top y
\notag\\
&\quad+
a_s\frac{1-m_s}{\delta_s}
\sum_{r=1}^{R}\gamma_{r|s}(x_s,y)\,\Sigma_{r|s}x_s
-
a_s\frac{m_s(1-m_s)}{\delta_s}
\sum_{r=1}^{R}\gamma_{r|s}(x_s,y)\,\Sigma_{r|s}y
\notag\\
&\quad+
b_s y + c_s x_s + \sigma_s z_s.
\label{eq:mog_reverse_expand_step2}
\end{align}
Finally collect the coefficients of $x_s$ and $y$:
\begin{equation}
\begin{aligned}
x_{s-1}
&=
\left[
c_s I
+
a_s\frac{1-m_s}{\delta_s}
\sum_{r=1}^{R}\gamma_{r|s}(x_s,y)\,\Sigma_{r|s}
\right]x_s
\\[3pt]
&\quad+
\left[
b_s I
+
\frac{a_s}{\sigma_y^2}
\sum_{r=1}^{R}\gamma_{r|s}(x_s,y)\,\Sigma_{r|s}H^\top
-
a_s\frac{m_s(1-m_s)}{\delta_s}
\sum_{r=1}^{R}\gamma_{r|s}(x_s,y)\,\Sigma_{r|s}
\right]y
\\[3pt]
&\quad+
a_s \sum_{r=1}^{R}\gamma_{r|s}(x_s,y)\,\Sigma_{r|s}\Sigma_r^{-1}\mu_r
+
\sigma_s z_s,
\qquad
z_s \sim \mathcal N(0,I).
\end{aligned}
\label{eq:mog_reverse_exact}
\end{equation}

\newpage

\section{Selected-label approximation, problem-adapted basis, and the induced Gaussian law}
\label{app:selected_label_formulas_v6}

This appendix derives the fixed-label reverse law used in Section~\ref{sec:mog_selection_basis}. It begins from the frozen-label BBDM update, moves to the component-wise basis in which the dynamics become diagonal, and writes the explicit Gaussian law induced by the selected-label approximation. It ends with proving the mean exactness and covariance deficit claims in ~\ref{prop:selected_label_mean_exactness} and ~\ref{prop:selected_label_covariance_deficit}.

\subsection*{D.1 Fixed-label reverse update}

The selected-label approximation draws an auxiliary label $J\in\{1,\dots,R\}$ from the measurement posterior,
\begin{equation}
\mathbb P(J=r\mid y)=\gamma_{r|y}(y),
\label{eq:mog_gamma_given_y_problem_basis}
\end{equation}
and then keeps that label fixed along the reverse chain.

Condition on $J=r$. The approximation uses the component-conditioned posterior mean
\begin{equation}
\mu_{r|s}(x_s,y)
=
\Sigma_{r|s}
\left(
\Sigma_{r|y}^{-1}\mu_{r|y}
+
\frac{1-m_s}{\delta_s}(x_s-m_s y)
\right),
\label{eq:mog_mu_rs_problem_basis_pre}
\end{equation}
with
\begin{equation}
\Sigma_{r|s}
=
\left(
\Sigma_{r|y}^{-1}
+
\frac{(1-m_s)^2}{\delta_s}I
\right)^{-1}.
\label{eq:mog_sigma_rs_problem_basis_pre}
\end{equation}
At \(s=S\), we use \(\hat x_0=\mu_{r|y}\). The formulas below are
for \(1\le s\le S-1\):

Insert $\hat x_0=\mu_{r|s}(x_s,y)$ into the BBDM update
\begin{equation}
x_{s-1}
=
a_s \hat x_0 + b_s y + c_s x_s + \sigma_s z_s,
\qquad
z_s\sim\mathcal N(0,I).
\label{eq:mog_reverse_generic_problem_basis}
\end{equation}
This gives
\begin{equation}
\begin{aligned}
x_{s-1}
&=
a_s\Sigma_{r|s}
\left(
\Sigma_{r|y}^{-1}\mu_{r|y}
+
\frac{1-m_s}{\delta_s}(x_s-m_s y)
\right)
+b_s y+c_s x_s+\sigma_s z_s
\\
&=
a_s\Sigma_{r|s}\Sigma_{r|y}^{-1}\mu_{r|y}
+
a_s\frac{1-m_s}{\delta_s}\Sigma_{r|s}x_s
-
a_s\frac{m_s(1-m_s)}{\delta_s}\Sigma_{r|s}y
+b_s y+c_s x_s+\sigma_s z_s.
\end{aligned}
\label{eq:mog_fixed_label_expand_step1}
\end{equation}
Collecting the $x_s$ and $y$ terms yields
\begin{equation}
\begin{aligned}
x_{s-1}
&=
\left(c_s I + a_s\frac{1-m_s}{\delta_s}\Sigma_{r|s}\right)x_s
\\
&\quad+
\left(b_s I - a_s\frac{m_s(1-m_s)}{\delta_s}\Sigma_{r|s}\right)y
\\
&\quad+
a_s\Sigma_{r|s}\Sigma_{r|y}^{-1}\mu_{r|y}
+\sigma_s z_s,
\qquad z_s\sim\mathcal N(0,I).
\end{aligned}
\label{eq:mog_fixed_label_reverse_update}
\end{equation}
Thus, once the label is frozen, the reverse update is affine in $x_s$ and $y$.

\subsection*{D.2 Component-wise problem-adapted basis}

Fix a component $r$. Since $\Sigma_{r|y}^{-1}$ is symmetric positive definite, there exist an orthogonal matrix $U_r$ and positive scalars $\lambda_{r,1},\dots,\lambda_{r,d}$ such that
\begin{equation}
\Sigma_{r|y}^{-1}
=
U_r\,\operatorname{diag}(\lambda_{r,1},\dots,\lambda_{r,d})\,U_r^\top.
\label{eq:mog_precision_eig_r}
\end{equation}
Define the projected variables
\begin{equation}
\begin{aligned}
x_s^{U_r}&\triangleq U_r^\top x_s,
\qquad y^{U_r}\triangleq U_r^\top y,
\\
\mu_{r|y}^{U_r}&\triangleq U_r^\top\mu_{r|y},
\qquad z_s^{U_r}\triangleq U_r^\top z_s.
\end{aligned}
\label{eq:projected_variables_appendix_v81}
\end{equation}
Because $U_r$ is orthogonal and $z_s\sim\mathcal N(0,I)$, we also have $z_s^{U_r}\sim\mathcal N(0,I)$.

For each step \(1\le s\le S-1\), define
\begin{equation}
\Lambda_{r\mid s}
\triangleq
\operatorname{diag}\!\left(
\frac{1}{\lambda_{r,1}+\frac{(1-m_s)^2}{\delta_s}},
\dots,
\frac{1}{\lambda_{r,d}+\frac{(1-m_s)^2}{\delta_s}}
\right).
\label{eq:mog_Lambda_problem_basis}
\end{equation}
Then
\begin{equation}
\Sigma_{r|s}
=
\left(
\Sigma_{r|y}^{-1}+\frac{(1-m_s)^2}{\delta_s}I
\right)^{-1}
=
U_r\Lambda_{r\mid s}U_r^\top.
\label{eq:mog_sigma_rs_problem_basis}
\end{equation}
Hence every $\Sigma_{r|s}$ is diagonal in the same basis $U_r$.

\subsection*{D.3 Reverse recursion in the basis $U_r$}

Project \eqref{eq:mog_fixed_label_reverse_update} onto $U_r$:
\begin{equation}
\begin{aligned}
x_{s-1}^{U_r}
&=
\left(c_s I + a_s\frac{1-m_s}{\delta_s}\Lambda_{r\mid s}\right)x_s^{U_r}
\\
&\quad+
\left(b_s I - a_s\frac{m_s(1-m_s)}{\delta_s}\Lambda_{r\mid s}\right)y^{U_r}
\\
&\quad+
a_s\Lambda_{r\mid s}\operatorname{diag}(\lambda_{r,1},\dots,\lambda_{r,d})\mu_{r|y}^{U_r}
+\sigma_s z_s^{U_r}.
\end{aligned}
\label{eq:mog_reverse_problem_basis_step1}
\end{equation}
For $1\le s\le S-1$, define the diagonal matrices
\begin{align}
G_r(s)
&\triangleq
c_s I+\frac{a_s(1-m_s)}{\delta_s}\Lambda_{r\mid s},
\label{eq:mog_Gr_problem_basis}
\\[2pt]
N_r(s)
&\triangleq
b_s I-\frac{a_s m_s(1-m_s)}{\delta_s}\Lambda_{r\mid s},
\label{eq:mog_Nr_problem_basis}
\\[2pt]
M_r(s)
&\triangleq
a_s\Lambda_{r\mid s}\operatorname{diag}(\lambda_{r,1},\dots,\lambda_{r,d}).
\label{eq:mog_Mr_problem_basis}
\end{align}

Then
\begin{equation}
x_{s-1}^{U_r}
=
G_r(s)x_s^{U_r}+N_r(s)y^{U_r}+M_r(s)\mu_{r|y}^{U_r}+\sigma_s z_s^{U_r}.
\label{eq:mog_reverse_problem_basis_vector}
\end{equation}

Since $G_r(s)$, $N_r(s)$, and $M_r(s)$ are diagonal, the recursion decouples coordinate-wise. For coordinate $k$,
\begin{equation}
x_{s-1,k}^{U_r}
=
g_{r,k}(s)x_{s,k}^{U_r}+n_{r,k}(s)y_k^{U_r}+m_{r,k}(s)\mu_{r|y,k}^{U_r}+\sigma_s z_{s,k}^{U_r},
\label{eq:mog_reverse_problem_basis_scalar}
\end{equation}
where $g_{r,k}(s)$, $n_{r,k}(s)$, and $m_{r,k}(s)$ are the diagonal entries of $G_r(s)$, $N_r(s)$, and $M_r(s)$. For \(s=S\), set
\[
G_r(S)=0,\qquad
N_r(S)=m_{S-1}I,\qquad
M_r(S)=(1-m_{S-1})I,\qquad
\sigma_S^2=\delta_{S-1}.
\]
Then \eqref{eq:mog_reverse_problem_basis_vector} holds for all \(s=1,\ldots,S\).

\subsection*{D.4 Unrolling the fixed-label recursion}

Repeated substitution of \eqref{eq:mog_reverse_problem_basis_vector} using the endpoint convention at
\(s=S\), yields, for any step $\ell\in\{0,\dots,S-1\}$,
\begin{align}
 x_\ell^{U_r}
&=
\left[\prod_{s'=\ell+1}^{S}G_r(s')\right]x_S^{U_r}
+
\sum_{i=\ell+1}^{S}
\left(\prod_{j=\ell+1}^{i-1}G_r(j)\right)N_r(i)\,y^{U_r}
\notag\\
&\quad+
\sum_{i=\ell+1}^{S}
\left(\prod_{j=\ell+1}^{i-1}G_r(j)\right)M_r(i)\,\mu_{r|y}^{U_r}
+
\sum_{i=\ell+1}^{S}
\left(\prod_{j=\ell+1}^{i-1}G_r(j)\right)\sigma_i z_i^{U_r}.
\label{eq:mog_unrolled_problem_basis_general}
\end{align}
Here the empty product is understood as the identity. In BBDM inference, the terminal latent is initialized from the conditioning signal, so $x_S=y$ and therefore $x_S^{U_r}=y^{U_r}$. Setting $\ell=0$ in \eqref{eq:mog_unrolled_problem_basis_general} gives
\begin{equation}
\hat x_0^{U_r,(r)}
=
D_1^{(r)}y^{U_r}
+
D_2^{(r)}\mu_{r|y}^{U_r}
+
\sum_{i=1}^{S}
\left(\prod_{j=1}^{i-1}G_r(j)\right)\sigma_i z_i^{U_r},
\label{eq:mog_x0hat_problem_basis}
\end{equation}
where
\begin{equation}
D_1^{(r)}
\triangleq
\prod_{s'=1}^{S}G_r(s')
+
\sum_{i=1}^{S}
\left(\prod_{j=1}^{i-1}G_r(j)\right)N_r(i),
\label{eq:mog_D1_problem_basis}
\end{equation}
and
\begin{equation}
D_2^{(r)}
\triangleq
\sum_{i=1}^{S}
\left(\prod_{j=1}^{i-1}G_r(j)\right)M_r(i).
\label{eq:mog_D2_problem_basis}
\end{equation}
Because $G_r(s)$, $N_r(s)$, and $M_r(s)$ are diagonal, both $D_1^{(r)}$ and $D_2^{(r)}$ are diagonal as well.

\subsection*{D.5 The selected-label Gaussian law}

Conditioned on $(y,J=r)$, the only randomness in \eqref{eq:mog_x0hat_problem_basis} comes from the reverse Gaussian innovations. Hence
\begin{equation}
p_{\mathrm{Select}}(\hat x_0^{U_r,(r)}\mid y,J=r)
=
\mathcal N\!\bigl(
\mu_{\mathrm{BBDM}}^{U_r,(r)}(y),
\Sigma_{\mathrm{BBDM}}^{U_r,(r)}
\bigr),
\label{eq:mog_BBDM_conditional_problem_basis}
\end{equation}
with mean
\begin{equation}
\mu_{\mathrm{BBDM}}^{U_r,(r)}(y)
=
D_1^{(r)}y^{U_r}+D_2^{(r)}\mu_{r|y}^{U_r},
\label{eq:mog_BBDM_mean_problem_basis}
\end{equation}
and covariance
\begin{equation}
\Sigma_{\mathrm{BBDM}}^{U_r,(r)}
=
\sum_{i=1}^{S}
\left(\prod_{j=1}^{i-1}G_r(j)\right)\sigma_i^2 I\left(\prod_{j=1}^{i-1}G_r(j)\right)^\top.
\label{eq:mog_BBDM_cov_problem_basis}
\end{equation}
Returning to the original coordinates,
\begin{equation}
p_{\mathrm{Select}}(\hat x_0\mid y,J=r)
=
\mathcal N\!\bigl(
\hat x_0;
\mu_{\mathrm{BBDM}}^{(r)}(y),
\Sigma_{\mathrm{BBDM}}^{(r)}
\bigr),
\label{eq:mog_BBDM_conditional_original_coords}
\end{equation}
where
\begin{equation}
\mu_{\mathrm{BBDM}}^{(r)}(y)
=
U_r\mu_{\mathrm{BBDM}}^{U_r,(r)}(y),
\qquad
\Sigma_{\mathrm{BBDM}}^{(r)}
=
U_r\Sigma_{\mathrm{BBDM}}^{U_r,(r)}U_r^\top.
\label{eq:mog_BBDM_mean_cov_original_coords}
\end{equation}
Finally, averaging over the selected label gives
\begin{equation}
p_{\mathrm{Select}}(\hat x_0\mid y)
=
\sum_{r=1}^{R}
\gamma_{r|y}(y)
\mathcal N\!\bigl(
\hat x_0;
\mu_{\mathrm{BBDM}}^{(r)}(y),
\Sigma_{\mathrm{BBDM}}^{(r)}
\bigr).
\label{eq:mog_selection_mixture_problem_basis}
\end{equation}

\subsection*{D.6 Mean exactness and covariance deficit}

We now prove the two structural claims used in the main text. The proofs use only the fixed-label recursion derived above.

\paragraph{Proof of Proposition~\ref{prop:selected_label_mean_exactness}.}
Condition on $J=r$ and work in the basis $U_r$. Since the Gaussian innovations have zero mean, the conditional mean of the selected-label chain obeys the deterministic recursion
\begin{equation}
\bar x_{s-1}^{U_r}
=
G_r(s)\bar x_s^{U_r}
+
N_r(s)y^{U_r}
+
M_r(s)\mu_{r|y}^{U_r},
\label{eq:app_mean_recursion_vector}
\end{equation}
where
\[
\bar x_s^{U_r}\triangleq \mathbb E[x_s^{U_r}\mid y,J=r].
\]

We prove by backward induction that, for every step $s$,
\begin{equation}
\bar x_s^{U_r}
=
m_s y^{U_r}+(1-m_s)\mu_{r|y}^{U_r}.
\label{eq:app_bridge_mean_vector_induction}
\end{equation}
The initialization holds at the terminal step because $x_S=y$ and $m_S\cong1$, hence
\[
\bar x_S^{U_r}
=
y^{U_r}
\cong
m_S y^{U_r}+(1-m_S)\mu_{r|y}^{U_r}.
\]

Now assume that \eqref{eq:app_bridge_mean_vector_induction} holds at step $s$. Substituting this induction hypothesis into \eqref{eq:app_mean_recursion_vector} gives
\begin{equation}
\begin{aligned}
\bar x_{s-1}^{U_r}
&=
G_r(s)\bigl(m_s y^{U_r}+(1-m_s)\mu_{r|y}^{U_r}\bigr)
+
N_r(s)y^{U_r}
+
M_r(s)\mu_{r|y}^{U_r}
\\
&=
\bigl(G_r(s)m_s+N_r(s)\bigr)y^{U_r}
+
\bigl(G_r(s)(1-m_s)+M_r(s)\bigr)\mu_{r|y}^{U_r}.
\end{aligned}
\label{eq:app_mean_recursion_before_cancellation}
\end{equation}

For the endpoint \(s=S\), the identities
\[
G_r(s)m_s+N_r(s)=m_{s-1}I,\qquad
G_r(s)(1-m_s)+M_r(s)=(1-m_{s-1})I
\]
follow immediately from the endpoint convention. For
\(1\le s\le S-1\), they follow from the algebra below. Using,
\[
G_r(s)=c_sI+\frac{a_s(1-m_s)}{\delta_s}\Lambda_{r|s},
\qquad
N_r(s)=b_sI-\frac{a_sm_s(1-m_s)}{\delta_s}\Lambda_{r|s},
\]
we get
\begin{equation}
\begin{aligned}
G_r(s)m_s+N_r(s)
&=
\left(c_sI+\frac{a_s(1-m_s)}{\delta_s}\Lambda_{r|s}\right)m_s
+
b_sI-\frac{a_sm_s(1-m_s)}{\delta_s}\Lambda_{r|s}
\\
&=
(c_sm_s+b_s)I
+
\frac{a_sm_s(1-m_s)}{\delta_s}\Lambda_{r|s}
-
\frac{a_sm_s(1-m_s)}{\delta_s}\Lambda_{r|s}
\\
&=
(c_sm_s+b_s)I
\\
&=
m_{s-1}I.
\end{aligned}
\label{eq:app_y_coeff_cancellation}
\end{equation}
The last equality follows from the definitions
\[
b_s=m_{s-1}-m_sc_s,
\qquad
c_s=\sqrt{\frac{\delta_{s-1}-\sigma_s^2}{\delta_s}}.
\]

Second, using
\[
M_r(s)=a_s\Lambda_{r|s}\diag(\lambda_{r,1},\dots,\lambda_{r,d}),
\]
we have
\begin{equation}
\begin{aligned}
G_r(s)(1-m_s)+M_r(s)
&=
\left(c_sI+\frac{a_s(1-m_s)}{\delta_s}\Lambda_{r|s}\right)(1-m_s)
+
a_s\Lambda_{r|s}\diag(\lambda_{r,1},\dots,\lambda_{r,d})
\\
&=
c_s(1-m_s)I
+
a_s\Lambda_{r|s}
\left(
\frac{(1-m_s)^2}{\delta_s}I
+
\diag(\lambda_{r,1},\dots,\lambda_{r,d})
\right).
\end{aligned}
\label{eq:app_mu_coeff_cancellation_step1}
\end{equation}
By definition,
\[
\Lambda_{r|s}
=
\left(
\diag(\lambda_{r,1},\dots,\lambda_{r,d})
+
\frac{(1-m_s)^2}{\delta_s}I
\right)^{-1},
\]
and therefore
\[
\Lambda_{r|s}
\left(
\frac{(1-m_s)^2}{\delta_s}I
+
\diag(\lambda_{r,1},\dots,\lambda_{r,d})
\right)
=
I.
\]
Substituting this into \eqref{eq:app_mu_coeff_cancellation_step1} gives
\begin{equation}
\begin{aligned}
G_r(s)(1-m_s)+M_r(s)
&=
c_s(1-m_s)I+a_sI
&=
\bigl(c_s(1-m_s)+a_s\bigr)I
&=
(1-m_{s-1})I.
\end{aligned}
\label{eq:app_mu_coeff_cancellation}
\end{equation}
The last equality follows from the definition
\[
a_s=(1-m_{s-1})-(1-m_s)c_s.
\]

Combining \eqref{eq:app_mean_recursion_before_cancellation}, \eqref{eq:app_y_coeff_cancellation}, and \eqref{eq:app_mu_coeff_cancellation}, we obtain
\[
\bar x_{s-1}^{U_r}
=
m_{s-1}y^{U_r}
+
(1-m_{s-1})\mu_{r|y}^{U_r}.
\]
Thus the induction is complete. At $s=0$, since $m_0=0$,
\[
\mathbb E[\hat x_0^{U_r,(r)}\mid y,J=r]
=
\mu_{r|y}^{U_r}.
\]
Returning to the original coordinates gives
\[
\mathbb E[\hat x_0\mid y,J=r]
=
\mu_{r|y}.
\]
Comparing this final mean with the deterministic part of \eqref{eq:mog_x0hat_problem_basis} gives
\[
D_1^{(r)}=0,
\qquad
D_2^{(r)}=I.
\]

\paragraph{Proof of Proposition~\ref{prop:selected_label_covariance_deficit}.}
Condition on component \(r\) and work in the basis \(U_r\). We compare two
chains that both start from the same deterministic endpoint \(x_S=y\).
The first chain is the selected-label chain conditioned on \((y,J=r)\). This is
the chain that uses the posterior mean \(\mu_{r|s}(x_s,y)\) inside the BBDM
reverse update.
The second chain is the exact component-conditioned bridge reverse chain
conditioned on \((y,C=r)\). At each reverse step, this exact chain uses the same
BBDM reverse formula before replacing \(x_0\) by its posterior mean. Thus, for
one reverse transition, it samples \(x_0\) from the appropriate conditional
posterior and then applies the bridge update.

We first consider the terminal reverse update \(s=S\). By the endpoint
convention, the selected-label chain uses \(\hat x_0=\mu_{r|y}\). Therefore,
after projection to the \(U_r\)-basis,
\[
x_{S-1}^{U_r,\mathrm{sel}}
=
m_{S-1}y^{U_r}
+
(1-m_{S-1})\mu_{r|y}^{U_r}
+
\sqrt{\delta_{S-1}}z_S^{U_r}.
\]
Conditioned on \((y,J=r)\), the first two terms are deterministic and
\(z_S^{U_r}\sim\mathcal N(0,I)\). Hence
\[
\operatorname{Cov}(x_{S-1}^{U_r,\mathrm{sel}}\mid y,J=r)
=
\delta_{S-1}I.
\]

For the exact component-conditioned chain, the terminal update uses a posterior
sample
\[
x_0\mid y,C=r\sim\mathcal N(\mu_{r|y},\Sigma_{r|y}).
\]
Thus
\[
x_{S-1}^{U_r,\mathrm{ex}}
=
m_{S-1}y^{U_r}
+
(1-m_{S-1})x_0^{U_r}
+
\sqrt{\delta_{S-1}}z_S^{U_r},
\]
where \(z_S^{U_r}\sim\mathcal N(0,I)\) is independent of \(x_0^{U_r}\). Again,
the term \(m_{S-1}y^{U_r}\) is deterministic after conditioning on \(y\), so
\[
\begin{aligned}
\operatorname{Cov}(x_{S-1}^{U_r,\mathrm{ex}}\mid y,C=r)
&=
(1-m_{S-1})^2
\operatorname{Cov}(x_0^{U_r}\mid y,C=r)
+
\delta_{S-1}I
=
(1-m_{S-1})^2
U_r^\top\Sigma_{r|y}U_r
+
\delta_{S-1}I.
\end{aligned}
\]
Since
\[
U_r^\top\Sigma_{r|y}U_r
=
\diag\!\left(
\frac{1}{\lambda_{r,1}},\ldots,\frac{1}{\lambda_{r,d}}
\right),
\]
the covariance gap at step \(S-1\) is
\[
\begin{aligned}
\Delta_{S-1}
&\triangleq
\operatorname{Cov}(x_{S-1}^{U_r,\mathrm{ex}}\mid y,C=r)
-
\operatorname{Cov}(x_{S-1}^{U_r,\mathrm{sel}}\mid y,J=r)
\\
&=
(1-m_{S-1})^2
\diag\!\left(
\frac{1}{\lambda_{r,1}},\ldots,\frac{1}{\lambda_{r,d}}
\right)
\succeq0.
\end{aligned}
\]

We now show that this covariance ordering propagates through every interior
reverse step. Fix \(1\le s\le S-1\). For the selected-label chain, we already
have the projected recursion
\[
x_{s-1}^{U_r,\mathrm{sel}}
=
G_r(s)x_s^{U_r,\mathrm{sel}}
+
N_r(s)y^{U_r}
+
M_r(s)\mu_{r|y}^{U_r}
+
\sigma_s z_s^{U_r}.
\]
Conditioned on \((y,J=r)\), the quantities \(y^{U_r}\) and
\(\mu_{r|y}^{U_r}\) are deterministic. Therefore taking covariance gives
\[
\operatorname{Cov}(x_{s-1}^{U_r,\mathrm{sel}}\mid y,J=r)
=
G_r(s)
\operatorname{Cov}(x_s^{U_r,\mathrm{sel}}\mid y,J=r)
G_r(s)^\top
+
\sigma_s^2I.
\]

For the exact component-conditioned chain, one reverse transition has the form
\[
x_{s-1}
=
a_sx_0+b_sy+c_sx_s+\sigma_s z_s.
\]
After projection to the \(U_r\)-basis,
\[
x_{s-1}^{U_r,\mathrm{ex}}
=
a_sx_0^{U_r}
+
b_sy^{U_r}
+
c_sx_s^{U_r,\mathrm{ex}}
+
\sigma_s z_s^{U_r}.
\]
For \(1\le s\le S-1\), the conditional posterior of \(x_0\) is
\[
x_0\mid x_s,y,C=r
\sim
\mathcal N(\mu_{r|s}(x_s,y),\Sigma_{r|s}),
\]
and in the \(U_r\)-basis,
\[
U_r^\top\Sigma_{r|s}U_r=\Lambda_{r|s}.
\]
Moreover,
\[
U_r^\top\mu_{r|s}(x_s,y)
=
\Lambda_{r|s}
\left(
\diag(\lambda_{r,1},\ldots,\lambda_{r,d})\mu_{r|y}^{U_r}
+
\frac{1-m_s}{\delta_s}(x_s^{U_r}-m_sy^{U_r})
\right).
\]
Therefore, conditioned on \(x_s^{U_r,\mathrm{ex}}\), \(y\), and \(C=r\),
\[
\begin{aligned}
&\mathbb E[x_{s-1}^{U_r,\mathrm{ex}}
\mid x_s^{U_r,\mathrm{ex}},y,C=r]
\\
&=
a_s U_r^\top\mu_{r|s}(x_s^{\mathrm{ex}},y)
+
b_sy^{U_r}
+
c_sx_s^{U_r,\mathrm{ex}}
\\
&=
\left(c_sI+\frac{a_s(1-m_s)}{\delta_s}\Lambda_{r|s}\right)
x_s^{U_r,\mathrm{ex}}
+
\left(b_sI-\frac{a_sm_s(1-m_s)}{\delta_s}\Lambda_{r|s}\right)
y^{U_r}
\\
&\qquad
+
a_s\Lambda_{r|s}
\diag(\lambda_{r,1},\ldots,\lambda_{r,d})
\mu_{r|y}^{U_r}
\\
&=
G_r(s)x_s^{U_r,\mathrm{ex}}
+
N_r(s)y^{U_r}
+
M_r(s)\mu_{r|y}^{U_r}.
\end{aligned}
\]
Thus the exact chain and the selected-label chain have the same affine
conditional mean. The difference is in the conditional covariance. Given
\(x_s^{U_r,\mathrm{ex}}\), \(y\), and \(C=r\), the only random terms in
\[
x_{s-1}^{U_r,\mathrm{ex}}
=
a_sx_0^{U_r}
+
b_sy^{U_r}
+
c_sx_s^{U_r,\mathrm{ex}}
+
\sigma_s z_s^{U_r}
\]
are \(x_0^{U_r}\) and \(z_s^{U_r}\). These two are independent, and
\[
\operatorname{Cov}(x_0^{U_r}\mid x_s^{U_r,\mathrm{ex}},y,C=r)
=
\Lambda_{r|s}.
\]
Hence
\[
\operatorname{Cov}(x_{s-1}^{U_r,\mathrm{ex}}
\mid x_s^{U_r,\mathrm{ex}},y,C=r)
=
a_s^2\Lambda_{r|s}
+
\sigma_s^2I.
\]

Now apply the law of total covariance:
\[
\begin{aligned}
&\operatorname{Cov}(x_{s-1}^{U_r,\mathrm{ex}}\mid y,C=r)
\\
&=
\operatorname{Cov}\!\left(
\mathbb E[x_{s-1}^{U_r,\mathrm{ex}}
\mid x_s^{U_r,\mathrm{ex}},y,C=r]
\mid y,C=r
\right)
\\
&\qquad
+
\mathbb E\!\left[
\operatorname{Cov}(x_{s-1}^{U_r,\mathrm{ex}}
\mid x_s^{U_r,\mathrm{ex}},y,C=r)
\mid y,C=r
\right].
\end{aligned}
\]
Using the conditional mean and conditional covariance computed above gives
\[
\operatorname{Cov}(x_{s-1}^{U_r,\mathrm{ex}}\mid y,C=r)
=
G_r(s)
\operatorname{Cov}(x_s^{U_r,\mathrm{ex}}\mid y,C=r)
G_r(s)^\top
+
\sigma_s^2I
+
a_s^2\Lambda_{r|s}.
\]

We can now subtract the selected covariance recursion from the exact covariance
recursion. For \(s=0,\ldots,S-1\), define
\[
\Delta_s
\triangleq
\operatorname{Cov}(x_s^{U_r,\mathrm{ex}}\mid y,C=r)
-
\operatorname{Cov}(x_s^{U_r,\mathrm{sel}}\mid y,J=r).
\]
For \(1\le s\le S-1\), the subtraction gives
\[
\Delta_{s-1}
=
G_r(s)\Delta_sG_r(s)^\top
+
a_s^2\Lambda_{r|s}.
\]
If \(\Delta_s\succeq0\), then
\[
G_r(s)\Delta_sG_r(s)^\top\succeq0,
\]
and since \(\Lambda_{r|s}\succeq0\), also
\[
a_s^2\Lambda_{r|s}\succeq0.
\]
Therefore
\[
\Delta_s\succeq0
\quad\Longrightarrow\quad
\Delta_{s-1}\succeq0.
\]
We already proved that \(\Delta_{S-1}\succeq0\). Applying the implication
successively for \(s=S-1,S-2,\ldots,1\) yields
\[
\Delta_s\succeq0,
\qquad
s=0,\ldots,S-1.
\]

It remains only to identify what \(\Delta_0\succeq0\) says. The exact
component-conditioned chain is the reverse-time conditional chain of the same
bridge model after conditioning on \(y\) and \(C=r\). Therefore its clean
endpoint has marginal
\[
x_0\mid y,C=r
\sim
\mathcal N(\mu_{r|y},\Sigma_{r|y}).
\]
Consequently,
\[
\operatorname{Cov}(x_0^{U_r,\mathrm{ex}}\mid y,C=r)
=
U_r^\top\Sigma_{r|y}U_r.
\]
At the same time, the selected-label chain endpoint is exactly
\(\hat x_0^{U_r,(r)}\), whose covariance conditioned on \((y,J=r)\) is
\[
\operatorname{Cov}(\hat x_0^{U_r,(r)}\mid y,J=r)
=
U_r^\top\Sigma_{\mathrm{BBDM}}^{(r)}U_r.
\]
Thus \(\Delta_0\succeq0\) gives
\[
U_r^\top\Sigma_{r|y}U_r
-
U_r^\top\Sigma_{\mathrm{BBDM}}^{(r)}U_r
\succeq0.
\]
Multiplying by \(U_r\) and \(U_r^\top\) gives
\[
\Sigma_{r|y}-\Sigma_{\mathrm{BBDM}}^{(r)}\succeq0.
\]

Finally, both covariances are diagonal in the \(U_r\)-basis:
\[
U_r^\top\Sigma_{r|y}U_r
=
\diag\!\left(
\frac{1}{\lambda_{r,1}},\ldots,\frac{1}{\lambda_{r,d}}
\right),
\]
and
\[
U_r^\top\Sigma_{\mathrm{BBDM}}^{(r)}U_r
=
\diag\!\left(
\sigma_{\mathrm{BBDM},r,1}^2,\ldots,
\sigma_{\mathrm{BBDM},r,d}^2
\right).
\]
The positive-semidefinite comparison therefore implies, coordinate by
coordinate,

\[
0
\le
\sigma_{\mathrm{BBDM},r,k}^2
\le
\frac{1}{\lambda_{r,k}},
\qquad
k=1,\ldots,d.
\]
This proves the proposition.

\newpage

\section{Schedule objectives and bounded schedule family}
\label{app:schedule_objectives_derivations}

This appendix derives the two schedule objectives used in Section~\ref{sec:comparison_v6} and makes explicit the coupling choices behind the objectives. It also explains why the scheduling parameterization that is offered is valid under the bridge constraints. 

\subsection*{E.1 True and selected laws and the matched-index coupling}

From Appendix~\ref{app:mog_exact_posterior_derivation}, the true measurement posterior is
\begin{equation}
p(x_0\mid y)
=
\sum_{r=1}^{R}
\gamma_{r|y}(y)\,
\mathcal N\!\bigl(x_0;\mu_{r|y},\Sigma_{r|y}\bigr).
\label{eq:mog_true_posterior_mog_problem_basis}
\end{equation}
The selected-label law is
\begin{equation}
p_{\mathrm{Select}}(\hat x_0\mid y)
=
\sum_{r=1}^{R}
\gamma_{r|y}(y)
\mathcal N\!\bigl(
\hat x_0;
\mu_{\mathrm{BBDM}}^{(r)}(y),
\Sigma_{\mathrm{BBDM}}^{(r)}
\bigr).
\label{eq:app_selected_law_repeat_e}
\end{equation}
Thus the two laws have the same mixture weights. We couple them by first drawing a common index
\[
R_y\sim \mathrm{Categorical}(\gamma_{1|y}(y),\dots,\gamma_{R|y}(y)),
\]
and then, conditioned on $R_y=r$, coupling the two Gaussian components. This gives the valid transport plan. Therefore, from the definition of the W2 distance:
\begin{align}
&W_2^2\!\left(p(x_0\mid y),p_{\mathrm{Select}}(\hat x_0\mid y)\right)
\notag\\
&\qquad\le
\sum_{r=1}^{R}\gamma_{r|y}(y)
W_2^2\!\left(
\mathcal N(\mu_{r|y},\Sigma_{r|y}),
\mathcal N(\mu_{\mathrm{BBDM}}^{(r)}(y),\Sigma_{\mathrm{BBDM}}^{(r)})
\right).
\label{eq:app_matched_component_w2_bound_e}
\end{align}
The inequality may be strict because the optimal coupling between the two full mixtures is allowed to split mass across different component labels.

\subsection*{E.2 Gaussian $W_2$ formula and cancellation of the mean term}

For two Gaussian laws
\[
P_1=\mathcal N(\mu_1,\Sigma_1),
\qquad
P_2=\mathcal N(\mu_2,\Sigma_2),
\]
the squared 2-Wasserstein distance has the standard closed form
\citep{dowson1982frechet,peyre2019computational}:
\begin{equation}
W_2^2(P_1,P_2)
=
\|\mu_1-\mu_2\|_2^2
+
\tr\!\left(
\Sigma_1+
\Sigma_2-
2\left(\Sigma_2^{1/2}\Sigma_1\Sigma_2^{1/2}\right)^{1/2}
\right).
\label{eq:app_gaussian_w2_full_formula_e}
\end{equation}
For the matched component $r$, Proposition~\ref{prop:selected_label_mean_exactness} gives
\begin{equation}
\mu_{\mathrm{BBDM}}^{(r)}(y)=\mu_{r|y},
\qquad
\|\mu_{r|y}-\mu_{\mathrm{BBDM}}^{(r)}(y)\|_2^2=0.
\label{eq:app_mean_term_cancels_e}
\end{equation}
It remains only to compare the covariances. In the basis $U_r$,
\begin{equation}
U_r^\top\Sigma_{r|y}U_r
=
\diag\!\left(\frac{1}{\lambda_{r,1}},\dots,\frac{1}{\lambda_{r,d}}\right),
\label{eq:app_true_cov_diag_e}
\end{equation}
whereas
\begin{equation}
U_r^\top\Sigma_{\mathrm{BBDM}}^{(r)}U_r
=
\diag\!\left(\sigma_{\mathrm{BBDM},r,1}^2,
\dots,
\sigma_{\mathrm{BBDM},r,d}^2\right).
\label{eq:app_bbdm_cov_diag_e}
\end{equation}
Because the two covariance matrices are diagonal in the same basis, the covariance part of \eqref{eq:app_gaussian_w2_full_formula_e} reduces coordinate-wise. Hence
\begin{equation}
W_2^2\!\left(
\mathcal N(\mu_{r|y},\Sigma_{r|y}),
\mathcal N(\mu_{\mathrm{BBDM}}^{(r)}(y),\Sigma_{\mathrm{BBDM}}^{(r)})
\right)
=
\sum_{k=1}^{d}
\left(
\sigma_{\mathrm{BBDM},r,k}-\frac{1}{\sqrt{\lambda_{r,k}}}
\right)^2.
\label{eq:mog_pointwise_upper_bound_problem_basis}
\end{equation}

\subsection*{E.3 The Wasserstein upper-bound schedule objective}

Average \eqref{eq:mog_pointwise_upper_bound_problem_basis} over the marginal law of $y$ and use
\[
\gamma_{r|y}(y)p(y)=\pi_rp(y\mid C=r).
\]
Since the covariance term does not depend on the realized value of $y$, the matched-component Wasserstein upper-bound objective reduces to
\begin{equation}
\JMOGW
=
\sum_{r=1}^{R}\pi_r
\sum_{k=1}^{d}
\left(\sigma_{\mathrm{BBDM},r,k}-\frac{1}{\sqrt{\lambda_{r,k}}}\right)^2.
\label{eq:mog_w2_final_objective}
\end{equation}
This is the expression stated in Corollary~\ref{cor:wasserstein_schedule_objective}.

\subsection*{E.4 The expected-MSE schedule objective and its coupling}

Fix $y$. First draw a label
\[
R_y\sim \gamma_{\cdot|y}(y).
\]
Then, conditioned on $R_y=r$, draw
\[
X\sim\mathcal N(\mu_{r|y},\Sigma_{r|y}),
\qquad
\hat X\sim
\mathcal N(\mu_{\mathrm{BBDM}}^{(r)}(y),
\Sigma_{\mathrm{BBDM}}^{(r)}),
\]
independently. Thus the two samples use the same mixture component, but are independent inside that component.

For a fixed matched component $r$,
\begin{align}
\mathbb E\bigl[\|X-\hat X\|_2^2\mid y,R_y=r\bigr]
&=
\mathbb E\bigl[
\|(X-\mu_{r|y})
-(\hat X-\mu_{\mathrm{BBDM}}^{(r)}(y))
\nonumber\\
&\qquad\qquad
+(\mu_{r|y}-\mu_{\mathrm{BBDM}}^{(r)}(y))\|_2^2
\bigr]
\notag\\
&=
\tr(\Sigma_{r|y})
+
\tr(\Sigma_{\mathrm{BBDM}}^{(r)})
+
\|\mu_{r|y}-\mu_{\mathrm{BBDM}}^{(r)}(y)\|_2^2.
\label{eq:app_mse_matched_component_before_mean_exact}
\end{align}
The cross terms vanish because the two centered Gaussian residuals have zero mean and are sampled independently. By Proposition~\ref{prop:selected_label_mean_exactness},
\[
\mu_{\mathrm{BBDM}}^{(r)}(y)=\mu_{r|y},
\]
so the mean term in \eqref{eq:app_mse_matched_component_before_mean_exact} cancels. Hence
\begin{align}
\mathbb E\bigl[\|X-\hat X\|_2^2\mid y,R_y=r\bigr]
&=
\tr(\Sigma_{r|y})
+
\tr(\Sigma_{\mathrm{BBDM}}^{(r)}).
\label{eq:app_mse_matched_component_trace}
\end{align}
In the basis $U_r$,
\[
U_r^\top \Sigma_{r|y}U_r
=
\diag\!\left(
\frac{1}{\lambda_{r,1}},\dots,\frac{1}{\lambda_{r,d}}
\right),
\]
and
\[
U_r^\top \Sigma_{\mathrm{BBDM}}^{(r)}U_r
=
\diag\!\left(
\sigma_{\mathrm{BBDM},r,1}^2,\dots,
\sigma_{\mathrm{BBDM},r,d}^2
\right).
\]
Since the trace is invariant under orthogonal changes of basis,
\begin{align}
\mathbb E\bigl[\|X-\hat X\|_2^2\mid y,R_y=r\bigr]
&=
\sum_{k=1}^{d}
\left(
\frac{1}{\lambda_{r,k}}
+
\sigma_{\mathrm{BBDM},r,k}^2
\right).
\label{eq:mog_conditional_mse_basis}
\end{align}

Again Averaging \eqref{eq:mog_conditional_mse_basis} over the marginal law of $y$ and using,
\[
\gamma_{r|y}(y)p(y)=\pi_rp(y\mid C=r).
\]

we get,

\begin{equation}
\JMOGM
=
\sum_{r=1}^{R}\pi_r
\sum_{k=1}^{d}
\left(
\sigma_{\mathrm{BBDM},r,k}^2
+
\frac{1}{\lambda_{r,k}}
\right).
\label{eq:mog_expected_mse_final}
\end{equation}

It is important to distinguish this matched-label product coupling from an independent-label coupling. Suppose instead that, for fixed $y$, the true posterior sample uses a label $C$ and the selected-label reconstruction uses a label $J$, where
\[
C\sim\gamma_{\cdot|y}(y),
\qquad
J\sim\gamma_{\cdot|y}(y),
\]
independently. Conditioned on $C=r$ and $J=j$,
\begin{align}
\mathbb E\bigl[\|X-\hat X\|_2^2\mid y,C=r,J=j\bigr]
&=
\tr(\Sigma_{r|y})
+
\tr(\Sigma_{\mathrm{BBDM}}^{(j)})
\nonumber\\
&\quad+
\|\mu_{r|y}-\mu_{\mathrm{BBDM}}^{(j)}(y)\|_2^2.
\end{align}
Using again $\mu_{\mathrm{BBDM}}^{(j)}(y)=\mu_{j|y}$, and averaging over the independent labels, gives
\begin{align}
\mathbb E\|X-\hat X\|_2^2
&=
\sum_{r=1}^{R}\gamma_{r|y}(y)\tr(\Sigma_{r|y})
+
\sum_{j=1}^{R}\gamma_{j|y}(y)\tr(\Sigma_{\mathrm{BBDM}}^{(j)})
\notag\\
&\quad+
\sum_{r=1}^{R}\sum_{j=1}^{R}
\gamma_{r|y}(y)\gamma_{j|y}(y)
\|\mu_{r|y}-\mu_{j|y}\|_2^2.
\label{eq:app_independent_label_mse_e}
\end{align}
The matched-label MSE and the independent-label MSE generally have different absolute values, because the independent-label version includes an additional between-component mean mismatch, but they lead to the same schedule minimizer.

\subsection*{E.5 Formulation of the optimization problem and bounded parameter box}

The schedule-design problem is to choose
\[
\Theta=\{(m_s,\delta_s)\}_{s=1}^{S}
\]
so as to minimize one of the two criteria,
\begin{equation}
\min_{\Theta}\;\mathcal J(\Theta),
\qquad
\mathcal J\in\{\JMOGW,\JMOGM\}.
\label{eq:mog_generic_optimization_problem}
\end{equation}
Because direct optimization over the discrete schedule values is inconvenient,
we use the four-parameter family
\begin{equation}
m_s
=
1-(1-\tau_s^\alpha)^\beta,
\qquad
\delta_s
=
c\,[4m_s(1-m_s)]^\gamma,
\qquad
\tau_s=\frac{s}{S}.
\label{eq:mog_schedule_family_parametric}
\end{equation}
A simple sufficient bridge-valid parameter region for this family is
\[
\alpha>0,\qquad \beta>0,\qquad c>0,\qquad 0<\gamma\le 2.
\]

In the experiments and schedule search we use the bounded regular subset
\begin{equation}
\begin{aligned}
\alpha&\in[1,2],
&\beta&\in[1,2],\\
c&\in[0.2,2],
&\gamma&\in[0.2,2].
\end{aligned}
\label{eq:app_bounded_parameter_box_e}
\end{equation}

The bridge constraints are
\begin{enumerate}
    \item $m_0=0$, $m_S=1$, $\delta_0=0$, $\delta_S=0$;
    \item $0<m_1<\cdots<m_{S-1}<1$;
    \item $\delta_s\ge 0$;
    \item $\delta_s-\delta_{s-1}\frac{(1-m_s)^2}{(1-m_{s-1})^2}\ge 0$.
\end{enumerate}
At $\tau_0=0$, \eqref{eq:mog_schedule_family_parametric} gives
$m_0=0$, hence $\delta_0=0$. At $\tau_S=1$, it gives
$m_S=1$, hence $\delta_S=0$. Moreover, for $\tau\in(0,1)$,
\[
\frac{dm}{d\tau}
=
\alpha\beta\,\tau^{\alpha-1}(1-\tau^\alpha)^{\beta-1}>0
\]
whenever $\alpha>0$ and $\beta>0$. Hence $m_s$ is strictly
increasing and lies in $(0,1)$ for all interior steps. Since
$4m_s(1-m_s)>0$ there, we also have $\delta_s>0$ for all
interior steps whenever $c>0$ and $\gamma>0$.

It remains to verify the forward-transition constraint. The endpoint
cases $s=1$ and $s=S$ are automatic from $\delta_0=0$ and
$(1-m_S)^2=0$. For the remaining interior transitions, the constraint
is equivalent to
\[
\frac{\delta_s}{(1-m_s)^2}
\ge
\frac{\delta_{s-1}}{(1-m_{s-1})^2}.
\]
Define
\[
g(m)=\frac{\delta}{(1-m)^2}
=
c\,4^\gamma m^\gamma(1-m)^{\gamma-2}.
\]
Differentiating with respect to $m$ gives
\[
g'(m)
=
c\,4^\gamma m^{\gamma-1}(1-m)^{\gamma-3}
\bigl[\gamma-2m(\gamma-1)\bigr].
\]
For $m\in(0,1)$, all factors outside the bracket are positive. If
$0<\gamma\le1$, the bracket is nonnegative for all $m\in(0,1)$. If
$\gamma>1$, its minimum over $(0,1)$ is approached as $m\to1$, where
it tends to $2-\gamma$. Hence $0<\gamma\le2$ guarantees
$g'(m)\ge0$ for all $m\in(0,1)$. Since $m_s$ is increasing, this
implies $g(m_s)\ge g(m_{s-1})$, which gives the forward-transition
validity condition.

Thus the family \eqref{eq:mog_schedule_family_parametric} satisfies
the Brownian-bridge constraints throughout the sufficient validity
region
\[
\alpha>0,\qquad \beta>0,\qquad c>0,\qquad 0<\gamma\le2.
\]
The box in \eqref{eq:app_bounded_parameter_box_e} is a bounded,
regular subset of this region. The bounds $\alpha,\beta\in[1,2]$ are
not imposed by the discrete bridge constraints themselves. They avoid
endpoint-singular slopes of the interpolation curve $m(\tau)$ and keep
the search in a moderate range. The bounds on $c$ and $\gamma$ keep the
bridge variance non-degenerate, while the upper bound $\gamma\le2$
preserves the sufficient transition condition above.

The need for a bounded search box is also motivated by the behavior of
the schedule objectives. By Proposition~\ref{prop:selected_label_covariance_deficit},
each selected-label standard deviation satisfies
\[
0\le \sigma_{\mathrm{BBDM},r,k}\le \frac{1}{\sqrt{\lambda_{r,k}}}.
\]
The MSE objective decreases as $\sigma_{\mathrm{BBDM},r,k}$ moves
toward $0$, whereas the Wasserstein objective decreases as
$\sigma_{\mathrm{BBDM},r,k}$ moves toward
$1/\sqrt{\lambda_{r,k}}$. Driving the sampler close to either boundary
can make the bridge parameters increasingly extreme, while the
objective improvements exhibit diminishing returns once the standard
deviations are already close to the preferred boundary. The bounded box
keeps the schedules in a moderate, non-degenerate range while still
allowing low-loss solutions for both objectives.
\clearpage

\section{MoG-Free Schedule Heuristics from a Variance Bound}
\label{app:mog_free_schedule_heuristics}

The schedule objectives derived in Corollaries~\ref{cor:wasserstein_schedule_objective}
and~\ref{cor:expected_mse_schedule_objective} are explicit once the MoG posterior
precisions $\lambda_{r,k}$ are known. However, fitting a reliable high-dimensional
MoG is generally impractical for image-scale data. For example, images of size
$256\times256\times3$ live in dimension $196{,}608$, and estimating mixture
covariances in those dimensions is extremely demanding.

This appendix derives simple schedule-parameter heuristics that do not require
fitting the full MoG.

\subsection*{F.1 A bridge precision scale}

Define, for $s=1,\dots,S-1$,
\begin{equation}
\rho_s
\triangleq
\frac{(1-m_s)^2}{\delta_s}.
\label{eq:appI_rho_definition}
\end{equation}
This quantity is the scalar added to the component posterior precision
$\lambda_{r,k}$ in the covariance
\[
\frac{1}{
\lambda_{r,k}+\frac{(1-m_s)^2}{\delta_s}
}
=
\frac{1}{\lambda_{r,k}+\rho_s}.
\]
We also use the endpoint conventions
\[
\rho_0=+\infty,
\qquad
\rho_S=0.
\]

The bridge transition constraint from Appendix~\ref{app:schedule_objectives_derivations} is
\begin{equation}
\delta_s-\delta_{s-1}
\frac{(1-m_s)^2}{(1-m_{s-1})^2}
\ge 0.
\label{eq:appI_bridge_constraint_original}
\end{equation}
Using
\[
\delta_s=\frac{(1-m_s)^2}{\rho_s},
\qquad
\delta_{s-1}=\frac{(1-m_{s-1})^2}{\rho_{s-1}},
\]
we get
\begin{align}
\delta_s-\delta_{s-1}
\frac{(1-m_s)^2}{(1-m_{s-1})^2}
&=
\frac{(1-m_s)^2}{\rho_s}
-
\frac{(1-m_{s-1})^2}{\rho_{s-1}}
\frac{(1-m_s)^2}{(1-m_{s-1})^2}
\nonumber\\
&=
\frac{(1-m_s)^2}{\rho_s}
-
\frac{(1-m_s)^2}{\rho_{s-1}}
\nonumber\\
&=
(1-m_s)^2
\left(
\frac{1}{\rho_s}
-
\frac{1}{\rho_{s-1}}
\right).
\label{eq:appI_bridge_constraint_rho}
\end{align}
Therefore the bridge constraint is equivalent to
\begin{equation}
\rho_s\le \rho_{s-1}.
\label{eq:appI_rho_monotone_constraint}
\end{equation}
Thus a valid bridge corresponds to a decreasing sequence
\[
+\infty=\rho_0\ge \rho_1\ge \rho_2\ge\cdots\ge \rho_{S-1}>0,
\qquad
\rho_S=0.
\]

For the bounded schedule family,
\[
\delta_s=c[4m_s(1-m_s)]^\gamma.
\]
Substituting this into \eqref{eq:appI_rho_definition} gives
\begin{equation}
\rho_s
=
\frac{(1-m_s)^2}{c[4m_s(1-m_s)]^\gamma}
=
\frac{1}{c4^\gamma}
m_s^{-\gamma}(1-m_s)^{2-\gamma}.
\label{eq:appI_rho_bounded_family}
\end{equation}
For $0<\gamma\le 2$, this is decreasing as a function of $m_s$. Indeed,
\begin{equation}
\frac{\partial}{\partial m}
\log\left[
\frac{1}{c4^\gamma}m^{-\gamma}(1-m)^{2-\gamma}
\right]
=
-\frac{\gamma}{m}
-
\frac{2-\gamma}{1-m}
<0.
\label{eq:appI_rho_decreasing_in_m}
\end{equation}

\subsection*{F.2 Exact selected-label variance in the $\rho$ scale}

We now rewrite the selected-label BBDM variance from
Proposition~\ref{prop:selected_label_covariance_deficit} in terms of $\rho_s$.

First, the BBDM bridge increment is
\begin{align}
\delta_{s|s-1}
&=
\delta_s-\delta_{s-1}
\frac{(1-m_s)^2}{(1-m_{s-1})^2}
\nonumber\\
&=
(1-m_s)^2
\left(
\frac{1}{\rho_s}
-
\frac{1}{\rho_{s-1}}
\right).
\label{eq:appI_delta_increment_rho}
\end{align}
Therefore
\begin{align}
\sigma_s^2
&=
\frac{\delta_{s|s-1}\delta_{s-1}}{\delta_s}
\nonumber\\
&=
\frac{
(1-m_s)^2
\left(
\frac{1}{\rho_s}
-
\frac{1}{\rho_{s-1}}
\right)
\cdot
\frac{(1-m_{s-1})^2}{\rho_{s-1}}
}{
\frac{(1-m_s)^2}{\rho_s}
}
\nonumber\\
&=
(1-m_{s-1})^2
\frac{\rho_{s-1}-\rho_s}{\rho_{s-1}^2}.
\label{eq:appI_sigma_s_rho}
\end{align}

Next, compute $c_s$. Since
\[
c_s=
\sqrt{
\frac{\delta_{s-1}-\sigma_s^2}{\delta_s}
},
\]
we first simplify the numerator:
\begin{align}
\delta_{s-1}-\sigma_s^2
&=
\frac{(1-m_{s-1})^2}{\rho_{s-1}}
-
(1-m_{s-1})^2
\frac{\rho_{s-1}-\rho_s}{\rho_{s-1}^2}
\nonumber\\
&=
(1-m_{s-1})^2
\left[
\frac{1}{\rho_{s-1}}
-
\frac{\rho_{s-1}-\rho_s}{\rho_{s-1}^2}
\right]
\nonumber\\
&=
(1-m_{s-1})^2
\frac{\rho_s}{\rho_{s-1}^2}.
\label{eq:appI_delta_minus_sigma_rho}
\end{align}
Hence
\begin{align}
c_s
&=
\sqrt{
\frac{
(1-m_{s-1})^2
\frac{\rho_s}{\rho_{s-1}^2}
}{
\frac{(1-m_s)^2}{\rho_s}
}
}
\nonumber\\
&=
\frac{1-m_{s-1}}{1-m_s}
\frac{\rho_s}{\rho_{s-1}}.
\label{eq:appI_cs_rho}
\end{align}
Then
\begin{align}
a_s
&=
(1-m_{s-1})-(1-m_s)c_s
\nonumber\\
&=
(1-m_{s-1})
-
(1-m_s)
\frac{1-m_{s-1}}{1-m_s}
\frac{\rho_s}{\rho_{s-1}}
\nonumber\\
&=
(1-m_{s-1})
\frac{\rho_{s-1}-\rho_s}{\rho_{s-1}}.
\label{eq:appI_as_rho}
\end{align}

The scalar diagonal entry of $G_r(s)$ is
\[
g_{r,k}(s)
=
c_s
+
\frac{a_s(1-m_s)}{\delta_s}
\frac{1}{\lambda_{r,k}+\rho_s}.
\]
Using
\[
\delta_s=\frac{(1-m_s)^2}{\rho_s},
\]
we have
\[
\frac{a_s(1-m_s)}{\delta_s}
=
\frac{a_s\rho_s}{1-m_s}.
\]
Therefore
\begin{align}
g_{r,k}(s)
&=
\frac{1-m_{s-1}}{1-m_s}
\frac{\rho_s}{\rho_{s-1}}
+
\frac{
(1-m_{s-1})
\frac{\rho_{s-1}-\rho_s}{\rho_{s-1}}
\cdot
\rho_s
}{
(1-m_s)(\lambda_{r,k}+\rho_s)
}
\nonumber\\
&=
\frac{1-m_{s-1}}{1-m_s}
\frac{\rho_s}{\rho_{s-1}}
\left[
1+
\frac{\rho_{s-1}-\rho_s}{\lambda_{r,k}+\rho_s}
\right]
\nonumber\\
&=
\frac{1-m_{s-1}}{1-m_s}
\frac{\rho_s}{\rho_{s-1}}
\frac{\lambda_{r,k}+\rho_{s-1}}{\lambda_{r,k}+\rho_s}.
\label{eq:appI_gain_rho}
\end{align}

Now telescope the product in Proposition~\ref{prop:selected_label_covariance_deficit}. For $i\ge2$,
\begin{align}
\prod_{j=1}^{i-1}g_{r,k}(j)^2
&=
\left[
\prod_{j=1}^{i-1}
\frac{1-m_{j-1}}{1-m_j}
\frac{\rho_j}{\rho_{j-1}}
\frac{\lambda_{r,k}+\rho_{j-1}}{\lambda_{r,k}+\rho_j}
\right]^2
\nonumber\\
&=
\left[
\frac{1}{1-m_{i-1}}
\cdot
\frac{\rho_{i-1}}{\rho_0}
\cdot
\frac{\lambda_{r,k}+\rho_0}{\lambda_{r,k}+\rho_{i-1}}
\right]^2.
\label{eq:appI_product_before_limit}
\end{align}
Using the endpoint convention $\rho_0=+\infty$,
\[
\frac{\rho_{i-1}}{\rho_0}
\frac{\lambda_{r,k}+\rho_0}{\lambda_{r,k}+\rho_{i-1}}
\longrightarrow
\frac{\rho_{i-1}}{\lambda_{r,k}+\rho_{i-1}}.
\]
Hence
\begin{equation}
\prod_{j=1}^{i-1}g_{r,k}(j)^2
=
\frac{\rho_{i-1}^2}{
(1-m_{i-1})^2(\lambda_{r,k}+\rho_{i-1})^2
}.
\label{eq:appI_product_rho}
\end{equation}
Multiplying \eqref{eq:appI_product_rho} with \eqref{eq:appI_sigma_s_rho} gives
\begin{align}
\sigma_i^2
\prod_{j=1}^{i-1}g_{r,k}(j)^2
&=
(1-m_{i-1})^2
\frac{\rho_{i-1}-\rho_i}{\rho_{i-1}^2}
\cdot
\frac{\rho_{i-1}^2}{
(1-m_{i-1})^2(\lambda_{r,k}+\rho_{i-1})^2
}
\nonumber\\
&=
\frac{\rho_{i-1}-\rho_i}{(\lambda_{r,k}+\rho_{i-1})^2}.
\label{eq:appI_single_variance_term_rho}
\end{align}
The $i=1$ term vanishes under the convention $\rho_0=+\infty$. Therefore
\begin{equation}
\sigma_{\mathrm{BBDM},r,k}^2
=
\sum_{i=2}^{S}
\frac{
\rho_{i-1}-\rho_i
}{
(\lambda_{r,k}+\rho_{i-1})^2
}.
\label{eq:appI_exact_sigma_rho}
\end{equation}
This identity is exact inside the selected-label model.

\subsection*{F.3 MSE upper bound and edge heuristic}

By Corollary~\ref{cor:expected_mse_schedule_objective},
\[
\JMOGM
=
\sum_{r=1}^{R}\pi_r
\sum_{k=1}^{d}
\left(
\frac{1}{\lambda_{r,k}}
+
\sigma_{\mathrm{BBDM},r,k}^2
\right).
\]
The first term is independent of the schedule, so the schedule-dependent part is
only
\[
\sum_{r=1}^{R}\pi_r
\sum_{k=1}^{d}
\sigma_{\mathrm{BBDM},r,k}^2.
\]

For each fixed pair $(r,k)$, the function
\[
\rho\mapsto \frac{1}{(\lambda_{r,k}+\rho)^2}
\]
is decreasing in $\rho$. From \eqref{eq:appI_exact_sigma_rho},
\[
\sigma_{\mathrm{BBDM},r,k}^2
=
\sum_{i=2}^{S}
(\rho_{i-1}-\rho_i)
\frac{1}{(\lambda_{r,k}+\rho_{i-1})^2}.
\]
On the interval $[\rho_i,\rho_{i-1}]$, the endpoint value at $\rho_{i-1}$ is the
smallest value. Hence
\begin{equation}
(\rho_{i-1}-\rho_i)
\frac{1}{(\lambda_{r,k}+\rho_{i-1})^2}
\le
\int_{\rho_i}^{\rho_{i-1}}
\frac{d\rho}{(\lambda_{r,k}+\rho)^2}.
\label{eq:appI_lower_sum_mse_step}
\end{equation}
Summing over $i=2,\dots,S$ gives
\begin{align}
\sigma_{\mathrm{BBDM},r,k}^2
&\le
\int_0^{\rho_1}
\frac{d\rho}{(\lambda_{r,k}+\rho)^2}
\nonumber\\
&=
\frac{1}{\lambda_{r,k}}
-
\frac{1}{\lambda_{r,k}+\rho_1}
\nonumber\\
\label{eq:appI_sigma_mse_upper_rho}
\end{align}
Thus the MSE upper bound is minimized by making $\rho_1$ as small as possible.

For the bounded schedule family,
\[
\rho_1
=
\frac{1}{c4^\gamma}
m_1^{-\gamma}(1-m_1)^{2-\gamma},
\qquad
m_1=1-(1-S^{-\alpha})^\beta.
\]
We now inspect the parameter directions.

First,
\[
\frac{\partial \log \rho_1}{\partial c}
=
-\frac{1}{c}<0,
\]
so increasing $c$ decreases $\rho_1$.

Second,
\[
\frac{\partial \log \rho_1}{\partial m_1}
=
-\frac{\gamma}{m_1}
-
\frac{2-\gamma}{1-m_1}
<0
\]
for $0<\gamma\le2$. Therefore increasing $m_1$ decreases $\rho_1$.

Now
\[
m_1=1-(1-S^{-\alpha})^\beta.
\]
For $\beta$,
\[
\frac{\partial m_1}{\partial \beta}
=
-(1-S^{-\alpha})^\beta
\log(1-S^{-\alpha})
>0,
\]
because $0<1-S^{-\alpha}<1$. Thus increasing $\beta$ increases $m_1$ and
therefore decreases $\rho_1$.

For $\alpha$,
\[
\frac{\partial m_1}{\partial \alpha}
=
-\beta(1-S^{-\alpha})^{\beta-1}
(\log S)S^{-\alpha}
<0.
\]
Thus decreasing $\alpha$ increases $m_1$ and therefore decreases $\rho_1$.

Finally,
\[
\frac{\partial \log\rho_1}{\partial \gamma}
=
-\log(4m_1(1-m_1)).
\]
Since $4m_1(1-m_1)\le1$, we have
\[
-\log(4m_1(1-m_1))\ge0.
\]
Thus decreasing $\gamma$ decreases $\rho_1$.

Consequently, over the bounded box
\[
\alpha\in[1,2],
\qquad
\beta\in[1,2],
\qquad
c\in[0.2,2],
\qquad
\gamma\in[0.2,2],
\]
the MSE upper bound is minimized by the edge choice
\begin{equation}
\boxed{
(\alpha,\beta,c,\gamma)=(1,2,2,0.2).
}
\label{eq:appI_mse_edge_schedule}
\end{equation}
This conclusion does not require the numerical values of the $\lambda_{r,k}$'s.

\subsection*{F.4 W2 heuristic: use the opposite edge}

By Corollary~\ref{cor:wasserstein_schedule_objective}, the selected-label
Wasserstein objective is
\[
\JMOGW
=
\sum_{r=1}^{R}\pi_r
\sum_{k=1}^{d}
\left(
\sigma_{\mathrm{BBDM},r,k}
-
\frac{1}{\sqrt{\lambda_{r,k}}}
\right)^2.
\]
By Proposition~\ref{prop:selected_label_covariance_deficit},
\[
0
\le
\sigma_{\mathrm{BBDM},r,k}
\le
\frac{1}{\sqrt{\lambda_{r,k}}}.
\]
Therefore, coordinate-wise, the Wasserstein objective is reduced when
$\sigma_{\mathrm{BBDM},r,k}$ is increased toward the posterior standard deviation.

The bound in \eqref{eq:appI_sigma_mse_upper_rho} gives
\[
\sigma_{\mathrm{BBDM},r,k}^2
\le
\frac{1}{\lambda_{r,k}}
-
\frac{1}{\lambda_{r,k}+\rho_1}
\]

The right-hand side is increasing in $\rho_1$. Hence, as a simple MoG-free W2
heuristic, we take the opposite direction from the MSE rule: instead of minimizing
$\rho_1$, we choose the parameters that increase $\rho_1$.

Thus, over the bounded box, the W2-oriented heuristic reverses the MSE edge:
\[
\alpha \uparrow,
\qquad
\beta \downarrow,
\qquad
c \downarrow,
\qquad
\gamma \uparrow.
\]
Equivalently, the edge candidate is
\begin{equation}
\boxed{
(\alpha,\beta,c,\gamma)=(2,1,0.2,2).
}
\label{eq:appI_w2_opposite_edge}
\end{equation}
This is not a certificate of a universal W2 optimum, but a practical schedule
heuristic: it increases the upper bound on the sampler variance, which is preferred for matching the posterior variance.

\clearpage

\section{Selected-chain self-consistency}
\label{app:onehot_proof_outline_v6}

This appendix states and proves a self-consistency result supporting the
selected-label approximation. The result shows that, in a linearly separated
shared-covariance MoG regime, freezing the correct component label is
self-consistent: along the selected-label chain, the exact MoG responsibilities
continue to favor the same label with high probability.

\begin{assumption}
\label{ass:shared_covariance_main}
All mixture covariances are shared:
\[
\Sigma_r=\Sigma\succ0
\qquad
\text{for all } r\in\{1,\dots,R\}.
\]
\end{assumption}

\begin{theorem}
\label{thm:selected_chain_self_consistency}

Define the pairwise measurement separation
\begin{equation}
M_{rj}
\triangleq
\bigl(H(\mu_r-\mu_j)\bigr)^\top
\bigl(H\Sigma H^\top+\sigma_y^2I\bigr)^{-1}
\bigl(H(\mu_r-\mu_j)\bigr).
\label{eq:measurement_separation_mainpaper_v6}
\end{equation}Fix a true component $r^\star$. Suppose that $R$ and $S$ are fixed,
$\pi_r\ge\pi_{\min}>0$, and for some $\alpha>0$,
\begin{equation}
M_{r^\star j}\ge \alpha d
\qquad
\text{for all } j\neq r^\star.
\label{eq:main_linear_separation_assumption_v6}
\end{equation}
Generate $y$ under $C=r^\star$, and run the selected-label chain with the label
forced to $J=r^\star$, producing $X_s^{\mathrm{sel}}$. Let
$\mathcal E^{\mathrm{sel}}_{r^\star,\varepsilon}$ be the event that
\[
\begin{gathered}
\gamma_{r^\star\mid y}(y)\ge 1-\varepsilon,\\
\gamma_{r^\star\mid s}(X_s^{\mathrm{sel}},y)\ge 1-\varepsilon
\quad\forall s=1,\dots,S-1.
\end{gathered}
\]
For every fixed $\varepsilon\in(0,1/2)$, there exist $c,K>0$ and
$d_0\in\mathbb N$, independent of $d$, such that for all $d\ge d_0$,
\begin{equation}
\PP\!\left(
\mathcal E^{\mathrm{sel}}_{r^\star,\varepsilon}
\mid C=r^\star
\right)
\ge 1-KSe^{-cd}.
\label{eq:selected_chain_self_consistency_main}
\end{equation}
Since $J\sim\gamma_{\cdot\mid y}(y)$, also
\[
\PP(J=r^\star\mid C=r^\star)
\ge
1-\varepsilon-KSe^{-cd}.
\]
\end{theorem}

\subsection*{Proof outline}

The proof has two steps. First, we prove a bridge-marginal one-hot
concentration result. Second, we transfer the same concentration mechanism to
the selected-label chain, using the fact that the frozen surrogate is mean exact
and covariance deficient compared to the target posterior. We also record a
deterministic local perturbation bound between the exact MoG update and the
frozen-label update.

\subsection*{Shared-covariance model}

Under Assumption~\ref{ass:shared_covariance_main}, all components share the
same covariance. Thus,
\[
x_0\mid (C=r)\sim \mathcal N(\mu_r,\Sigma),
\qquad
y\mid x_0 \sim \mathcal N(Hx_0,\sigma_y^2 I).
\]

The
measurement-conditioned posterior covariance becomes component-independent:
\begin{equation}
\Sigma_{|y}^{-1}
=
\Sigma^{-1}
+
\frac{1}{\sigma_y^2}H^\top H.
\label{eq:mog_sigma_given_y_shared_problem_basis}
\end{equation}
The corresponding posterior mean for component \(r\) is
\begin{equation}
\mu_{r|y}
=
\Sigma_{|y}
\left(
\Sigma^{-1}\mu_r
+
\frac{1}{\sigma_y^2}H^\top y
\right).
\label{eq:mog_mu_given_y_shared_problem_basis}
\end{equation}

Therefore, for every pair \(r\neq j\),
\begin{equation}
\mu_{r|y}-\mu_{j|y}
=
\Sigma_{|y}\Sigma^{-1}(\mu_r-\mu_j),
\label{eq:mog_mu_diff_given_y_shared_problem_basis}
\end{equation}
which is independent of the realization of \(y\).

\subsection*{The conditional bridge law under a fixed component}

Under the shared-covariance assumption, the conditional law of the bridge
variable \(x_s\) given \((y,C=r)\) also has a covariance that does not depend on
\(r\). Indeed,
\begin{equation}
x_s\mid (y,C=r)
\sim
\mathcal N\!\bigl(u_{r,s}(y),V_s\bigr),
\label{eq:mog_bridge_law_shared_problem_basis}
\end{equation}
where
\begin{equation}
u_{r,s}(y)
=
(1-m_s)\mu_{r|y}+m_s y,
\label{eq:mog_bridge_mean_shared_problem_basis}
\end{equation}
and
\begin{equation}
V_s
=
(1-m_s)^2\Sigma_{|y}+\delta_s I.
\label{eq:mog_bridge_cov_shared_problem_basis}
\end{equation}

Using \eqref{eq:mog_mu_diff_given_y_shared_problem_basis}, the difference
between the component-conditioned bridge means is
\begin{equation}
u_{r,s}(y)-u_{j,s}(y)
=
(1-m_s)\Sigma_{|y}\Sigma^{-1}(\mu_r-\mu_j),
\label{eq:mog_bridge_mean_diff_shared_problem_basis}
\end{equation}
which is again independent of the realization of \(y\).

\subsection*{G.3 Two separation quantities}

The proof will be expressed through two pairwise separations.

\paragraph{Measurement separation.}
For each pair of components \(r\neq j\), define
\begin{equation}
M_{rj}
\triangleq
\bigl(H(\mu_r-\mu_j)\bigr)^\top
\bigl(H\Sigma H^\top+\sigma_y^2 I\bigr)^{-1}
\bigl(H(\mu_r-\mu_j)\bigr).
\label{eq:mog_measurement_separation_problem_basis}
\end{equation}

Equivalently, if we define
\[
A
\triangleq
H^\top\bigl(H\Sigma H^\top+\sigma_y^2 I\bigr)^{-1}H,
\]
then
\begin{equation}
M_{rj}
=
(\mu_r-\mu_j)^\top A(\mu_r-\mu_j).
\label{eq:mog_measurement_separation_A_problem_basis}
\end{equation}

\paragraph{Bridge separation at step \(s\).}
For each pair \(r\neq j\), define
\begin{equation}
B_{rj,s}
\triangleq
\bigl(u_{r,s}(y)-u_{j,s}(y)\bigr)^\top
V_s^{-1}
\bigl(u_{r,s}(y)-u_{j,s}(y)\bigr).
\label{eq:mog_bridge_separation_problem_basis}
\end{equation}
By \eqref{eq:mog_bridge_mean_diff_shared_problem_basis}, this quantity is
deterministic and does not depend on the realization of \(y\). It can be written as
\begin{equation}
B_{rj,s}
=
(1-m_s)^2
(\mu_r-\mu_j)^\top
\Sigma^{-1}\Sigma_{|y}\,V_s^{-1}\,\Sigma_{|y}\Sigma^{-1}
(\mu_r-\mu_j).
\label{eq:mog_bridge_separation_explicit_problem_basis}
\end{equation}

\subsection*{G.4 Pairwise log-posterior ratios}

We now study the pairwise log-ratio between the correct component and an
incorrect one.

\paragraph{Lemma 1}
For \(r\neq j\), define
\begin{equation}
L_{rj}^{(y)}(y)
\triangleq
\log\frac{\gamma_{r|y}(y)}{\gamma_{j|y}(y)}.
\label{eq:mog_measurement_logratio_def_problem_basis}
\end{equation}
Then, under the event \(C=r\),
\begin{equation}
L_{rj}^{(y)}(y)
\sim
\mathcal N\!\left(
\log\frac{\pi_r}{\pi_j}+\frac12 M_{rj},
\;
M_{rj}
\right).
\label{eq:mog_measurement_logratio_law_problem_basis}
\end{equation}

\paragraph{Proof.}
Under the shared-covariance model,
\[
y\mid (C=r)\sim \mathcal N(H\mu_r,\;H\Sigma H^\top+\sigma_y^2 I).
\]
Set
\[
S_y \triangleq H\Sigma H^\top+\sigma_y^2 I,
\qquad
m_r \triangleq H\mu_r,
\qquad
m_j \triangleq H\mu_j.
\]
Using the formula for \(\gamma_{r|y}(y)\), we obtain:
\begin{align}
L_{rj}^{(y)}(y)
&= \log \frac{\pi_r \mathcal{N}(y; m_r, S_y)}{\pi_j \mathcal{N}(y; m_j, S_y)} \nonumber \\
&= \log \frac{\pi_r \exp\left(-\frac{1}{2} (y-m_r)^\top S_y^{-1} (y-m_r)\right)}{\pi_j \exp\left(-\frac{1}{2} (y-m_j)^\top S_y^{-1} (y-m_j)\right)} \nonumber \\
&= \log\frac{\pi_r}{\pi_j} - \frac{1}{2} (y-m_r)^\top S_y^{-1} (y-m_r) + \frac{1}{2} (y-m_j)^\top S_y^{-1} (y-m_j).
\end{align}

Expanding the quadratic forms and using the symmetry of $S_y^{-1}$ (which implies $y^\top S_y^{-1} m_r = m_r^\top S_y^{-1} y$), the $y^\top S_y^{-1} y$ terms cancel out:
\begin{align}
L_{rj}^{(y)}(y)
&= \log\frac{\pi_r}{\pi_j} - \frac{1}{2} \left( -2 m_r^\top S_y^{-1} y + m_r^\top S_y^{-1} m_r \right) + \frac{1}{2} \left( -2 m_j^\top S_y^{-1} y + m_j^\top S_y^{-1} m_j \right) \nonumber \\
&= \log\frac{\pi_r}{\pi_j} + (m_r - m_j)^\top S_y^{-1} y - \frac{1}{2} m_r^\top S_y^{-1} m_r + \frac{1}{2} m_j^\top S_y^{-1} m_j.
\end{align}

Finally, we can factor the remaining terms by noting that $(m_r-m_j)^\top S_y^{-1} (m_r+m_j) = m_r^\top S_y^{-1} m_r - m_j^\top S_y^{-1} m_j$ because the cross-terms cancel. This yields:
\begin{equation}
L_{rj}^{(y)}(y)
=
\log\frac{\pi_r}{\pi_j}
+
(m_r-m_j)^\top S_y^{-1}
\left(
y-\frac12(m_r+m_j)
\right).
\label{eq:mog_measurement_logratio_expanded_problem_basis}
\end{equation}
If \(C=r\), then \(y=m_r+S_y^{1/2}\xi\) with \(\xi\sim\mathcal N(0,I)\). Substituting into
\eqref{eq:mog_measurement_logratio_expanded_problem_basis} gives
\[
L_{rj}^{(y)}(y)
=
\log\frac{\pi_r}{\pi_j}
+
\frac12(m_r-m_j)^\top S_y^{-1}(m_r-m_j)
+
\bigl(S_y^{-1/2}(m_r-m_j)\bigr)^\top \xi.
\]
The last term is Gaussian with mean \(0\) and variance
\((m_r-m_j)^\top S_y^{-1}(m_r-m_j)=M_{rj}\). This proves
\eqref{eq:mog_measurement_logratio_law_problem_basis}. 

\paragraph{Lemma 2}
For \(r\neq j\) and each step \(s\), define
\begin{equation}
Q_{rj,s}(x_s,y)
\triangleq
\log\frac{p(x_s\mid y,C=r)}{p(x_s\mid y,C=j)}.
\label{eq:mog_bridge_logratio_def_problem_basis}
\end{equation}
Then, conditioned on \((y,C=r)\),
\begin{equation}
Q_{rj,s}(x_s,y)
\sim
\mathcal N\!\left(
\frac12 B_{rj,s},
\;
B_{rj,s}
\right).
\label{eq:mog_bridge_logratio_law_problem_basis}
\end{equation}

\paragraph{Proof.}
From \eqref{eq:mog_bridge_law_shared_problem_basis},
\[
x_s\mid (y,C=r)\sim \mathcal N(u_{r,s}(y),V_s),
\qquad
x_s\mid (y,C=j)\sim \mathcal N(u_{j,s}(y),V_s).
\]
Since both Gaussians have the same covariance \(V_s\), using similar steps to Lemma 1,
\begin{equation}
Q_{rj,s}(x_s,y)
=
\bigl(u_{r,s}(y)-u_{j,s}(y)\bigr)^\top
V_s^{-1}
\left(
x_s-\frac12(u_{r,s}(y)+u_{j,s}(y))
\right).
\label{eq:mog_bridge_logratio_expanded_problem_basis}
\end{equation}
If \(C=r\), then
\[
x_s
=
u_{r,s}(y)+V_s^{1/2}\zeta_s,
\qquad
\zeta_s\sim\mathcal N(0,I).
\]
Substituting into \eqref{eq:mog_bridge_logratio_expanded_problem_basis} yields
\[
Q_{rj,s}(x_s,y)
=
\frac12
\bigl(u_{r,s}(y)-u_{j,s}(y)\bigr)^\top
V_s^{-1}
\bigl(u_{r,s}(y)-u_{j,s}(y)\bigr)
+
\bigl(V_s^{-1/2}(u_{r,s}(y)-u_{j,s}(y))\bigr)^\top \zeta_s.
\]
The first term is \(\frac12 B_{rj,s}\) and the second term is centered Gaussian
with variance \(B_{rj,s}\). This proves
\eqref{eq:mog_bridge_logratio_law_problem_basis}. 

\paragraph{Lemma 3 }
For \(r\neq j\), define
\begin{equation}
L_{rj,s}(x_s,y)
\triangleq
\log\frac{\gamma_{r|s}(x_s,y)}{\gamma_{j|s}(x_s,y)}.
\label{eq:mog_stepwise_logratio_def_problem_basis}
\end{equation}
Then
\begin{equation}
L_{rj,s}(x_s,y)
=
L_{rj}^{(y)}(y)+Q_{rj,s}(x_s,y).
\label{eq:mog_stepwise_logratio_decomposition_problem_basis}
\end{equation}

\paragraph{Proof.}
By Bayes' rule,
\[
\gamma_{r|s}(x_s,y)\propto \gamma_{r|y}(y)\,p(x_s\mid y,C=r).
\]
Taking the ratio between components \(r\) and \(j\) gives
\[
\frac{\gamma_{r|s}(x_s,y)}{\gamma_{j|s}(x_s,y)}
=
\frac{\gamma_{r|y}(y)}{\gamma_{j|y}(y)}
\cdot
\frac{p(x_s\mid y,C=r)}{p(x_s\mid y,C=j)}.
\]
Taking logarithms yields
\eqref{eq:mog_stepwise_logratio_decomposition_problem_basis}. 

\subsection*{G.5 Turning pairwise margins into one-hot responsibilities}

The next elementary lemma converts a lower bound on all pairwise log-ratios
into a lower bound on the winning responsibility.

\paragraph{Lemma 4 }
Suppose that for some index \(r\) and some threshold \(T\ge 0\),
\begin{equation}
\log\frac{\gamma_r}{\gamma_j}\ge T
\qquad \text{for all } j\neq r.
\label{eq:mog_margin_assumption_problem_basis}
\end{equation}
Then
\begin{equation}
\gamma_r
\ge
\frac{1}{1+(R-1)e^{-T}}.
\label{eq:mog_margin_to_onehot_problem_basis}
\end{equation}

\paragraph{Proof.}
From \eqref{eq:mog_margin_assumption_problem_basis},
\[
\gamma_j\le e^{-T}\gamma_r
\qquad \text{for all } j\neq r.
\]
Therefore,
\[
1
=
\gamma_r+\sum_{j\neq r}\gamma_j
\le
\gamma_r\bigl(1+(R-1)e^{-T}\bigr),
\]
which implies \eqref{eq:mog_margin_to_onehot_problem_basis}. 

\subsection*{G.6 Main theorem: the same component stays dominant across all noise levels}

We now state the main theorem. Its role is to make the concentration mechanism completely explicit: once the measurement term is large enough, the bridge fluctuations are not strong enough to overturn the winning component.

\paragraph{Theorem 1}
Assume the shared-covariance model
\eqref{ass:shared_covariance_main}. Fix a true component
\(r^\star\) and suppose that the sample is generated under \(C=r^\star\).

Let \(T_0>T_1>0\), and define the event
\begin{equation}
\mathcal E(T_0,T_1)
\triangleq
\left\{
\begin{array}{l}
L_{r^\star j}^{(y)}(y)\ge T_0
\quad \text{for all } j\neq r^\star, \\[4pt]
L_{r^\star j,s}(x_s,y)\ge T_1
\quad \text{for all } j\neq r^\star
\text{ and all } s=1,\dots,S-1
\end{array}
\right\}.
\label{eq:mog_good_event_problem_basis}
\end{equation}
Then
\begin{equation}
\mathbb P\!\bigl(\mathcal E(T_0,T_1)^c \mid C=r^\star\bigr) \le \sum_{j\neq r^\star} \exp\!\left( -\frac{\left[\log\frac{\pi_{r^\star}}{\pi_j}+\frac12 M_{r^\star j}-T_0\right]_+^2}{2M_{r^\star j}} \right) + (S-1)(R-1)e^{-(T_0-T_1)}.
\label{eq:mog_good_event_bound_problem_basis}
\end{equation}

Moreover, on the event \(\mathcal E(T_0,T_1)\), the following hold
simultaneously:

\begin{enumerate}
    \item the measurement posterior is concentrated on the true component:
    \begin{equation}
    \gamma_{r^\star|y}(y)
    \ge
    \frac{1}{1+(R-1)e^{-T_0}};
    \label{eq:mog_measurement_onehot_conclusion_problem_basis}
    \end{equation}

    \item for every noise level \(s=1,\dots,S-1\), the stepwise
    responsibilities are concentrated on the same component:
    \begin{equation}
    \gamma_{r^\star|s}(x_s,y)
    \ge
    \frac{1}{1+(R-1)e^{-T_1}};
    \label{eq:mog_stepwise_onehot_conclusion_problem_basis}
    \end{equation}

    \item the same component \(r^\star\) is strictly dominant for all
    \(s=1,\dots,S-1\).
\end{enumerate}

\paragraph{Proof.}
We split the proof into two parts.

\medskip
\textit{Part 1: concentration of the measurement posterior.}

Define
\begin{equation}
\mathcal E_y(T_0)
\triangleq
\left\{
L_{r^\star j}^{(y)}(y)\ge T_0
\quad \text{for all } j\neq r^\star
\right\}.
\label{eq:mog_measurement_event_problem_basis}
\end{equation}
By Lemma 1, for each \(j\neq r^\star\),
\[
L_{r^\star j}^{(y)}(y)\mid (C=r^\star)
\sim
\mathcal N\!\left(
\log\frac{\pi_{r^\star}}{\pi_j}+\frac12 M_{r^\star j},
\;
M_{r^\star j}
\right).
\]
Let
\[
\Delta_{r^\star j}(T_0)
\triangleq
\left[\log\frac{\pi_{r^\star}}{\pi_j}+\frac12 M_{r^\star j}-T_0\right]_+.
\]
If \(\Delta_{r^\star j}(T_0)=0\), then the trivial bound
\[
\mathbb P\!\left(
L_{r^\star j}^{(y)}(y)<T_0
\mid C=r^\star
\right)\le 1
\]
already yields
\[
\mathbb P\!\left(
L_{r^\star j}^{(y)}(y)<T_0
\mid C=r^\star
\right)
\le
\exp\!\left(-\frac{\Delta_{r^\star j}(T_0)^2}{2M_{r^\star j}}\right).
\]
If instead \(\Delta_{r^\star j}(T_0)>0\), then \(T_0\) lies below the Gaussian
mean, and the standard Gaussian tail bound \(\Phi(-u)\le e^{-u^2/2}\) gives
\[
\mathbb P\!\left(
L_{r^\star j}^{(y)}(y)<T_0
\mid C=r^\star
\right)
\le
\exp\!\left(
-\frac{\Delta_{r^\star j}(T_0)^2}{2M_{r^\star j}}
\right).
\]
Applying a union bound over all \(j\neq r^\star\) gives
\begin{align}
\mathbb P\!\bigl(\mathcal E_y(T_0)^c \mid C=r^\star\bigr)
\le
\sum_{j\neq r^\star}
\exp\!\left(
-\frac{\left[\log\frac{\pi_{r^\star}}{\pi_j}+\frac12 M_{r^\star j}-T_0\right]_+^2}{2M_{r^\star j}}
\right).
\label{eq:mog_measurement_event_bound_problem_basis}
\end{align}

On the event \(\mathcal E_y(T_0)\), we have
\[
\log\frac{\gamma_{r^\star|y}(y)}{\gamma_{j|y}(y)}\ge T_0
\qquad \text{for all } j\neq r^\star,
\]
so Lemma 4 implies
\eqref{eq:mog_measurement_onehot_conclusion_problem_basis}.

\medskip
\textit{Part 2: stability across all noise levels.}

Fix a pair \((j,s)\) with \(j\neq r^\star\) and
\(s\in\{1,\dots,S-1\}\). On the event \(\mathcal E_y(T_0)\), if
\[
L_{r^\star j,s}(x_s,y)<T_1,
\]
then by Lemma 3,
\[
Q_{r^\star j,s}(x_s,y)
=
L_{r^\star j,s}(x_s,y)-L_{r^\star j}^{(y)}(y)
<
-(T_0-T_1).
\]
Therefore,
\begin{equation}
\mathbb P\!\left(
L_{r^\star j,s}(x_s,y)<T_1
\mid C=r^\star,\mathcal E_y(T_0),y
\right)
\le
\mathbb P\!\left(
Q_{r^\star j,s}(x_s,y)<-(T_0-T_1)
\mid C=r^\star,y
\right).
\label{eq:mog_bridge_bad_to_Q_bad_problem_basis}
\end{equation}

By Lemma 2,
\[
Q_{r^\star j,s}(x_s,y)\mid (C=r^\star,y)
\sim
\mathcal N\!\left(\frac12 B_{r^\star j,s},\,B_{r^\star j,s}\right).
\]
If \(B_{r^\star j,s}=0\), then \(Q_{r^\star j,s}(x_s,y)=0\) almost surely, so
the probability on the right-hand side of
\eqref{eq:mog_bridge_bad_to_Q_bad_problem_basis} is \(0\), and the desired bound
is trivial. Otherwise, \(B_{r^\star j,s}>0\), and
\begin{equation}
\mathbb P\!\left(
Q_{r^\star j,s}(x_s,y)<-(T_0-T_1)
\mid C=r^\star,y
\right)
\le
\exp\!\left(
-\frac{\left(T_0-T_1+\frac12 B_{r^\star j,s}\right)^2}{2B_{r^\star j,s}}
\right).
\label{eq:mog_bridge_tail_exact_problem_basis}
\end{equation}
Set \(a\triangleq T_0-T_1>0\) and \(B\triangleq B_{r^\star j,s}\). Then
\[
\frac{(a+\frac12 B)^2}{2B}
=
\frac{a^2}{2B}+\frac{a}{2}+\frac{B}{8}
\ge a,
\]
because
\[
\frac{a^2}{2B}+\frac{B}{8}\ge \frac{a}{2}
\]
by the arithmetic-geometric mean inequality. Therefore
\begin{equation}
\mathbb P\!\left(
L_{r^\star j,s}(x_s,y)<T_1
\mid C=r^\star,\mathcal E_y(T_0),y
\right)
\le
e^{-(T_0-T_1)}.
\label{eq:mog_bridge_tail_simple_problem_basis}
\end{equation}

This bound is uniform in \(y\), so after averaging over \(y\) it remains valid:
\begin{equation}
\mathbb P\!\left(
L_{r^\star j,s}(x_s,y)<T_1
\mid C=r^\star,\mathcal E_y(T_0)
\right)
\le
e^{-(T_0-T_1)}.
\label{eq:mog_bridge_tail_uniform_problem_basis}
\end{equation}

Applying a union bound over all \(j\neq r^\star\) and all
\(s=1,\dots,S-1\),
\begin{align}
&\mathbb P\!\left(
\exists\,j\neq r^\star,\exists\,s\in\{1,\dots,S-1\}
:
L_{r^\star j,s}(x_s,y)<T_1
\mid C=r^\star,\mathcal E_y(T_0)
\right)
\notag\\
&\qquad\le
(S-1)(R-1)e^{-(T_0-T_1)}.
\label{eq:mog_bridge_union_bound_problem_basis}
\end{align}

Combining \eqref{eq:mog_measurement_event_bound_problem_basis} and
\eqref{eq:mog_bridge_union_bound_problem_basis} proves
\eqref{eq:mog_good_event_bound_problem_basis}.

Finally, on the event \(\mathcal E(T_0,T_1)\), we have
\[
\log\frac{\gamma_{r^\star|s}(x_s,y)}{\gamma_{j|s}(x_s,y)}\ge T_1
\qquad
\text{for all } j\neq r^\star \text{ and all } s=1,\dots,S-1.
\]
Applying Lemma 4 at each noise level yields
\eqref{eq:mog_stepwise_onehot_conclusion_problem_basis}. Since \(T_1>0\), these
pairwise log-ratios are strictly positive, hence
\(\gamma_{r^\star|s}(x_s,y)>\gamma_{j|s}(x_s,y)\) for every \(j\neq r^\star\)
and every \(s=1,\dots,S-1\), so the same component is strictly dominant across
all noise levels. 

\subsection*{G.7 A high-dimensional corollary}

The following corollary reformulates the previous theorem into the statement used in the main text: when the measurement separation is proportional to the dimension, the failure probability decays exponentially in $d$.

\paragraph{Corollary 1}
Assume the setting of Theorem~1, and suppose that there exists a constant
\(\alpha>0\) such that
\begin{equation}
M_{rj}\ge \alpha d
\qquad \text{for all } r\neq j.
\label{eq:mog_linear_measurement_separation_problem_basis}
\end{equation}
Assume also that the mixture weights are uniformly bounded below:
\begin{equation}
\pi_r \ge \pi_{\min}>0
\qquad \text{for all } r.
\label{eq:mog_pi_min_problem_basis}
\end{equation}

Fix \(R\), \(S\), and \(\varepsilon\in(0,1/2)\), and define
\begin{equation}
T_\varepsilon
\triangleq
\log\frac{(R-1)(1-\varepsilon)}{\varepsilon}.
\label{eq:mog_T_epsilon_problem_basis}
\end{equation}
Then there exist constants \(c>0\), \(K>0\), and \(d_0\in\mathbb N\) depending
only on \(\alpha\), \(\pi_{\min}\), \(\varepsilon\), and \(R\), such that for
every \(d\ge d_0\),
\begin{equation}
\begin{aligned}
&\mathbb P \Bigl( \gamma_{r^\star|y}(y) \ge 1-\varepsilon, \, \gamma_{r^\star|s}(x_s,y) \ge 1-\varepsilon, \text{ and } r^\star \text{ is dominant } \forall s=1,\dots,S-1 \Bigm| C=r^\star \Bigr) \\
&\qquad \ge 1 - K S e^{-c d}.
\end{aligned}
\label{eq:mog_highdim_corollary_problem_basis}
\end{equation}

\paragraph{Proof.}
Choose
\begin{equation}
T_1 = T_\varepsilon,
\qquad
T_0 = \frac{\alpha d}{4}.
\label{eq:mog_T0_T1_choice_problem_basis}
\end{equation}
By Lemma 4, the condition
\[
\log\frac{\gamma_{r^\star|s}}{\gamma_{j|s}}\ge T_\varepsilon
\quad \forall j\neq r^\star
\]
implies
\[
\gamma_{r^\star|s}\ge \frac{1}{1+(R-1)e^{-T_\varepsilon}} = 1-\varepsilon.
\]
The same implication also holds for \(\gamma_{r^\star|y}(y)\).

It therefore suffices to control the failure probability in
\eqref{eq:mog_good_event_bound_problem_basis}. Since \(M_{r^\star j}\ge \alpha d\)
and \(\pi_{r^\star}/\pi_j \ge \pi_{\min}\), for sufficiently large \(d\) we have
\[
\log\frac{\pi_{r^\star}}{\pi_j}+\frac12 M_{r^\star j}-T_0
\ge
\log \pi_{\min} + \frac14 M_{r^\star j}
\ge
\frac18 M_{r^\star j}.
\]
Therefore the first term in
\eqref{eq:mog_good_event_bound_problem_basis} is at most
\[
\sum_{j\neq r^\star}
\exp\!\left(-\frac{M_{r^\star j}}{128}\right)
\le
(R-1)e^{-\alpha d/128}.
\]
Similarly, for sufficiently large \(d\),
\[
T_0-T_1
=
\frac{\alpha d}{4} - T_\varepsilon
\ge
\frac{\alpha d}{8},
\]
so the second term in \eqref{eq:mog_good_event_bound_problem_basis} is at most
\[
(S-1)(R-1)e^{-\alpha d/8}.
\]
Combining the two bounds yields
\[
\mathbb P\!\bigl(\mathcal E(T_0,T_1)^c\mid C=r^\star\bigr)
\le
(R-1)e^{-\alpha d/128}+(S-1)(R-1)e^{-\alpha d/8},
\]
which implies \eqref{eq:mog_highdim_corollary_problem_basis} for suitable
constants \(c,K,d_0\). 

\subsection*{G.8 Random means imply linear measurement separation}

The high-dimensional corollary is driven by the pairwise measurement separation

\[
M_{rj}
=
(\mu_r-\mu_j)^\top A(\mu_r-\mu_j).
\]

\[
A
=
H^\top(H\Sigma H^\top+\sigma_y^2 I)^{-1}H.
\]
Since \(A\) is symmetric positive semidefinite, we can write its eigendecomposition as
\[
A=\sum_{k=1}^{d}\eta_k v_kv_k^\top,
\qquad
\eta_k\ge 0.
\]
Assume that \(A\) has a linear number of informative directions: there exists
\(I\subset\{1,\dots,d\}\) such that
\begin{equation}
|I|\ge \rho d
\qquad\text{and}\qquad
\eta_k\ge \eta_{\min}>0
\quad\text{for all }k\in I.
\label{eq:mog_many_informative_eigenvalues}
\end{equation}

Then
\[
A\succeq \eta_{\min}\sum_{k\in I}v_kv_k^\top.
\]

\paragraph{Proposition 1}
Suppose the component means are drawn independently in a fixed orthonormal
coordinate system:
\[
(\mu_r)_\ell
\overset{\mathrm{i.i.d.}}{\sim}
\mathrm{Unif}[-1,1],
\qquad
r=1,\dots,R,\quad \ell=1,\dots,d.
\]
Assume \eqref{eq:mog_many_informative_eigenvalues}. Then for every fixed pair
\(r\neq j\), there exist constants \(\alpha,c>0\), depending only on
\(\rho\) and \(\eta_{\min}\), such that for all sufficiently large \(d\),
\begin{equation}
\mathbb P(M_{rj}\le \alpha d)\le e^{-cd}.
\label{eq:mog_random_means_fixed_basis_linear_separation}
\end{equation}

\paragraph{Proof.}
Fix a pair \(r\neq j\), and define
\[
\Delta_{rj}\triangleq \mu_r-\mu_j.
\]
Since the coordinates of \(\mu_r\) and \(\mu_j\) are drawn independently from
\(\mathrm{Unif}[-1,1]\), the coordinates of \(\Delta_{rj}\) are independent.
For each coordinate \(\ell\),
\[
(\Delta_{rj})_\ell=(\mu_r)_\ell-(\mu_j)_\ell.
\]
Thus \((\Delta_{rj})_\ell\) is centered and bounded by \(2\) in absolute value.
Moreover,
\[
\begin{aligned}
\mathbb E[(\Delta_{rj})_\ell^2]
&=
\mathbb E[(U-V)^2]
\\
&=
\mathbb E[U^2]+\mathbb E[V^2]-2\mathbb E[U]\mathbb E[V]
\\
&=
\frac13+\frac13
=
\frac23,
\end{aligned}
\]
where
\[
U,V\overset{\mathrm{i.i.d.}}{\sim}\mathrm{Unif}[-1,1].
\]
Therefore
\[
\mathbb E[\Delta_{rj}]=0,
\qquad
\operatorname{Cov}(\Delta_{rj})=\frac23 I.
\]

Let
\[
P_I=\sum_{k\in I}v_kv_k^\top.
\]
We first record the elementary properties of \(P_I\) directly from this formula.
Since the vectors \(v_1,\dots,v_d\) form an orthonormal basis,
\[
v_k^\top v_\ell=
\begin{cases}
1, & k=\ell,\\
0, & k\neq \ell.
\end{cases}
\]

First,
\[
P_I^\top
=
\left(\sum_{k\in I}v_kv_k^\top\right)^\top
=
\sum_{k\in I}v_kv_k^\top
=
P_I.
\]
Second,
\[
\begin{aligned}
P_I^2
&=
\left(\sum_{k\in I}v_kv_k^\top\right)
\left(\sum_{\ell\in I}v_\ell v_\ell^\top\right)
\\
&=
\sum_{k\in I}\sum_{\ell\in I}
v_k(v_k^\top v_\ell)v_\ell^\top
\\
&=
\sum_{k\in I}v_kv_k^\top
=
P_I.
\end{aligned}
\]
Thus \(P_I\) is symmetric and satisfies \(P_I^2=P_I\).

Since
\[
A=\sum_{k=1}^{d}\eta_k v_kv_k^\top,
\]
we have, for every vector \(u\),
\[
\begin{aligned}
u^\top(A-\eta_{\min}P_I)u
&=
\sum_{k\in I}(\eta_k-\eta_{\min})(v_k^\top u)^2
+
\sum_{k\notin I}\eta_k(v_k^\top u)^2.
\end{aligned}
\]
Every term on the right-hand side is nonnegative, because
\(\eta_k\ge \eta_{\min}\) for \(k\in I\) and \(\eta_k\ge0\) for all \(k\).
Therefore
\[
A\succeq \eta_{\min}P_I.
\]
Applying this with \(u=\Delta_{rj}\) gives
\begin{equation}
M_{rj}
=
\Delta_{rj}^\top A\Delta_{rj}
\ge
\eta_{\min}\Delta_{rj}^\top P_I\Delta_{rj}.
\label{eq:mog_random_means_projection_lower_bound}
\end{equation}

We now study
\[
Q\triangleq \Delta_{rj}^\top P_I\Delta_{rj}.
\]
Using \(P_I^\top=P_I\) and \(P_I^2=P_I\),
\[
\|P_I\Delta_{rj}\|_2^2
=
(P_I\Delta_{rj})^\top(P_I\Delta_{rj})
=
\Delta_{rj}^\top P_I^\top P_I\Delta_{rj}
=
\Delta_{rj}^\top P_I^2\Delta_{rj}
=
\Delta_{rj}^\top P_I\Delta_{rj}
=
Q.
\]

Next, compute the expectation of \(Q\). For any deterministic matrix \(B\) and
any centered random vector \(X\),
\[
X^\top B X
=
\operatorname{tr}(X^\top B X)
=
\operatorname{tr}(BXX^\top).
\]
Therefore
\[
\mathbb E[X^\top B X]
=
\operatorname{tr}\!\left(B\,\mathbb E[XX^\top]\right)
=
\operatorname{tr}\!\left(B\operatorname{Cov}(X)\right).
\]
Using this with \(X=\Delta_{rj}\) and \(B=P_I\), we get
\[
\begin{aligned}
\mathbb E[Q]
&=
\operatorname{tr}\!\left(P_I\operatorname{Cov}(\Delta_{rj})\right)
\\
&=
\operatorname{tr}\!\left(P_I\frac23 I\right)
\\
&=
\frac23\operatorname{tr}(P_I).
\end{aligned}
\]
Now
\[
\operatorname{tr}(P_I)
=
\operatorname{tr}\!\left(\sum_{k\in I}v_kv_k^\top\right)
=
\sum_{k\in I}\operatorname{tr}(v_kv_k^\top).
\]
For each \(k\),
\[
\operatorname{tr}(v_kv_k^\top)=v_k^\top v_k=1.
\]
Hence
\[
\operatorname{tr}(P_I)=|I|,
\]
and therefore
\begin{equation}
\mathbb E[Q]=\frac23 |I|.
\label{eq:mog_random_means_projection_expectation}
\end{equation}

We now control the probability that \(Q\) is much smaller than its mean. We use
the Hanson--Wright inequality ~\citep{rudelson2013hanson} in the following standard form: if \(X\) has
independent, centered, bounded coordinates and \(B\) is deterministic, then
there exists a universal constant \(c_{\mathrm{HW}}>0\) such that, for every
\(t>0\),
\begin{equation}
\mathbb P\!\left(
|X^\top B X-\mathbb E[X^\top B X]|\ge t
\right)
\le
2\exp\!\left[
-c_{\mathrm{HW}}
\min\left(
\frac{t^2}{\|B\|_F^2},
\frac{t}{\|B\|_{\mathrm{op}}}
\right)
\right].
\label{eq:hw_inequality_used}
\end{equation}
Here
\[
\|B\|_{\mathrm{op}}
=
\sup_{\|x\|_2=1}\|Bx\|_2
\]
and
\[
\|B\|_F^2
=
\operatorname{tr}(B^\top B).
\]

We apply this with
\[
X=\Delta_{rj},
\qquad
B=P_I.
\]
The assumptions apply because the coordinates of \(\Delta_{rj}\) are
independent, centered, and bounded by \(2\).

We now compute the two norms of \(P_I\). First,
\[
\|P_I\|_F^2
=
\operatorname{tr}(P_I^\top P_I)
=
\operatorname{tr}(P_I^2)
=
\operatorname{tr}(P_I)
=
|I|.
\]
Second, for any vector \(x\),
\[
P_Ix
=
\sum_{k\in I}v_k(v_k^\top x).
\]
Using orthonormality,
\[
\|P_Ix\|_2^2
=
\sum_{k\in I}(v_k^\top x)^2
\le
\sum_{k=1}^{d}(v_k^\top x)^2
=
\|x\|_2^2.
\]
Thus, for every \(x\) with \(\|x\|_2=1\),
\[
\|P_Ix\|_2\le1,
\]
so
\[
\|P_I\|_{\mathrm{op}}\le1.
\]
If \(I\) is nonempty, choosing \(x=v_k\) for some \(k\in I\) gives
\(P_Ix=x\), so in fact \(\|P_I\|_{\mathrm{op}}=1\). 

By \eqref{eq:mog_random_means_projection_expectation},
\[
\mathbb E[Q]=\frac23|I|.
\]
The event
\[
Q\le \frac13|I|
\]
implies
\[
\mathbb E[Q]-Q
\ge
\frac23|I|-\frac13|I|
=
\frac13|I|.
\]
Therefore
\[
\left\{
Q\le \frac13|I|
\right\}
\subseteq
\left\{
|Q-\mathbb E[Q]|\ge \frac13|I|
\right\}.
\]

Using \eqref{eq:hw_inequality_used} with \(t=|I|/3\), \(B=P_I\), and the norm
bounds above,
\[
\begin{aligned}
\mathbb P\!\left(
Q\le \frac13|I|
\right)
&\le
\mathbb P\!\left(
|Q-\mathbb E[Q]|\ge \frac13|I|
\right)
\\
&\le
2\exp\!\left[
-c_{\mathrm{HW}}
\min\left(
\frac{(|I|/3)^2}{|I|},
\frac{|I|/3}{1}
\right)
\right]
\\
&=
2\exp\!\left[
-c_{\mathrm{HW}}
\min\left(
\frac{|I|}{9},
\frac{|I|}{3}
\right)
\right].
\end{aligned}
\]
Absorbing the numerical constants into a new universal constant \(c_0>0\), we
obtain
\begin{equation}
\mathbb P\!\left(
Q\le \frac13|I|
\right)
\le
2e^{-c_0|I|}.
\label{eq:mog_random_means_projection_concentration}
\end{equation}

Since \(|I|\ge \rho d\), the event
\[
Q\le \frac{\rho d}{3}
\]
is contained in the event
\[
Q\le \frac{|I|}{3}.
\]
Therefore
\[
\mathbb P\!\left(
Q\le \frac{\rho d}{3}
\right)
\le
2e^{-c_0|I|}
\le
2e^{-c_0\rho d}.
\]

Finally, by \eqref{eq:mog_random_means_projection_lower_bound},
\[
M_{rj}\ge \eta_{\min}Q.
\]
Hence, if
\[
M_{rj}\le \frac{\rho\eta_{\min}}{3}d,
\]
then necessarily
\[
Q\le \frac{\rho d}{3}.
\]
Thus
\[
\mathbb P\!\left(
M_{rj}\le \frac{\rho\eta_{\min}}{3}d
\right)
\le
\mathbb P\!\left(
Q\le \frac{\rho d}{3}
\right)
\le
2e^{-c_0\rho d}.
\]
For sufficiently large \(d\), the prefactor \(2\) can be absorbed into the
exponential by reducing the exponent constant. Thus the claim holds with
\[
\alpha=\frac{\rho\eta_{\min}}{3}.
\]

\paragraph{All-pairs consequence.}
If \(R\) is fixed, then a union bound gives
\begin{equation}
\mathbb P\!\left(
\min_{r\neq j}M_{rj}\le \alpha d
\right)
\le
\binom{R}{2}e^{-cd}.
\label{eq:mog_random_means_all_pairs_fixed_basis}
\end{equation}
Hence, for a fixed number of components \( R \), the constant prefactor \( \binom{R}{2} \) is asymptotically dominated by the exponential decay in \( d \), guaranteeing that the linear-separation condition holds with overwhelming probability in high dimensions.

\subsection*{G.9 Transfer to the selected-label chain and local perturbation}

We now transfer the bridge-marginal concentration bound to the selected-label
chain. 

\paragraph{Selected-chain transfer.}
Fix the true component \(r^\star\), condition on \(y\), and run the
selected-label chain with the label forced to \(J=r^\star\). The proof of
Proposition~\ref{prop:selected_label_mean_exactness} gives, at every step,
\begin{equation}
\mathbb E[X_s^{\mathrm{sel}}\mid y,J=r^\star]
=
u_{r^\star,s}(y).
\label{eq:selected_transfer_same_mean_short}
\end{equation}
Similarly, the covariance recursion used in the proof of
Proposition~\ref{prop:selected_label_covariance_deficit} gives the
intermediate-time domination
\begin{equation}
\operatorname{Cov}(X_s^{\mathrm{sel}}\mid y,J=r^\star)
\preceq
V_s,
\qquad
s=1,\dots,S-1.
\label{eq:selected_transfer_cov_domination_short}
\end{equation}
Thus, relative to the component-conditioned bridge marginal
\[
x_s\mid(y,C=r^\star)\sim
\mathcal N(u_{r^\star,s}(y),V_s),
\]
the selected chain has the same mean and no larger covariance.

Now fix \(j\neq r^\star\) and \(s\in\{1,\dots,S-1\}\). By
\eqref{eq:mog_bridge_logratio_expanded_problem_basis}, the pairwise bridge
log-likelihood ratio
\[
Q_{r^\star j,s}(x,y)
=
\log\frac{p(x\mid y,C=r^\star)}{p(x\mid y,C=j)}
\]
is affine in \(x\). Therefore, applying this same affine functional to
\(X_s^{\mathrm{sel}}\), equations
\eqref{eq:selected_transfer_same_mean_short}--\eqref{eq:selected_transfer_cov_domination_short}
imply that
\begin{equation}
\mathbb E[
Q_{r^\star j,s}(X_s^{\mathrm{sel}},y)
\mid y,J=r^\star]
=
\frac12 B_{r^\star j,s},
\label{eq:selected_Q_same_mean_short}
\end{equation}
and
\begin{equation}
\operatorname{Var}[
Q_{r^\star j,s}(X_s^{\mathrm{sel}},y)
\mid y,J=r^\star]
\le
B_{r^\star j,s}.
\label{eq:selected_Q_smaller_var_short}
\end{equation}
Since the
selected-chain margin has the same mean and smaller variance, the Gaussian tail
bound used in Part 2 of Theorem~1 applies with \(x_s\) replaced by
\(X_s^{\mathrm{sel}}\). Hence, on the event
\[
L_{r^\star j}^{(y)}(y)\ge T_0
\qquad\text{for all }j\neq r^\star,
\]
we get, exactly as in \eqref{eq:mog_bridge_tail_simple_problem_basis},
\begin{equation}
\mathbb P\!\left(
L_{r^\star j,s}(X_s^{\mathrm{sel}},y)<T_1
\mid C=r^\star,\mathcal E_y(T_0),y
\right)
\le
e^{-(T_0-T_1)} .
\label{eq:selected_transfer_tail_short}
\end{equation}
A union bound over \(j\neq r^\star\) and \(s=1,\dots,S-1\), together with the
measurement part of Theorem~1, gives the selected-chain analogue of
\eqref{eq:mog_good_event_bound_problem_basis}. Choosing
\[
T_1=T_\varepsilon,
\qquad
T_0=\frac{\alpha d}{4},
\]
as in Corollary~1 yields
\begin{equation}
\mathbb P\!\left(
\mathcal E^{\mathrm{sel}}_{r^\star,\varepsilon}
\mid C=r^\star
\right)
\ge
1-KSe^{-cd}.
\end{equation}
which proves Theorem~\ref{thm:selected_chain_self_consistency}.

It remains to justify the sampled-label statement. Since
\(J\sim\gamma_{\cdot\mid y}(y)\),
\[
\mathbb P(J\neq r^\star\mid C=r^\star)
=
\mathbb E[
1-\gamma_{r^\star\mid y}(y)
\mid C=r^\star].
\]
On the event \(\gamma_{r^\star\mid y}(y)\ge1-\varepsilon\), the integrand is at
most \(\varepsilon\), and otherwise it is at most \(1\). Therefore,
\[
\mathbb P(J\neq r^\star\mid C=r^\star)
\le
\varepsilon
+
\mathbb P(
\gamma_{r^\star\mid y}(y)<1-\varepsilon
\mid C=r^\star)
\le
\varepsilon+KSe^{-cd}.
\]
Equivalently,
\[
\mathbb P(J=r^\star\mid C=r^\star)
\ge
1-\varepsilon-KSe^{-cd}.
\]

\paragraph{Local exact-versus-frozen perturbation bound.}
Finally, we record the deterministic one-step consequence of high
responsibility. Fix a component \(r\), a step \(s\), and a state \((x,y)\).
If
\[
\gamma_{r\mid s}(x,y)\ge 1-\varepsilon,
\]
then
\begin{align}
\left\|
\hat x_{0,s}^{\mathrm{MMSE}}(x,y)-\mu_{r\mid s}(x,y)
\right\|_2
&=
\left\|
\sum_{j\neq r}
\gamma_{j\mid s}(x,y)
\bigl(\mu_{j\mid s}(x,y)-\mu_{r\mid s}(x,y)\bigr)
\right\|_2
\notag\\
&\le
\varepsilon
\max_{j\neq r}
\left\|
\mu_{j\mid s}(x,y)-\mu_{r\mid s}(x,y)
\right\|_2 .
\label{eq:local_denoiser_perturbation_short}
\end{align}
Coupling the exact MoG update and the frozen-label update with the same
Gaussian innovation \(z_s\), the shared terms \(b_sy+c_sx+\sigma_sz_s\) cancel,
and therefore
\begin{align}
&\left\|
\bigl(a_s\hat x_{0,s}^{\mathrm{MMSE}}(x,y)+b_sy+c_sx+\sigma_sz_s\bigr)
-
\bigl(a_s\mu_{r\mid s}(x,y)+b_sy+c_sx+\sigma_sz_s\bigr)
\right\|_2
\notag\\
&\qquad\le
|a_s|\varepsilon
\max_{j\neq r}
\left\|
\mu_{j\mid s}(x,y)-\mu_{r\mid s}(x,y)
\right\|_2 .
\label{eq:local_update_perturbation_short}
\end{align}
Thus, along states where the frozen label has high responsibility, one exact
MoG step and one frozen-label step are close. A full pathwise tracking theorem
would require additional stability assumptions controlling the accumulation of
these local errors, so we validate the whole-chain discrepancy empirically.

\clearpage
\section{MoG analysis of conditional DDIM with an oracle posterior denoiser}
\label{app:conditional_ddim_mog_derivation}

This appendix shows that the selected-label MoG analysis is not specific to the Brownian-bridge machinery. We repeat the same pipeline for a standard non-bridge conditional DDIM sampler. The forward process is the usual data-to-noise diffusion marginal, while the observation $y$ is incorporated only through the denoiser. 

\subsection*{H.1 Conditional DDIM model and notation}

In contrast to BBDM, DDIM does not use a bridge endpoint at $y$. Its forward marginal is
\begin{equation}
q_{\mathrm D}(x_s\mid x_0)
=
\N\!\left(x_s;\sqrt{\bar\alpha_s}\,x_0,(1-\bar\alpha_s)I\right),
\label{eq:ddim_forward_marginal_app}
\end{equation}
where $s\in\{0,\dots,S\}$, $\bar\alpha_0=1$, and $\bar\alpha_S$ is close to zero. \\
The measurement model and MoG prior are unchanged:
\begin{equation}
y\mid x_0\sim\N(Hx_0,\sigma_y^2I),
\qquad
p(x_0)=\sum_{r=1}^{R}\pi_r\N(x_0;\mu_r,\Sigma_r).
\label{eq:ddim_measurement_prior_app}
\end{equation}

Given an estimate $\hat x_0(x_s,y)$, the deterministic DDIM update can be written in the affine form
\begin{equation}
x_{s-1}=a_s^{\mathrm D}x_s+b_s^{\mathrm D}\hat x_0(x_s,y),
\label{eq:ddim_update_affine_app}
\end{equation}
where
\begin{equation}
a_s^{\mathrm D}
=
\frac{\sqrt{1-\bar\alpha_{s-1}}}{\sqrt{1-\bar\alpha_s}},
\label{eq:ddim_a_coeff_app}
\end{equation}
\begin{equation}
b_s^{\mathrm D}
=
\sqrt{\bar\alpha_{s-1}}
-
\frac{\sqrt{\bar\alpha_s}\sqrt{1-\bar\alpha_{s-1}}}{\sqrt{1-\bar\alpha_s}}.
\label{eq:ddim_b_coeff_app}
\end{equation}

The sampler is initialized as in standard DDIM with
\begin{equation}
x_S\sim\N(0,I),
\label{eq:ddim_terminal_init_app}
\end{equation}
independently of $y$. 

\subsection*{H.2 Exact posterior and oracle conditional DDIM denoiser}

We now compute the posterior $p(x_0\mid x_s,y)$ under the DDIM forward marginal. The measurement-only posterior within a fixed component is identical to the one used in the BBDM analysis:
\begin{equation}
\Sigma_{r\mid y}^{-1}
=
\Sigma_r^{-1}+\frac{1}{\sigma_y^2}H^\top H,
\qquad
\mu_{r\mid y}
=
\Sigma_{r\mid y}
\left(\Sigma_r^{-1}\mu_r+\frac{1}{\sigma_y^2}H^\top y\right).
\label{eq:ddim_component_y_posterior_app}
\end{equation}
Thus,
\begin{equation}
x_0\mid (y,C=r)\sim \N(\mu_{r\mid y},\Sigma_{r\mid y}).
\label{eq:ddim_x0_given_y_cr_app}
\end{equation}
Conditioned on $(y,C=r)$, the DDIM latent $x_s$ is an affine Gaussian observation of $x_0$:
\[
x_s=\sqrt{\bar\alpha_s}\,x_0+\sqrt{1-\bar\alpha_s}\,\epsilon.
\]
Applying the Gaussian conditioning identity from Appendix~\ref{app:linear_gaussian_identities}, with $A=\sqrt{\bar\alpha_s}I$ and $R=(1-\bar\alpha_s)I$, gives
\begin{equation}
p_{\mathrm D}(x_0\mid x_s,y,C=r)
=
\N\!\left(x_0;\mu^{\mathrm D}_{r\mid s}(x_s,y),\Sigma^{\mathrm D}_{r\mid s}\right),
\label{eq:ddim_component_posterior_app}
\end{equation}
where
\begin{equation}
\Sigma^{\mathrm D}_{r\mid s}
=
\left(\Sigma_{r\mid y}^{-1}+\frac{\bar\alpha_s}{1-\bar\alpha_s}I\right)^{-1},
\label{eq:ddim_component_sigma_app}
\end{equation}
\begin{equation}
\mu^{\mathrm D}_{r\mid s}(x_s,y)
=
\Sigma^{\mathrm D}_{r\mid s}
\left(
\Sigma_{r\mid y}^{-1}\mu_{r\mid y}
+
\frac{\sqrt{\bar\alpha_s}}{1-\bar\alpha_s}x_s
\right).
\label{eq:ddim_component_mu_compact_app}
\end{equation}
Equivalently, using \eqref{eq:ddim_component_y_posterior_app},
\begin{equation}
\mu^{\mathrm D}_{r\mid s}(x_s,y)
=
\Sigma^{\mathrm D}_{r\mid s}
\left(
\Sigma_r^{-1}\mu_r+\frac{1}{\sigma_y^2}H^\top y+
\frac{\sqrt{\bar\alpha_s}}{1-\bar\alpha_s}x_s
\right).
\label{eq:ddim_component_mu_expanded_app}
\end{equation}

The component probabilities are also obtained exactly. First,
\begin{equation}
\gamma_{r\mid y}(y)
=
\frac{
\pi_r\N\!\left(y;H\mu_r,H\Sigma_rH^\top+\sigma_y^2I\right)
}{
\sum_{j=1}^{R}\pi_j\N\!\left(y;H\mu_j,H\Sigma_jH^\top+\sigma_y^2I\right)
}.
\label{eq:ddim_gamma_y_app}
\end{equation}
Next, conditioned on $(y,C=r)$,
\begin{equation}
p_{\mathrm D}(x_s\mid y,C=r)
=
\N\!\left(
 x_s;\sqrt{\bar\alpha_s}\,\mu_{r\mid y},
 \bar\alpha_s\Sigma_{r\mid y}+(1-\bar\alpha_s)I
\right).
\label{eq:ddim_xs_given_y_cr_app}
\end{equation}
Therefore the DDIM responsibilities are
\begin{equation}
\gamma^{\mathrm D}_{r\mid s}(x_s,y)
\triangleq
\PP(C=r\mid x_s,y)
=
\frac{
\gamma_{r\mid y}(y)p_{\mathrm D}(x_s\mid y,C=r)
}{
\sum_{j=1}^{R}\gamma_{j\mid y}(y)p_{\mathrm D}(x_s\mid y,C=j)
}.
\label{eq:ddim_gamma_s_app}
\end{equation}
Combining the component posteriors and responsibilities yields
\begin{equation}
p_{\mathrm D}(x_0\mid x_s,y)
=
\sum_{r=1}^{R}
\gamma^{\mathrm D}_{r\mid s}(x_s,y)
\N\!\left(x_0;\mu^{\mathrm D}_{r\mid s}(x_s,y),\Sigma^{\mathrm D}_{r\mid s}\right).
\label{eq:ddim_exact_mog_posterior_app}
\end{equation}
Hence the oracle conditional DDIM denoiser is
\begin{equation}
\hat x^{\mathrm D}_{0,s}(x_s,y)
=
\E[x_0\mid x_s,y]
=
\sum_{r=1}^{R}
\gamma^{\mathrm D}_{r\mid s}(x_s,y)
\mu^{\mathrm D}_{r\mid s}(x_s,y).
\label{eq:ddim_exact_denoiser_app}
\end{equation}

\subsection*{H.3 Exact conditional DDIM reverse dynamics and loss of global affinity}

Substituting \eqref{eq:ddim_exact_denoiser_app} into the DDIM update \eqref{eq:ddim_update_affine_app} gives
\begin{equation}
x_{s-1}
=
a_s^{\mathrm D}x_s
+
b_s^{\mathrm D}
\sum_{r=1}^{R}\gamma^{\mathrm D}_{r\mid s}(x_s,y)
\mu^{\mathrm D}_{r\mid s}(x_s,y).
\label{eq:ddim_exact_reverse_before_expand_app}
\end{equation}
Using \eqref{eq:ddim_component_mu_compact_app}, this becomes
\begin{align}
x_{s-1}
&=
\left(
 a_s^{\mathrm D}I
 +b_s^{\mathrm D}\frac{\sqrt{\bar\alpha_s}}{1-\bar\alpha_s}
 \sum_{r=1}^{R}\gamma^{\mathrm D}_{r\mid s}(x_s,y)
 \Sigma^{\mathrm D}_{r\mid s}
\right)x_s
\notag\\
&\quad+
 b_s^{\mathrm D}
 \sum_{r=1}^{R}\gamma^{\mathrm D}_{r\mid s}(x_s,y)
 \Sigma^{\mathrm D}_{r\mid s}
 \Sigma_{r\mid y}^{-1}\mu_{r\mid y}.
\label{eq:ddim_exact_reverse_compact_app}
\end{align}

The structural obstruction is identical to the one found for BBDM. If $R=1$, then $\gamma^{\mathrm D}_{1\mid s}\equiv 1$ and the reverse update is globally affine. If $R>1$, the coefficients in \eqref{eq:ddim_exact_reverse_compact_app} depend on the current state through $\gamma^{\mathrm D}_{r\mid s}(x_s,y)$. Therefore, the exact MoG conditional DDIM chain cannot be unrolled by simply composing one fixed sequence of linear maps.

\subsection*{H.4 Selected-label approximation for conditional DDIM}

We now apply the same selected-label approximation used for BBDM. Draw
\begin{equation}
\PP(J=r\mid y)=\gamma_{r\mid y}(y),
\label{eq:ddim_selected_label_rule_app}
\end{equation}
and keep $J$ fixed along the DDIM reverse chain. Conditioned on $J=r$, replace the mixture denoiser by the component-conditioned posterior mean
\[
\hat x_0(x_s,y)=\mu^{\mathrm D}_{r\mid s}(x_s,y).
\]
Inserting \eqref{eq:ddim_component_mu_compact_app} into \eqref{eq:ddim_update_affine_app} gives the fixed-label conditional DDIM recursion
\begin{equation}
x_{s-1}
=
\left(a_s^{\mathrm D}I+b_s^{\mathrm D}\frac{\sqrt{\bar\alpha_s}}{1-\bar\alpha_s}\Sigma^{\mathrm D}_{r\mid s}\right)x_s
+
b_s^{\mathrm D}\Sigma^{\mathrm D}_{r\mid s}\Sigma_{r\mid y}^{-1}\mu_{r\mid y}.
\label{eq:ddim_fixed_label_update_app}
\end{equation}
The update is affine once the label is frozen. Unlike the BBDM update \eqref{eq:selected_label_reverse_update_main_v74}, there is no explicit bridge endpoint term proportional to $y$ and no reverse innovation term. The measurement affects the recursion only through the posterior mean $\mu_{r\mid y}$ and precision $\Sigma_{r\mid y}^{-1}$.

\subsection*{H.5 Component-wise basis and diagonal scalar recursions}

For each component $r$, use the same problem-adapted basis as in Section~\ref{sec:mog_selection_basis}:
\begin{equation}
\Sigma_{r\mid y}^{-1}
=
U_r\diag(\lambda_{r,1},\dots,\lambda_{r,d})U_r^\top.
\label{eq:ddim_basis_app}
\end{equation}
Since $\Sigma^{\mathrm D}_{r\mid s}$ is obtained by adding a scalar multiple of the identity to $\Sigma_{r\mid y}^{-1}$ and inverting, the same basis diagonalizes all DDIM posterior covariances:
\begin{equation}
\Sigma^{\mathrm D}_{r\mid s}=U_r\Lambda^{\mathrm D}_{r\mid s}U_r^\top,
\label{eq:ddim_sigma_diag_app}
\end{equation}
where
\begin{equation}
\Lambda^{\mathrm D}_{r\mid s}
=
\diag\!\left(
\frac{1}{\lambda_{r,1}+\frac{\bar\alpha_s}{1-\bar\alpha_s}},
\dots,
\frac{1}{\lambda_{r,d}+\frac{\bar\alpha_s}{1-\bar\alpha_s}}
\right).
\label{eq:ddim_lambda_diag_app}
\end{equation}
Projecting \eqref{eq:ddim_fixed_label_update_app} onto this basis and defining
\[
x_s^{U_r}=U_r^\top x_s,
\qquad
\mu_{r\mid y}^{U_r}=U_r^\top\mu_{r\mid y},
\qquad
x_S^{U_r}=U_r^\top x_S,
\]
we obtain
\begin{equation}
x_{s-1}^{U_r}
=
G_r^{\mathrm D}(s)x_s^{U_r}
+
M_r^{\mathrm D}(s)\mu_{r\mid y}^{U_r},
\label{eq:ddim_fixed_label_basis_update_app}
\end{equation}
with diagonal matrices
\begin{equation}
G_r^{\mathrm D}(s)
=
a_s^{\mathrm D}I+b_s^{\mathrm D}\frac{\sqrt{\bar\alpha_s}}{1-\bar\alpha_s}\Lambda^{\mathrm D}_{r\mid s},
\label{eq:ddim_G_app}
\end{equation}
\begin{equation}
M_r^{\mathrm D}(s)
=
b_s^{\mathrm D}\Lambda^{\mathrm D}_{r\mid s}
\diag(\lambda_{r,1},\dots,\lambda_{r,d}).
\label{eq:ddim_M_app}
\end{equation}
Thus, conditioned on $J=r$, the DDIM chain also decouples into $d$ scalar recursions in the basis $U_r$.

\subsection*{H.6 Unrolled reconstruction law}

Starting from $x_S\sim\N(0,I)$, and using the convention that empty products are the identity, the recursion
\eqref{eq:ddim_fixed_label_basis_update_app} gives
\begin{equation}
\hat x_{0,\mathrm D}^{U_r,(r)}
=
\left(\prod_{s'=1}^{S}G_r^{\mathrm D}(s')\right)x_S^{U_r}
+
D_2^{(r),\mathrm D}\mu_{r\mid y}^{U_r},
\label{eq:ddim_unrolled_app}
\end{equation}
where
\begin{equation}
D_2^{(r),\mathrm D}
=
\sum_{i=1}^{S}
\left(\prod_{j=1}^{i-1}G_r^{\mathrm D}(j)\right)
M_r^{\mathrm D}(i).
\label{eq:ddim_D2_app}
\end{equation}
The matrix $D_2^{(r),\mathrm D}$ and the product
$\prod_{s'=1}^{S}G_r^{\mathrm D}(s')$ are diagonal. Since
$x_S^{U_r}\sim\N(0,I)$, the fixed-label conditional DDIM output satisfies
\begin{equation}
p^{\mathrm D}_{\mathrm{Select}}(\hat x_0\mid y,J=r)
=
\N\!\left(
\hat x_0;
\mu_{\mathrm{DDIM}}^{(r)}(y),
\Sigma_{\mathrm{DDIM}}^{(r)}
\right),
\label{eq:ddim_selected_component_law_app}
\end{equation}
with
\begin{equation}
\mu_{\mathrm{DDIM}}^{(r)}(y)
=
U_rD_2^{(r),\mathrm D}\mu_{r\mid y}^{U_r},
\label{eq:ddim_selected_mean_app}
\end{equation}
and
\begin{equation}
\Sigma_{\mathrm{DDIM}}^{(r)}
=
U_r
\left[
\left(\prod_{s'=1}^{S}G_r^{\mathrm D}(s')\right)
\left(\prod_{s'=1}^{S}G_r^{\mathrm D}(s')\right)^\top
\right]
U_r^\top.
\label{eq:ddim_selected_cov_app}
\end{equation}
Removing the conditioning on the selected label gives the mixture law
\begin{equation}
p^{\mathrm D}_{\mathrm{Select}}(\hat x_0\mid y)
=
\sum_{r=1}^{R}
\gamma_{r\mid y}(y)
\N\!\left(
\hat x_0;
\mu_{\mathrm{DDIM}}^{(r)}(y),
\Sigma_{\mathrm{DDIM}}^{(r)}
\right).
\label{eq:ddim_selected_mixture_law_app}
\end{equation}
This is the conditional-DDIM analogue of Corollary~\ref{cor:selected_label_reconstruction_law}. The weights are the same measurement-posterior weights as in the true posterior $p(x_0\mid y)$, while the Gaussian attached to each component is induced by the DDIM reverse dynamics.

We now use the same endpoint-idealized viewpoint as in the BBDM mean-exactness analysis. In practice, DDIM schedules take $\bar\alpha_S$ close to zero. Approximating this terminal value by
\[
\bar\alpha_S = 0,
\]
the initialization $x_S\sim\N(0,I)$ satisfies
\[
\E[x_S\mid y,J=r]=0=\sqrt{\bar\alpha_S}\,\mu_{r\mid y}.
\]
Repeating the same backward-induction argument used for the selected-label BBDM mean-exactness result gives
\begin{equation}
\E[\hat x_0\mid y,J=r]
=
\mu_{r\mid y}.
\label{eq:ddim_mean_exact_app}
\end{equation}
Equivalently, in the unrolled representation,
\begin{equation}
D_2^{(r),\mathrm D}=I.
\label{eq:ddim_D2_identity_app}
\end{equation}
Therefore, under the endpoint approximation $\bar\alpha_S\simeq 0$, the fixed-label conditional DDIM law simplifies to
\begin{equation}
p^{\mathrm D}_{\mathrm{Select}}(\hat x_0\mid y,J=r)
=
\N\!\left(
\hat x_0;
\mu_{r\mid y},
\Sigma_{\mathrm{DDIM}}^{(r)}
\right),
\label{eq:ddim_selected_component_law_mean_exact_app}
\end{equation}
and the selected-label mixture law becomes
\begin{equation}
p^{\mathrm D}_{\mathrm{Select}}(\hat x_0\mid y)
=
\sum_{r=1}^{R}
\gamma_{r\mid y}(y)
\N\!\left(
\hat x_0;
\mu_{r\mid y},
\Sigma_{\mathrm{DDIM}}^{(r)}
\right).
\label{eq:ddim_selected_mixture_law_mean_exact_app}
\end{equation}
Thus, after the same type of endpoint idealization used for BBDM, the selected-label conditional DDIM sampler is also component-wise mean exact. The remaining discrepancy from the true posterior is therefore in the component covariances.

\subsection*{H.7 Matched posterior comparison and DDIM schedule objectives}

The true measurement posterior remains
\begin{equation}
p(x_0\mid y)
=
\sum_{r=1}^{R}\gamma_{r\mid y}(y)
\N(x_0;\mu_{r\mid y},\Sigma_{r\mid y}).
\label{eq:ddim_true_y_posterior_app}
\end{equation}
Since \eqref{eq:ddim_selected_mixture_law_mean_exact_app} and
\eqref{eq:ddim_true_y_posterior_app} share the same component weights, the same matched-index coupling used in Section~\ref{sec:comparison_v6} gives
\begin{align}
&W_2^2\!\left(
p(x_0\mid y),
p^{\mathrm D}_{\mathrm{Select}}(\hat x_0\mid y)
\right)
\notag\\
&\le
\sum_{r=1}^{R}\gamma_{r\mid y}(y)\,
W_2^2\!\left(
\N(\mu_{r\mid y},\Sigma_{r\mid y}),
\N(\mu_{r\mid y},\Sigma_{\mathrm{DDIM}}^{(r)})
\right).
\label{eq:ddim_w2_bound_app}
\end{align}
Define
\begin{equation}
\sigma_{\mathrm{DDIM},r,k}
=
\left|
\left[
\prod_{s'=1}^{S}G_r^{\mathrm D}(s')
\right]_{kk}
\right|.
\label{eq:ddim_sigma_def_app}
\end{equation}
In the $U_r$ basis, the true posterior covariance is
\[
\diag\!\left(
\frac{1}{\lambda_{r,1}},
\dots,
\frac{1}{\lambda_{r,d}}
\right),
\]
while the DDIM-induced covariance is
\[
\diag\!\left(
\sigma_{\mathrm{DDIM},r,1}^2,
\dots,
\sigma_{\mathrm{DDIM},r,d}^2
\right).
\]
Because the means match under the endpoint approximation, the matched Gaussian Wasserstein cost contains no mean term:
\begin{align}
& W_2^2\!\left(
\N(\mu_{r\mid y},\Sigma_{r\mid y}),
\N(\mu_{r\mid y},\Sigma_{\mathrm{DDIM}}^{(r)})
\right)
\notag\\
&=
\sum_{k=1}^{d}
\left(
\sigma_{\mathrm{DDIM},r,k}
-
\frac{1}{\sqrt{\lambda_{r,k}}}
\right)^2.
\label{eq:ddim_matched_gaussian_cost_app}
\end{align}
Averaging over $y$ as in Appendix~\ref{app:schedule_objectives_derivations} and using
\[
\E_{p(y)}\sum_{r=1}^{R}\gamma_{r\mid y}(y)f_r(y)
=
\sum_{r=1}^{R}\pi_r\E_{p(y\mid C=r)}f_r(y),
\]
we obtain the endpoint-approximated DDIM Wasserstein schedule objective
\begin{equation}
J_{\mathrm{MOG,DDIM}}^{W_2}
=
\sum_{r=1}^{R}\pi_r
\sum_{k=1}^{d}
\left(
\sigma_{\mathrm{DDIM},r,k}
-
\frac{1}{\sqrt{\lambda_{r,k}}}
\right)^2.
\label{eq:ddim_w2_objective_app}
\end{equation}

The corresponding matched-label product-coupling expected-MSE objective is
\begin{equation}
J_{\mathrm{MOG,DDIM}}^{\mathrm{MSE}}
=
\sum_{r=1}^{R}\pi_r
\sum_{k=1}^{d}
\left(
\sigma_{\mathrm{DDIM},r,k}^2
+
\frac{1}{\lambda_{r,k}}
\right).
\label{eq:ddim_mse_objective_app}
\end{equation}

\clearpage
\section{Additional experimental figures}
\label{app:extra_figures_v43}

\begin{figure}[htbp]
    \centering
    \includegraphics[width=0.85\textwidth]{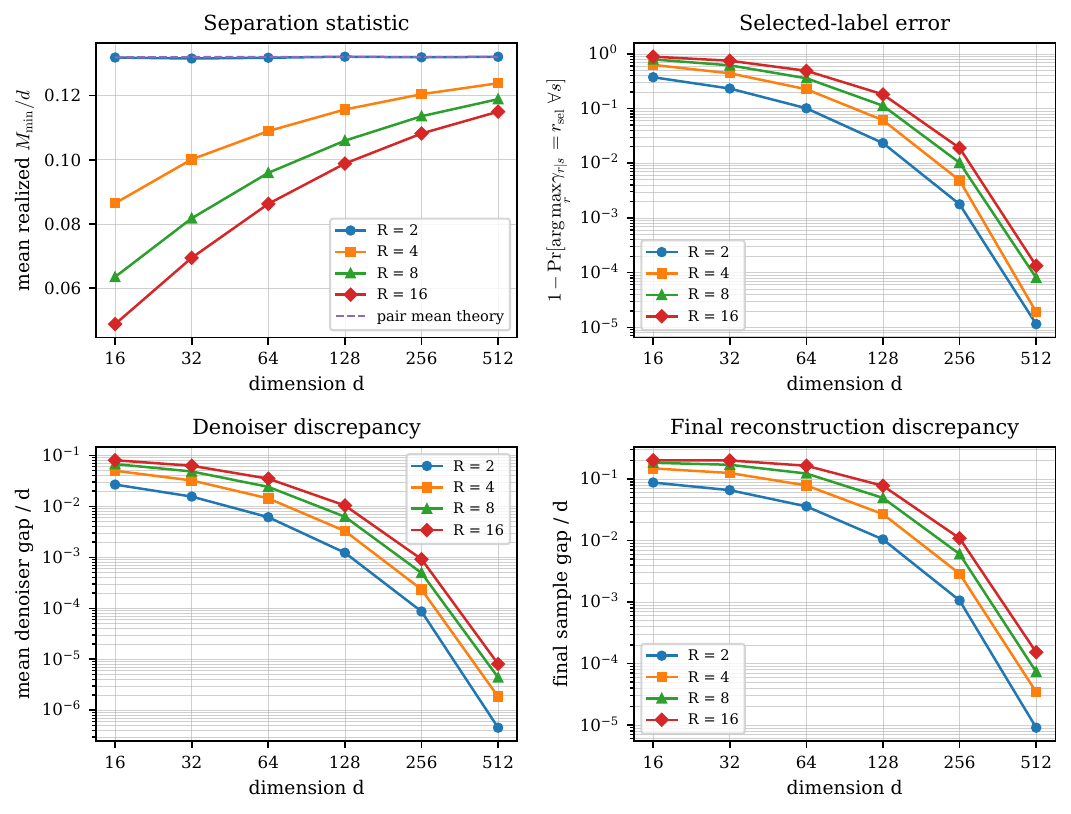}
    \caption{High-dimensional validation of the frozen-selection approximation in the controlled shared-covariance MoG setting. For each ambient dimension $d\in\{16,32,64,128,256,512\}$ and number of components $R\in\{2,4,8,16\}$, the component means are sampled independently as $\mu_{r,k}\sim\mathrm{Unif}[-1,1]$. The inverse problem is denoising with $H=I$ and $\sigma_y=2$, and all components share the diagonal covariance
    $\Sigma=\operatorname{diag}(\operatorname{geomspace}(0.5,2,d))$.
    The BBDM reverse chains use $S=20$ steps. In accordance with the selected-label approximation, the frozen label is sampled from the measurement posterior,
    $r_{\rm sel}\sim\mathrm{Categorical}(\gamma_{\cdot|y}(y))$,
    and is then kept fixed throughout the reverse chain. The exact MoG chain and the frozen selected-label chain are run with the same Gaussian innovations. The top-left panel reports the realized minimum measurement separation $M_{\min}/d$, where
    $M_{\min}=\min_{r\neq j}(\mu_r-\mu_j)^\top(\Sigma+\sigma_y^2 I)^{-1}(\mu_r-\mu_j)$,
    together with the pairwise mean-theory reference. The remaining panels show the selected-label error,  a time-averaged local
denoiser discrepancy per dimension, and the final reconstruction discrepancy per dimension. The states that both denoisers are evaluated on are the states of the frozen path. The selected-label error measures whether the sampled frozen label remains the dominant responsibility along the frozen chain. As the dimension increases, the measurement posterior becomes increasingly one-hot, the sampled label stabilizes, and the exact and frozen chains become increasingly close, consistent with the responsibility-concentration mechanism predicted by the theory.} 
    \label{fig:main_theorem_validation_app_v43}
\end{figure}

\begin{figure}[htbp]
    \centering
    \includegraphics[width=1\textwidth]{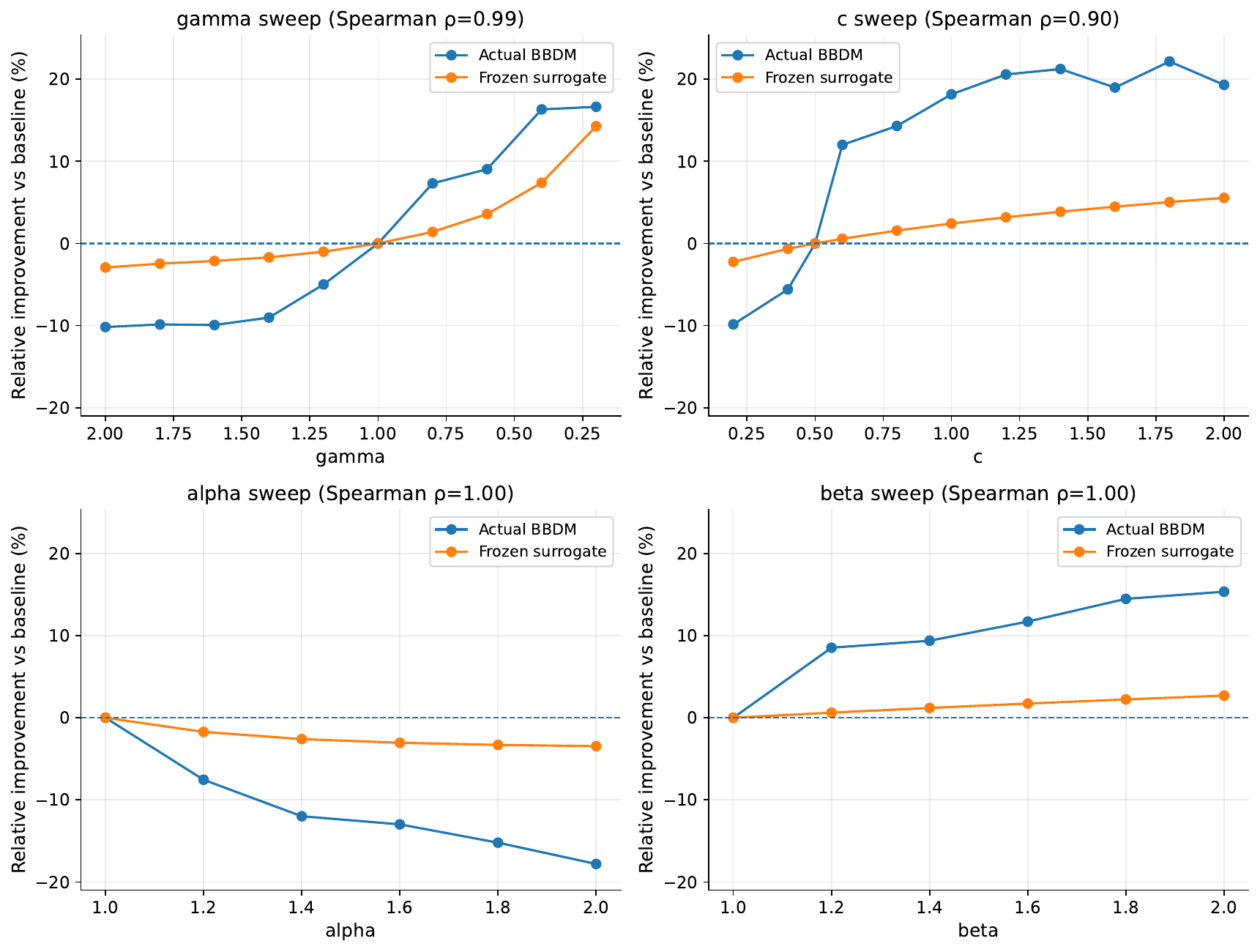}
    \caption{One-dimensional MNIST schedule sweeps for the baseline blur
setting $(V,\sigma_y)=(0.10,0.10)$. In each panel, one parameter in
$(\alpha,\beta,c,\gamma)$ is varied while the other three are fixed. The
curves compare relative improvement in reconstruction MSE to the
ground-truth clean digit over the MNIST test set for a trained BBDM and the
frozen selected-label surrogate fitted with 70 Gaussians per digit. Both
methods favor the same directions:
$\alpha\downarrow$, $\beta\uparrow$, $c\uparrow$, and $\gamma\downarrow$.
The corresponding Spearman rank correlations between the surrogate and
trained-BBDM sweeps are
$\rho_{\alpha}=1.00$, $\rho_{\beta}=1.00$,
$\rho_{c}=0.90$, and $\rho_{\gamma}=0.99$.
The comparison tests agreement in schedule ranking and favorable direction,
rather than equality of the absolute MSE values.}
    \label{fig:mnist_schedule_sweeps_app_v88}
\end{figure}

\begin{figure}[htbp]
    \centering
    \includegraphics[width=1\textwidth]{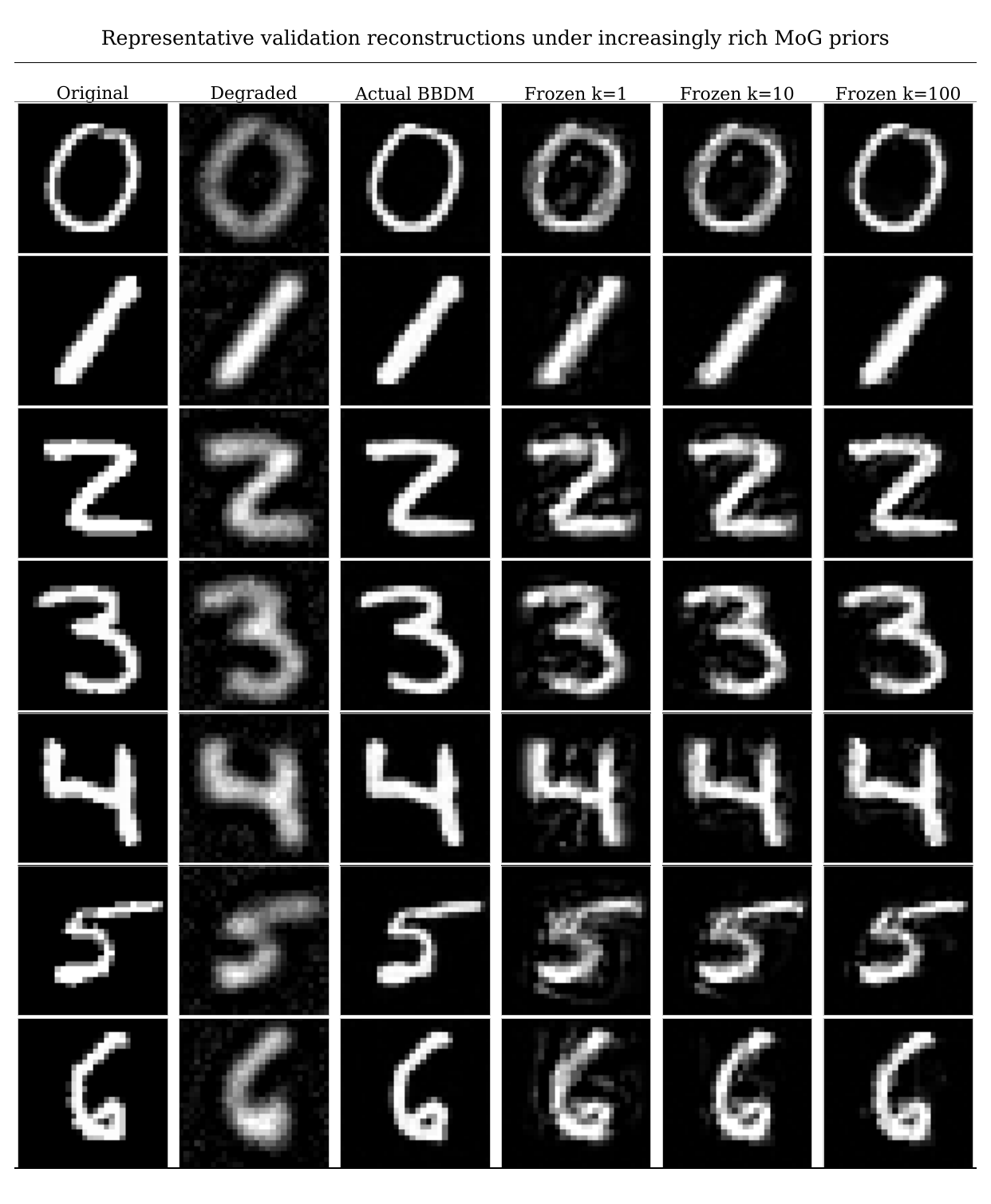}
\caption{Effect of prior richness in the frozen selected-label surrogate for
MNIST deblurring. The number of fitted Gaussian components is increased from
$k=1$ to $k=100$ per digit while the reconstruction rule and degradation
setting are kept fixed. Increasing the richness of the fitted MoG produces
cleaner and more stable reconstructions, supporting the use of a sufficiently
expressive prior in the surrogate schedule search. These are outputs of the
mathematical surrogate, not separately trained BBDM models.}
    \label{fig:digit_strip_progression_app_v43}
\end{figure}

\begin{figure}[htbp]
    \centering
    \includegraphics[width=0.88\textwidth]{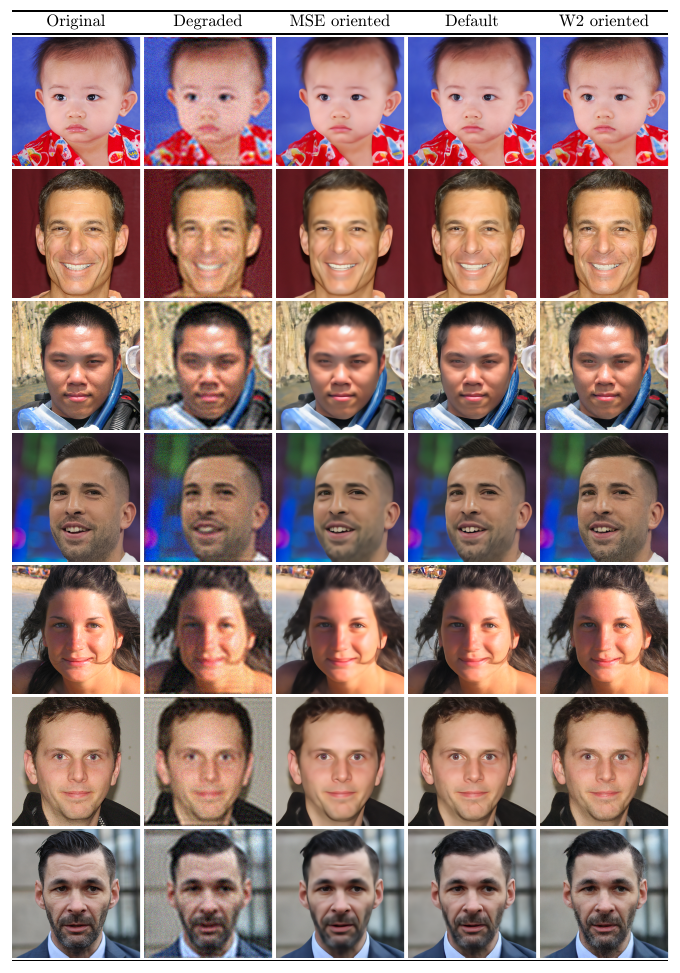}
    \caption{Qualitative FFHQ blur reconstructions for $(V,\sigma_y)=(0.03,0.10)$. Columns show the original image, degraded observation, MSE-oriented reconstruction, default reconstruction, and $W_2$-oriented reconstruction.}
    \label{fig:ffhq_v003_qualitative_app}
\end{figure}

\begin{figure}[htbp]
    \centering
    \includegraphics[width=0.88\textwidth]{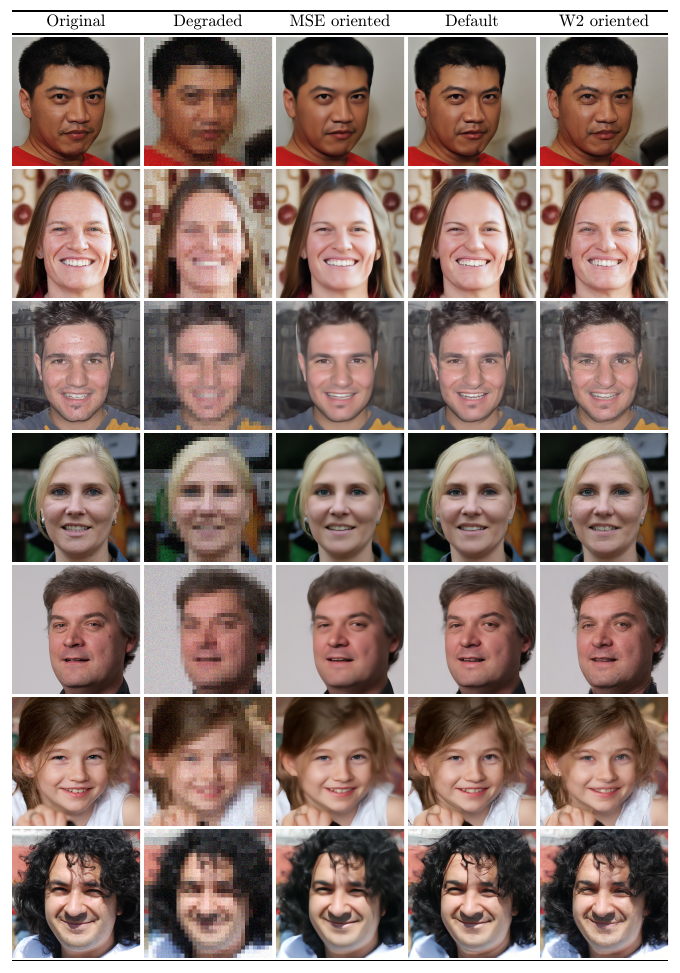}
    \caption{Qualitative FFHQ $8\times$ super-resolution reconstructions with $\sigma_y=0.1$. Columns show the original image, degraded observation, MSE-oriented reconstruction, default reconstruction, and $W_2$-oriented reconstruction.}
    \label{fig:ffhq_sr8_qualitative_app}
\end{figure}

\begin{figure}[htbp]
    \centering
    \includegraphics[width=0.88\textwidth]{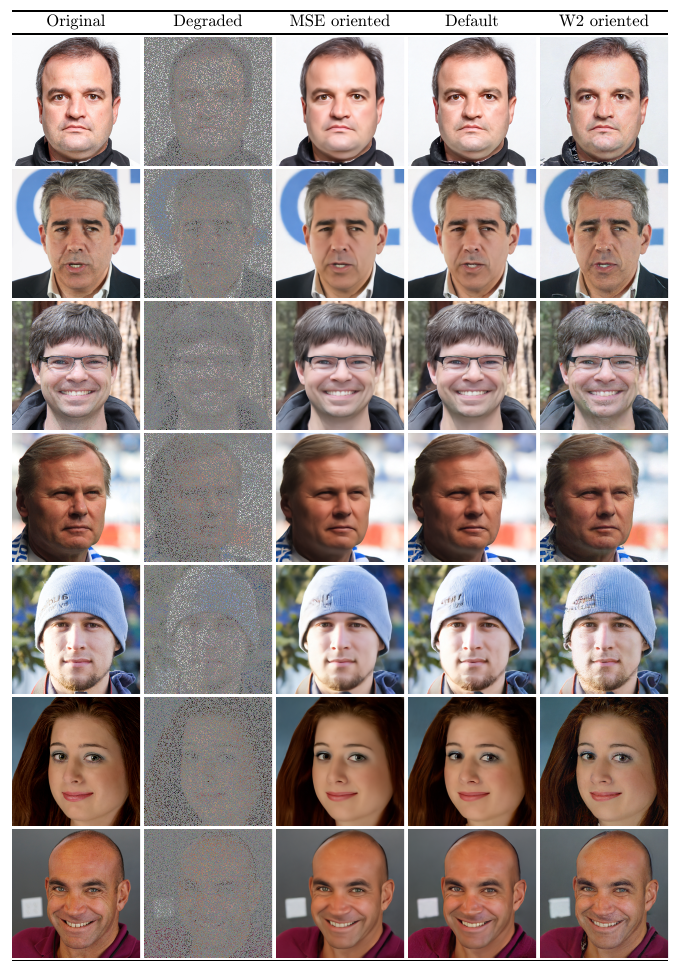}
    \caption{Qualitative FFHQ distributed inpainting reconstructions for $p=0.125$ and $\sigma_y=0.1$. Columns show the original image, degraded observation, MSE-oriented reconstruction, default reconstruction, and $W_2$-oriented reconstruction.}
    \label{fig:ffhq_inpaint12p5_qualitative_app}
\end{figure}

Figure~\ref{fig:main_theorem_validation_app_v43} provides the full
high-dimensional validation plot discussed in Section~\ref{sec:experiments_v7}.
Figure~\ref{fig:mnist_schedule_sweeps_app_v88} shows the four one-dimensional
MNIST schedule sweeps discussed in Section~\ref{sec:experiments_v7}.
Figure~\ref{fig:digit_strip_progression_app_v43} shows representative MNIST
reconstructions under increasingly rich fitted MoG priors for the deblurring
inverse problem. Table~\ref{tab:ffhq_sampling_steps_appendix} shows the FFHQ
inverse-problem results for different numbers of sampling steps.
Figures~\ref{fig:ffhq_v003_qualitative_app},
\ref{fig:ffhq_sr8_qualitative_app}, and
\ref{fig:ffhq_inpaint12p5_qualitative_app} show qualitative FFHQ
reconstructions for the blur setting \((V,\sigma_y)=(0.03,0.10)\), the
\(8\times\) super-resolution task, and the \(p=0.125\) distributed inpainting
inverse problem.

\FloatBarrier

\FloatBarrier

\begin{table}[p]
\caption{FFHQ schedule comparison across sampling budgets and
inverse-problem settings. }
\label{tab:ffhq_sampling_steps_appendix}

\begin{center}
\begingroup
\setlength{\tabcolsep}{2pt}

\begin{tabular*}{\linewidth}{
    @{\extracolsep{\fill}}
    cllrrrrrrrr
    @{}
}
\toprule
$S$ & Metric & Sch. &
$B_1$ & $B_2$ & $B_3$ & $B_4$ &
$\mathrm{SR}_4$ & $\mathrm{SR}_8$ &
$I_{25}$ & $I_{12.5}$ \\
\midrule


\multirow{12}{*}{20}
& \multirow{3}{*}{FID$\downarrow$}
& def.
& 14.544 & 19.785 & 9.938 & 21.876
& 14.520 & 26.046 & 17.609 & 19.742 \\

& & MSE
& 21.874 & 25.082 & 13.165 & 26.871
& 23.968 & 34.715 & 19.159 & 22.853 \\

& & $W_2$
& \textbf{9.533} & \textbf{14.666}
& \textbf{6.493} & \textbf{12.257}
& \textbf{11.201} & \textbf{17.778}
& \textbf{8.936} & \textbf{9.914} \\

\cmidrule(lr){2-11}

& \multirow{3}{*}{LPIPS$\downarrow$}
& def.
& 0.0850 & 0.1447 & 0.0450 & 0.1133
& 0.0990 & 0.1820 & 0.0860 & 0.1152 \\

& & MSE
& 0.1041 & 0.1741 & 0.0511 & 0.1225
& 0.1279 & 0.2069 & 0.0855 & 0.1221 \\

& & $W_2$
& \textbf{0.0741} & \textbf{0.1382}
& \textbf{0.0366} & \textbf{0.0955}
& \textbf{0.0919} & \textbf{0.1766}
& \textbf{0.0725} & \textbf{0.1128} \\

\cmidrule(lr){2-11}

& \multirow{3}{*}{PSNR$\uparrow$}
& def.
& 31.277 & 27.970 & 34.066 & 30.209
& 29.614 & 25.745 & 30.487 & 28.017 \\

& & MSE
& \textbf{31.814} & \textbf{28.579}
& \textbf{34.883} & \textbf{30.834}
& \textbf{30.205} & \textbf{26.236}
& \textbf{31.316} & \textbf{28.819} \\

& & $W_2$
& 30.882 & 28.007 & 33.467 & 29.797
& 29.384 & 25.588 & 28.618 & 25.873 \\

\cmidrule(lr){2-11}

& \multirow{3}{*}{SSIM$\uparrow$}
& def.
& 0.878 & 0.795 & 0.923 & 0.857
& 0.853 & 0.743 & 0.876 & 0.828 \\

& & MSE
& \textbf{0.888} & \textbf{0.807}
& \textbf{0.934} & \textbf{0.868}
& \textbf{0.866} & \textbf{0.758}
& \textbf{0.889} & \textbf{0.844} \\

& & $W_2$
& 0.866 & 0.788 & 0.913 & 0.844
& 0.837 & 0.736 & 0.839 & 0.771 \\

\midrule


\multirow{12}{*}{50}
& \multirow{3}{*}{FID$\downarrow$}
& def.
& 13.223 & 18.109 & 9.143 & 20.148
& 13.515 & 23.221 & 16.364 & 17.789 \\

& & MSE
& 21.878 & 25.066 & 13.169 & 26.871
& 23.997 & 34.823 & 19.199 & 22.861 \\

& & $W_2$
& \textbf{5.805} & \textbf{10.041}
& \textbf{3.653} & \textbf{7.134}
& \textbf{6.545} & \textbf{12.491}
& \textbf{5.238} & \textbf{8.874} \\

\cmidrule(lr){2-11}

& \multirow{3}{*}{LPIPS$\downarrow$}
& def.
& 0.0784 & 0.1351 & 0.0427 & 0.1067
& 0.0929 & 0.1715 & 0.0833 & \textbf{0.1113} \\

& & MSE
& 0.1039 & 0.1735 & 0.0511 & 0.1223
& 0.1277 & 0.2055 & 0.0855 & 0.1211 \\

& & $W_2$
& \textbf{0.0606} & \textbf{0.1195}
& \textbf{0.0297} & \textbf{0.0804}
& \textbf{0.0771} & \textbf{0.1586}
& \textbf{0.0663} & 0.1186 \\

\cmidrule(lr){2-11}

& \multirow{3}{*}{PSNR$\uparrow$}
& def.
& 30.973 & 27.608 & 33.820 & 29.886
& 29.289 & 25.369 & 30.170 & 27.641 \\

& & MSE
& \textbf{31.800} & \textbf{28.561}
& \textbf{34.876} & \textbf{30.822}
& \textbf{30.187} & \textbf{26.138}
& \textbf{31.315} & \textbf{28.803} \\

& & $W_2$
& 30.392 & 27.623 & 32.898 & 29.334
& 28.905 & 25.146 & 28.201 & 25.222 \\

\cmidrule(lr){2-11}

& \multirow{3}{*}{SSIM$\uparrow$}
& def.
& 0.873 & 0.786 & 0.920 & 0.851
& 0.841 & 0.731 & 0.871 & 0.819 \\

& & MSE
& \textbf{0.888} & \textbf{0.807}
& \textbf{0.934} & \textbf{0.869}
& \textbf{0.861} & \textbf{0.756}
& \textbf{0.890} & \textbf{0.845} \\

& & $W_2$
& 0.853 & 0.772 & 0.903 & 0.826
& 0.822 & 0.710 & 0.823 & 0.747 \\

\midrule


\multirow{12}{*}{100}
& \multirow{3}{*}{FID$\downarrow$}
& def.
& 12.369 & 17.133 & 8.656 & 18.597
& 12.800 & 20.845 & 15.347 & 16.390 \\

& & MSE
& 21.880 & 25.067 & 13.159 & 26.811
& 23.968 & 34.573 & 19.186 & 22.867 \\

& & $W_2$
& \textbf{5.101} & \textbf{9.352}
& \textbf{3.062} & \textbf{6.170}
& \textbf{5.366} & \textbf{11.832}
& \textbf{4.781} & \textbf{10.078} \\

\cmidrule(lr){2-11}

& \multirow{3}{*}{LPIPS$\downarrow$}
& def.
& 0.0744 & 0.1309 & 0.0414 & 0.1019
& 0.0891 & 0.1653 & 0.0812 & \textbf{0.1084} \\

& & MSE
& 0.1040 & 0.1734 & 0.0511 & 0.1223
& 0.1277 & 0.2052 & 0.0855 & 0.1211 \\

& & $W_2$
& \textbf{0.0572} & \textbf{0.1148}
& \textbf{0.0281} & \textbf{0.0763}
& \textbf{0.0729} & \textbf{0.1540}
& \textbf{0.0659} & 0.1229 \\

\cmidrule(lr){2-11}

& \multirow{3}{*}{PSNR$\uparrow$}
& def.
& 30.738 & 27.391 & 33.604 & 29.658
& 29.077 & 25.163 & 29.963 & 27.415 \\

& & MSE
& \textbf{31.794} & \textbf{28.551}
& \textbf{34.873} & \textbf{30.816}
& \textbf{30.176} & \textbf{26.099}
& \textbf{31.313} & \textbf{28.795} \\

& & $W_2$
& 30.144 & 27.430 & 32.629 & 29.098
& 28.666 & 24.955 & 28.000 & 24.955 \\

\cmidrule(lr){2-11}

& \multirow{3}{*}{SSIM$\uparrow$}
& def.
& 0.869 & 0.780 & 0.917 & 0.846
& 0.836 & 0.722 & 0.867 & 0.813 \\

& & MSE
& \textbf{0.888} & \textbf{0.807}
& \textbf{0.934} & \textbf{0.869}
& \textbf{0.860} & \textbf{0.756}
& \textbf{0.890} & \textbf{0.845} \\

& & $W_2$
& 0.845 & 0.762 & 0.898 & 0.817
& 0.812 & 0.699 & 0.814 & 0.734 \\

\bottomrule
\end{tabular*}

\vspace{4pt}

\parbox{\linewidth}{
\emph{Settings:}
$B_1$: blur with $(V,\sigma_y)=(0.10,0.10)$;
$B_2$: blur with $(V,\sigma_y)=(0.03,0.10)$;
$B_3$: blur with $(V,\sigma_y)=(0.30,0.10)$;
$B_4$: blur with $(V,\sigma_y)=(0.10,0.20)$;
$\mathrm{SR}_4$: $4\times$ super-resolution with $\sigma_y=0.10$;
$\mathrm{SR}_8$: $8\times$ super-resolution with $\sigma_y=0.10$;
$I_{25}$: inpainting with $(p,\sigma_y)=(0.25,0.10)$;
$I_{12.5}$: inpainting with $(p,\sigma_y)=(0.125,0.10)$.
}

\endgroup
\end{center}
\end{table}

\FloatBarrier

\end{document}